\documentclass{article} % For LaTeX2e
\usepackage{iclr2027_conference,times}

\usepackage{amsmath,amsfonts,bm}

\def\eqref#1{equation~\ref{#1}}
\def\1{\bm{1}}

\DeclareMathAlphabet{\mathsfit}{\encodingdefault}{\sfdefault}{m}{sl}
\SetMathAlphabet{\mathsfit}{bold}{\encodingdefault}{\sfdefault}{bx}{n}

\usepackage[dvipsnames]{xcolor}
\usepackage{xurl}
\usepackage[breaklinks,colorlinks,
citecolor = Blue,
]{hyperref}

\usepackage{url}
\usepackage{graphicx}
\usepackage{booktabs}
\usepackage{cleveref}
\usepackage{array}
\usepackage{makecell}
\usepackage{pifont}
\newcommand{\cmark}{\ding{51}}
\newcommand{\xmark}{\ding{55}}
\usepackage{subcaption}
\usepackage{xspace}
\usepackage{wrapfig}
\usepackage{float}
\usepackage{subcaption}
\usepackage{tcolorbox}
\usepackage{xcolor}
\usepackage{tablefootnote}
\usepackage{enumitem}
\usepackage{titletoc}

\def\eg{\textit{e.g.}\@\xspace}

\Crefname{section}{Section}{Sections}
\Crefname{table}{Table}{Tables}
\crefname{table}{Tab.}{Tabs.}
\crefname{appendix}{appendix}{appendices}
\Crefname{appendix}{Appendix}{Appendices}

\crefname{subappendix}{appendix}{appendices}
\Crefname{subappendix}{Appendix}{Appendices}

\AddToHook{cmd/appendix/before}{%
  \crefalias{section}{appendix}%
  \crefalias{subsection}{subappendix}%
}

\newcommand{\myparagraph}[1]{{\noindent\textbf{#1}}}

\title{CounterCount: Diagnosing Knowledge Bias in How Vision-Language Models Count Attributes}

\author{Reem Alzahrani\textsuperscript{1}\thanks{Correspondence to \texttt{reem.alzahrani@kaust.edu.sa}}\quad Bushra Bin Hemid\textsuperscript{1} \quad Hassan Alshanqiti\textsuperscript{2} \quad Zaid Alyafeai\textsuperscript{1} \quad\\ \textbf{Paul Gavrikov\textsuperscript{1} \quad Bernard Ghanem\textsuperscript{1} \quad Abdelrahman Eldesokey\textsuperscript{3}}  \\
\textsuperscript{1}KAUST\quad \textsuperscript{2}University of Edinburgh \quad \textsuperscript{3}HBKU \\
}

\iclrfinalcopy % Uncomment for camera-ready version, but NOT for submission.
\begin{document}

\maketitle
\lhead{Preprint}

\begin{abstract}
Do Vision-Language Models (VLMs) count what they see, or what they know about the world? 
We introduce \emph{CounterCount}, a diagnostic dataset of paired factual and counterfactual (CF) images, in which the CF image contains an atypical count of a subject's attribute (\eg, a rabbit with four ears), together with annotations localizing the counted attributes.
Humans answer both types of images nearly perfectly ($98.9\%$ and $97.5\%$), but across 16 VLMs (14 open-weight and 2 closed-source), average accuracy drops from $91.6\%$ on factual to $49.5\%$ on CF images, with most errors stemming from responses that match the canonical count, therefore revealing a bias towards knowledge priors rather than reliance on visual information.
Based on these findings, we perform an analysis that reveals two interesting insights behind this gap. First, models exhibit a ``System-1''-like response behavior, falling back on canonical counts after a shallow recognition of the subject. Second, model activations partially contain the visual evidence necessary to answer correctly, yet models underutilize it, attending less to the counted attributes when they provide biased answers. 
Input manipulations that break visual context, introduce cueing or reasoning, or direct amplification of the model's internal attention to relevant regions can reduce such biases. 
% \emph{CounterCount} goes beyond showing that VLMs rely on knowledge priors, revealing that once a subject is recognized, models trust their knowledge over the image and pay insufficient attention to the regions of the counted attributes rather than failing to see them.
\emph{CounterCount} goes beyond showing that VLMs fall back on knowledge priors, revealing that once a familiar context is recognized, models favor prior knowledge over available visual evidence rather than failing to encode that evidence.
\end{abstract}

\section{Introduction}
\label{sec:intro}

Vision-Language Models (VLMs) have advanced rapidly in recent years~\citep{Alayrac2022Flamingo, Li2022BLIP, Liu2023visual, bai2023qwenvl, bai2025qwen3vl, gemma2025gemma3, wang2025internvl35}, achieving strong performance on Visual Question Answering (VQA) and multimodal reasoning~\citep{Liu2023visual, Zhang2024vision, Yue2024MMMU, liu2024mmbench, Antol2015vqa, Hudson2019gqa, lu2024mathvista, Yu2024mmvet}.
Yet strong benchmark performance can mask a reliance on knowledge priors, learned from the statistical regularities of large-scale text corpora and the visual patterns encountered during training~\citep{vo2025bscore, sheng2019woman, gallegos2024bias, vo2025vision, vo2025vision2, Rahmanzadehgervi2024vision, lee2025vlind, fu2025hidden}, rather than genuine grounding in the image.

In general VQA, such reliance is often easy to expose: if a model describes the Statue of Liberty as holding a torch when the image shows a sword, the prior is clearly at work~\citep{lee2025vlind}.
Counting attributes is more subtle, as a wrong count may stem either from a default to the subject's canonical count or from a genuine limitation in counting ability.
Studies on general object counting, where no canonical prior is involved, help separate the two.
While some VLMs can still make errors at small counts~\citep{anh2026countingchorevlms, alghisi2025dereconstructing}, these studies consistently find that counting accuracy degrades as the number of objects increases~\citep{sengupta2025can, qharabagh2025lvlmcount, alghisi2025dereconstructing}.
Small counts therefore offer the setting where counting ability is least likely to be the bottleneck, and where errors most directly reflect prior-driven defaults.
But the few datasets that explicitly study counting bias~\citep{chen2026cdhbench, vo2025vision2} are limited in scale, often involve large counts that confound bias with counting difficulty, or contain ambiguous images.

\begin{figure}[htbp]
    \centering
    \includegraphics[width=\linewidth]{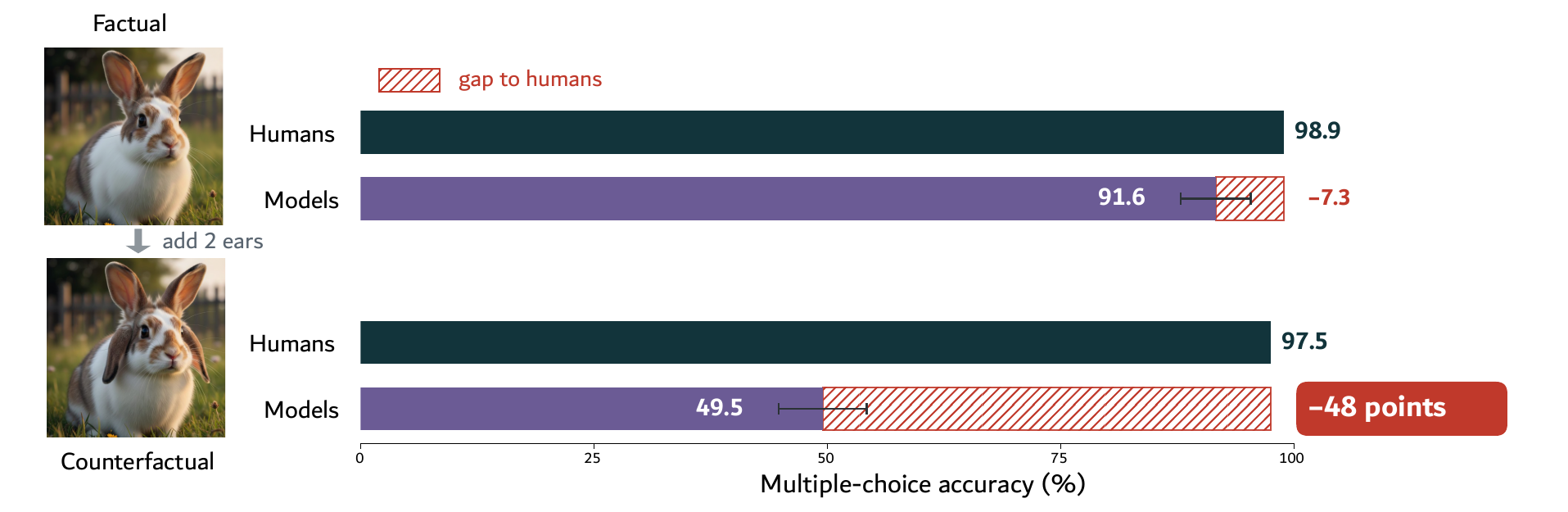}
    \caption{
    VLMs nearly match humans on factual images but collapse on counterfactual ones, while humans remain near ceiling, suggesting a reliance on knowledge priors over visual evidence.
    }
    \label{fig:intro_fig}
    \vspace{-4pt}
\end{figure}

To this end, we introduce \emph{CounterCount} (\Cref{sec:dataset}), a diagnostic dataset of paired factual and counterfactual (CF) images, where each pair targets an attribute with a strong canonical prior (\eg ears, wheels, or legs) at a small count.
A controlled human study confirms that the task is trivial: humans achieve near-perfect accuracy on both factual and CF images (98.9\% and 97.5\%; \Cref{fig:intro_fig}), showing that CF images pose no additional perceptual difficulty.
On the other hand, VLMs approach human accuracy on factual images (91.6\%), where the canonical count is also the correct answer, they collapse to 49.5\% on CF images, with over \emph{three quarters} of their CF errors returning exactly the canonical count.

To understand this failure, we conduct a systematic analysis that reveals two behaviors behind it (\Cref{sec:experiments}).
\textbf{First, models exhibit System-1-like behavior} (\Cref{sec:experiments:sys1}): the bias is triggered by a shallow recognition of the subject in the image rather than by how it is named in the prompt, and disturbing this recognition shortcut through blurring or cueing, or encouraging System-2-like reasoning~\citep{kahneman2011thinking} through thinking mode, moves predictions away from the canonical count.
\textbf{Second, models have the visual evidence but underuse it} (\Cref{sec:experiments:info}): failures coincide with under-attention to the attributes they are asked to count, and redirecting attention toward them helps models act on evidence they already have.
These findings suggest that recognizing the subject leads models to recall rather than count, even though the information required to count correctly is available to them.

In summary, our main contributions are:
\begin{itemize}
    \item \textbf{CounterCount.} A diagnostic dataset of 554 manually verified factual/CF image pairs across 379 subjects, 228 attribute types, and 17 categories, with localized oracle annotations (masks and bounding boxes) for fine-grained analysis of counting bias (\Cref{sec:dataset}).
    \item \textbf{Human and model evaluation.} An evaluation of humans and 16 VLMs (14 open-weight, 2 closed-source), showing that humans achieve near-ceiling accuracy on both factual and CF images while VLMs drop sharply on CF images, demonstrating that model failures are not due to task difficulty (\Cref{fig:intro_fig}).
    \item \textbf{Diagnostic analysis.} A systematic exploration yielding two key findings: (1) models exhibit System-1-like behavior, falling back on canonical counts based on knowledge after a shallow recognition of the subject, and (2) they overlook available visual evidence, attending less to count-relevant regions when they fail (\Cref{sec:experiments}).
\end{itemize}
\section{CounterCount Dataset}
\label{sec:dataset}

\begin{figure}
    \centering
    \includegraphics[width=1\linewidth]{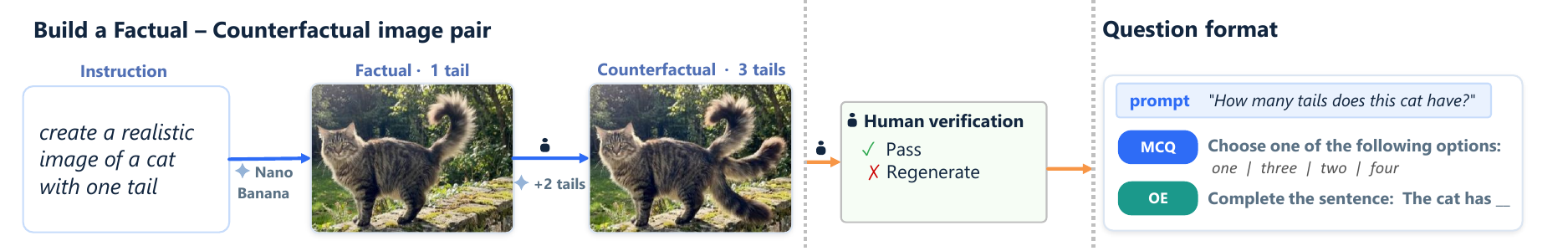}
    \caption{\textbf{CounterCount dataset construction and evaluation pipeline.} The pipeline follows four steps: (1) \textbf{Factual Image Generation}, where an instruction generates an image reflecting a canonical attribute, such as a cat with one tail; (2) \textbf{CF Image Generation}, where a numerical modifier (e.g., +2 tails) is applied to deliberately alter the canonical attribute and produce an edited counterpart; (3) \textbf{Human Verification}, where the image pair is reviewed to confirm the counts, resulting in either a pass or a command to regenerate; (4) \textbf{Question Formats}, Open-Ended (OE), and Multiple Choice (MCQ)}
    \label{fig:dataset_Pip}
    \vspace{-5pt}
\end{figure}

To test whether VLMs ground counting predictions in visual evidence or rely on canonical object priors, we construct paired factual and CF images.
An image pair differs only in a localized count-relevant edit, preserving viewpoint, background, lighting, and global composition while altering the number of a structural attribute (\eg ears, legs, wheels).
The edited regions are fully visible and manually verified, so the target count is explicit and unambiguous. This minimal design helps attribute performance differences between factual and CF images to conflicts between visual evidence and object-level priors rather than incidental scene variation.
To support analysis beyond answer accuracy, \emph{CounterCount} also provides localized masks and bounding boxes (BBs) for the modified count-relevant regions, enabling region-level analysis and controlled diagnosis during inference.

%%%%%%%%%%%%%%%%%%%%%%%%%%%%%%%%%%%%%%%%%%%%%%%%%%%%%%%%%%%%%%%%%%%%%%%%%%%%%

\subsection{Dataset Construction}

We chose common, widely recognized subjects to ensure broad coverage across familiar attributes, starting from broad groups (\eg, animals) and expanding into finer categories (\eg, mammals, birds). \emph{CounterCount} spans seventeen semantic categories across biological, manufactured, and functional domains, containing 379 unique subjects (\eg, dog, airplane, Starbucks logo) and 228 countable attribute types (e.g., leg, wing, crown), capturing a diverse set of canonical object priors.
The full category list and per-category breakdown are provided in \Cref{sup:dataset_Sta}.

\myparagraph{Attribute Selection and Canonical Priors.}
For each subject, we select attributes with widely recognized canonical counts and edit them to create counts that directly conflict with the expected object structure, such as a giraffe with five heads or a car with six wheels.
This places visual evidence in conflict with learned object priors, so that edited instances cannot be answered from prior knowledge alone; instead, models must ground their predictions in the visible count-relevant evidence.

\myparagraph{Image Generation and Quality Control.}
% Both images in every pair are fully synthetic. 
We generate a photorealistic factual image with canonical counts using
Nano-Banana \footnote{\url{https://deepmind.google/models/gemini-image/}}. 
Then, we prompt the same model for a CF variant that alters one count-relevant attribute while preserving viewpoint, lighting, background, and object identity. Each image pair is manually verified for photorealistic quality, semantic coherence, and visibility of the edited attribute; pairs that fail verification are regenerated until all criteria are met. 
This yields 554 CF images paired
with their factual counterparts (1{,}003 unique images in total since a single factual image can serve as the base for multiple CF edits targeting different attributes), all at a standardized
$480 \times 480$ resolution. 
Further pipeline details and examples are provided in
\Cref{fig:dataset_Pip} and \Cref{sup:pairs}.

 \myparagraph{Question Format.}
We evaluate each image pair by asking models how many instances of the target attribute the subject has.
All questions are generated from fixed templates filled with each image's manually annotated metadata (subject, attribute, factual count, and CF count).
We exclude binary yes/no questions, since VLMs show a systematic tendency to answer ``yes''~\citep{chan2026system}, which would confound our analysis.
We therefore use two formats:
(i)~\textit{four-option Multiple-Choice Questions (MCQ)}, where distractors are balanced around both the canonical and CF counts, and the option order is randomized to reduce positional bias~\citep{pezeshkpour2025benchmarking, zeno2025choosing}; and
(ii)~\textit{Open-Ended (OE) completion}, where the model completes a sentence with only the count and part name.
Templates and full construction details are provided in \Cref{sub:template}.

\myparagraph{Human Study.}
\label{sec2:human}We assess human performance on CounterCount to see how challenging the task is for humans, using human annotators recruited through Rapidata\footnote{\url{https://www.rapidata.ai/}}, a high-throughput platform for collecting human responses at scale.
We use the MCQ format to minimize ambiguity and reduce free-form response errors. Each image is independently annotated by 25 qualified annotators, and since each image is displayed for a few seconds, this encourages quick perceptual judgments rather than deliberate analysis. The final answer is determined by majority voting.
Full details of the human study are provided in Appendix~\ref{sup:human_study}.

%%%%%%%%%%%%%%%%%%%%%%%%%%%%%%%%%%%%%%%%%%%%%%%%%%%%%%%%%%%%%%%%%%%%%%%%%%%%%%%%%%%%
\subsection{Region-Level Annotations}
 \begin{figure}[htpb]
    \centering
    \includegraphics[width=.75\linewidth]{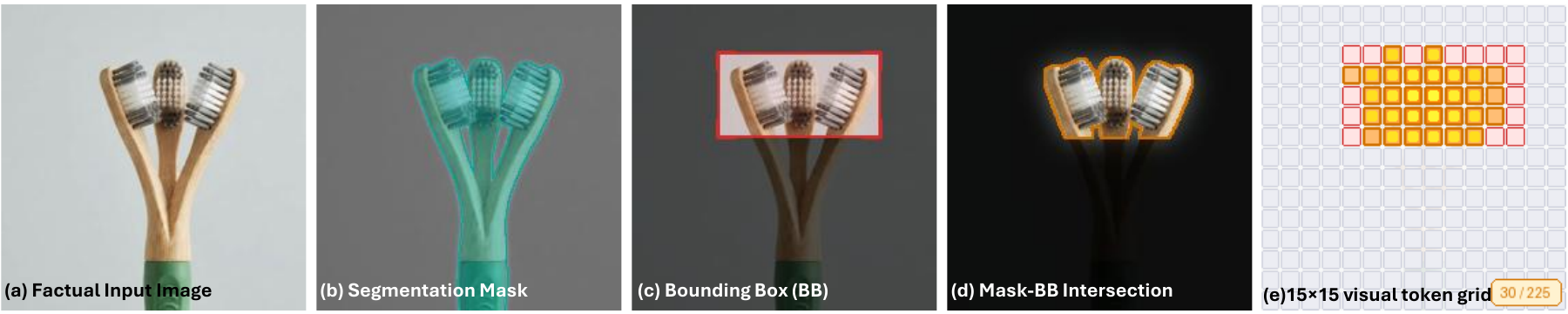}
    \caption{
    \textbf{Region-level annotation pipeline.}
}
    \label{fig:annotation}
    % \vspace{-11pt}
\end{figure}

In addition to paired factual-CF images, \emph{CounterCount} provides region-level annotations for each edited instance, including segmentation masks and bounding boxes (BB).
As shown in \Cref{fig:annotation}, these annotations localize the edited object and the specific count-relevant attribute.
Starting from the CF image (a), we use SAM~\citep{carion2026sam3} to obtain a segmentation mask for the full object containing the edit (b).
We then manually annotate a BB around the modified attribute (c).
The object mask captures the broader visual context, while the BB focuses on the localized structural change.
Their intersection, denoted Mask-BB, isolates the pixels corresponding to the edited count-relevant region (d).
We map the mask, BB, and Mask-BB regions onto the fixed visual-token grid of the vision encoder (e).
This region-to-token alignment identifies both object-level tokens and localized anomaly tokens, enabling analyses and diagnosis targeted at the visual evidence required for counting rather than the image as a whole.
\section{CounterCount Evaluation and Analysis}
\label{sec:results}
We evaluate VLMs on CounterCount's 554 image pairs using two question formats: MCQ and OE.
We report MCQ results in the main paper, as this matches the format of our human study, and provide OE results in \Cref{sup:oe_res}.

\myparagraph{Experimental Setup.}
We evaluate 16 VLMs: 14 open-weight models from six families, namely Qwen3-VL (4B, 8B, 32B)~\citep{bai2025qwen3vl}, Qwen3.5 (4B, 9B, 27B)~\citep{qwen35}, Qwen3.6 (27B)~\citep{qwen36_27b}, Qwen3.8 (27B)~\citep{qwen38repo}, Gemma-3 (4B, 12B, 27B)~\citep{gemma2025gemma3}, and Gemma-4 (E2B, 12B, 31B)~\citep{gemma4}, and two closed-source models, GPT-5.6 Luna~\citep{gpt56} and Sonnet 5~\citep{sonnet5}.
Open-weight models run locally on a single NVIDIA A100 80GB GPU with greedy decoding, while closed-source models are accessed through their APIs.
Unless otherwise stated, we disable reasoning and limit outputs to 128 tokens for all models.
For clarity, figures in the main paper report six representative models: the two closed-source models, the two strongest open-weight models (Gemma-4-31B and Qwen3.8-27B), and a medium (Gemma-3-12B) and a small (Qwen3.5-4B) model to cover different scales, with results for all models in \Cref{sup:experiments}.
Analyses that require model internals or thinking mode use a subset of models, specified in the corresponding sections.

\myparagraph{Evaluation Metrics.}
We extract numeric answers from model responses using regular expressions.
When a response contains multiple numbers, we prompt the same VLM in its baseline configuration to identify the intended answer.
On factual images, the canonical count is the correct answer, so we report accuracy only.
On CF images, we classify each prediction as \emph{correct} if it matches the edited count, \emph{biased} if it matches the canonical count, and \emph{other error} otherwise.
We report \textbf{\emph{accuracy}} as the fraction of correct predictions and \textbf{\emph{bias}} as the fraction of biased predictions, which measures how often models favor the canonical count over the image content.

\subsection{Baseline Results}
\begin{figure}[htpb]
    \centering
    \begin{subfigure}[b]{0.48\linewidth}
        \centering
        \includegraphics[width=\linewidth]{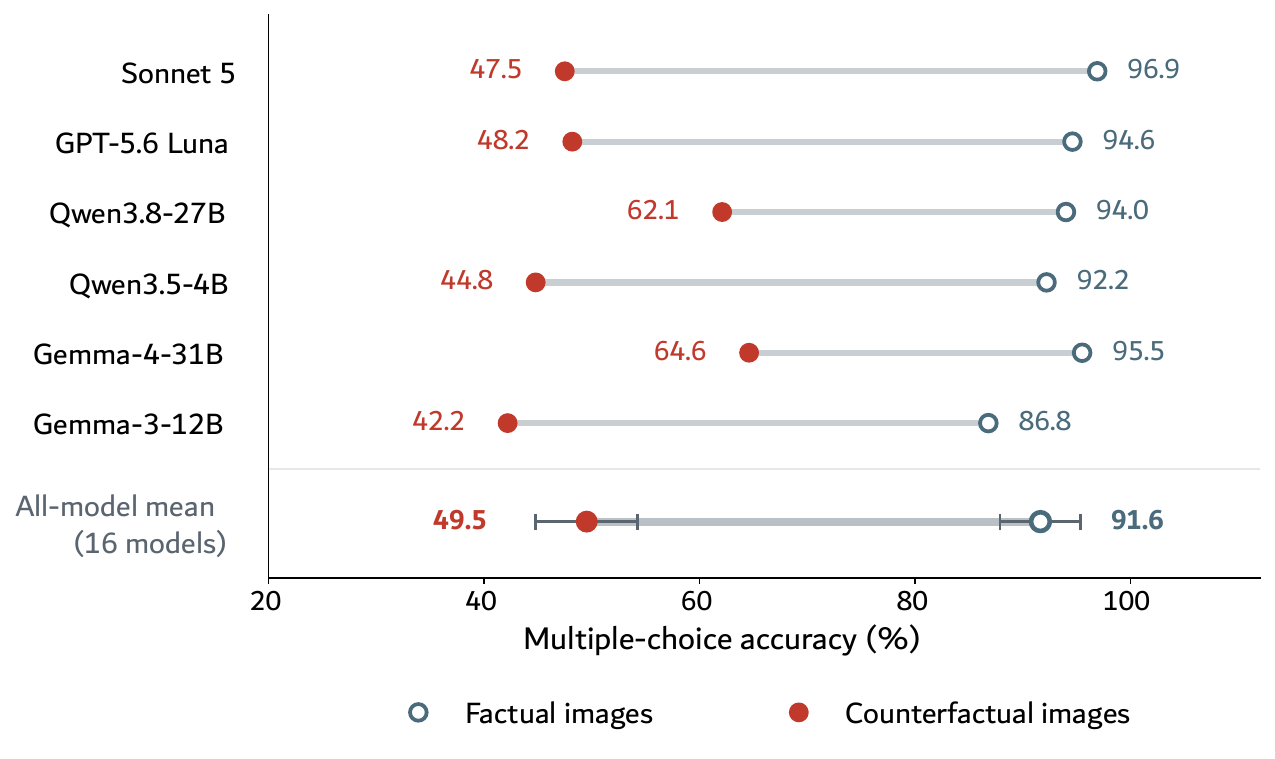}
        \caption{MCQ accuracy on factual and CF images.}
        \label{fig:res_a}
    \end{subfigure}
    \hfill
    \begin{subfigure}[b]{0.5\linewidth}
        \centering
        \includegraphics[width=\linewidth, height=4cm]{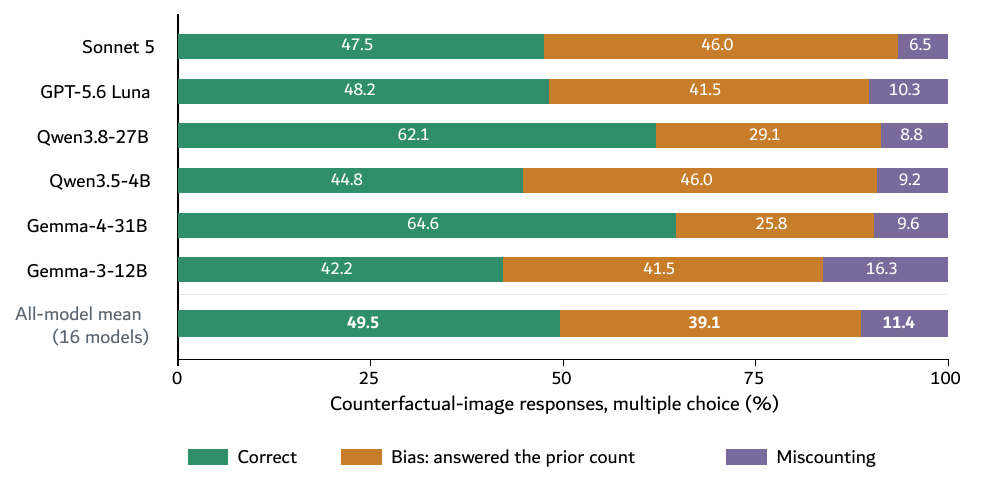}
        \caption{CF-image predictions split.}
        \label{fig:res_b}
    \end{subfigure}

    \caption{
    \textbf{Models' performance on CounterCount.}
}
    \label{fig:baseline}
    % \vspace{-5pt}
\end{figure}

\Cref{fig:res_a} presents MCQ results for the six selected models, along with the mean across all 16 models with 95\% confidence intervals.
Every model performs worse on CF images, with mean accuracy dropping from 91.6\% on factual images to 49.5\% on CF images, a 42.1-point drop.
In contrast, humans remain near ceiling on both (98.9\% and 97.5\%; \Cref{fig:intro_fig}), leaving a 48-point gap on CF images.
\setlength{\columnsep}{5pt}
\setlength{\intextsep}{3pt}

\begin{wrapfigure}[]{r}{0.38\textwidth}
    \centering
    % \vspace{-0.5em}
    \includegraphics[width=0.8\linewidth]{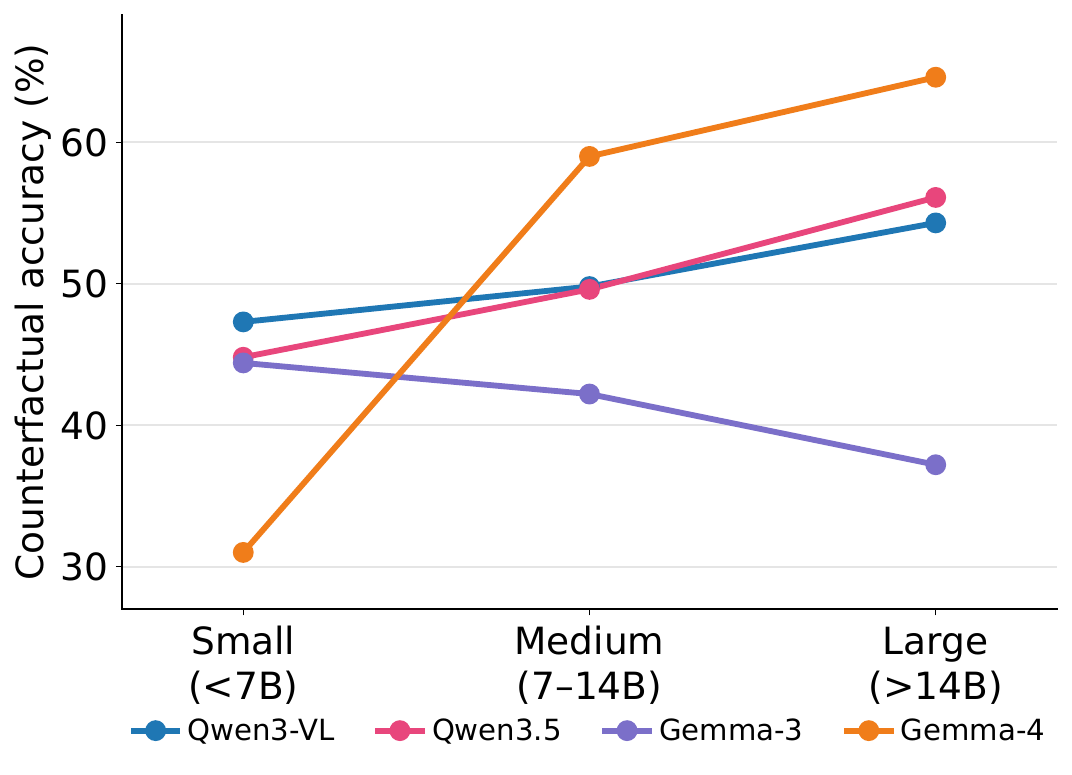}
    \par\vspace{4pt}
    \includegraphics[width=0.8\linewidth]{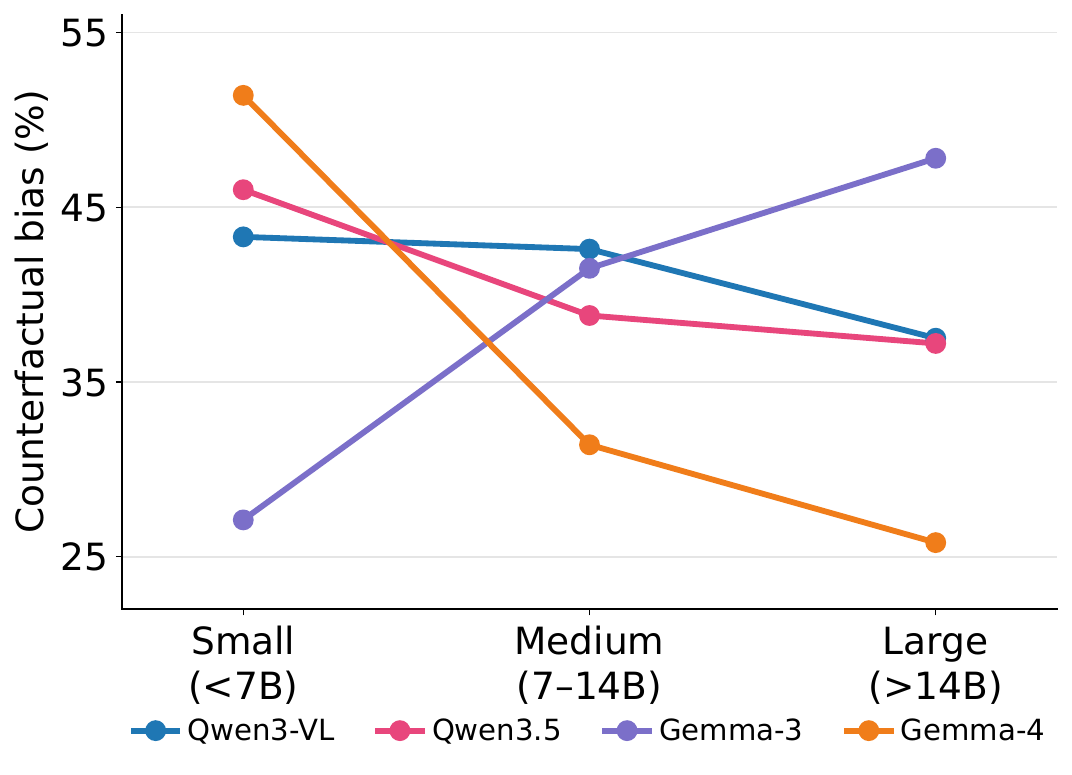}
    \caption{\textbf{Effect of model size.}
}
    \label{fig:scaling}
\end{wrapfigure}
Since the canonical count is also the correct answer on factual images, factual accuracy alone cannot reveal whether models count from the image or recall from prior knowledge.
For example, Sonnet 5 achieves the highest factual accuracy (96.9\%) but reaches only 47.5\% on CF images, below the open-weight Gemma-4-31B (64.6\%) and Qwen3.8-27B (62.1\%).

Most CF errors are not random miscounts.
As shown in \Cref{fig:res_b}, models return the canonical count in 39.1\% of CF predictions, compared with only 11.4\% for other errors, meaning that 77\% of errors default to the canonical count.
If failures stemmed from limited counting ability, errors would be spread across different counts rather than concentrated on the canonical one.
This indicates that models mostly favor prior knowledge over visual evidence, rather than simply failing to count.
OE results show the same trend, with mean accuracy falling from 90.9\% on factual to 46.8\% on CF images and a bias rate of 39.3\%.
Full MCQ and OE results are in \Cref{sup:mcq_res,sup:oe_res}.

\myparagraph{Does scaling improve performance?}
Within open-weight families, CF accuracy consistently increases and bias decreases with model size (\Cref{fig:scaling}), with Gemma-3 as the exception, where accuracy declines and bias rises.
However, even the best model remains far below its factual accuracy, and both closed-source models, despite their presumably larger scale, fall below the best open-weight models, suggesting that scale alone does not resolve the bias.

% As shown in \Cref{fig:res_b}, the fact that models return canonical counts in 39.1\% of edited-image predictions, compared with only 11.4\% for other miscounts, suggests that they often favor prior knowledge over visual evidence. OE results show the same pattern, with mean accuracy falling from 90.9\% to 46.8\%.Additional MCQ and full OE results appear in \Cref{sup:mcq_res,sup:oe_res}, respectively.\vspace{4pt}
% \hfill\break\indent\myparagraph{Does scaling improve performance?} On average, edited-image accuracy increases for large models, while bias decreases (\Cref{fig:scaling}), though accuracy remains well below that on original images. Gemma-4 and Gemma-3 show opposite trends, with performance improving for the former and declining for the latter as size increases. Both Qwen families show more consistent improvements.

\subsection{CounterCount Results Analysis}

\myparagraph{Are all categories equally difficult?}
Some categories are much harder than others (\Cref{sup:cat_res}).
Bugs and mythical creatures have the lowest CF accuracy (20\% and 29\%), while housing reaches 82\%, followed by flags and food (78\%).
Lower accuracy coincides with higher bias, as models return the canonical count for 73\% of CF bug images, compared with only 9\% for housing.
Symbols stand out with the lowest factual accuracy (66\%) and a smaller drop on CF images (54\%), suggesting that models struggle with this category even before any modifications.

\begin{figure}[htbp]
    \centering
    \begin{subfigure}[b]{0.48\linewidth}
        \centering
        \includegraphics[width=\linewidth]{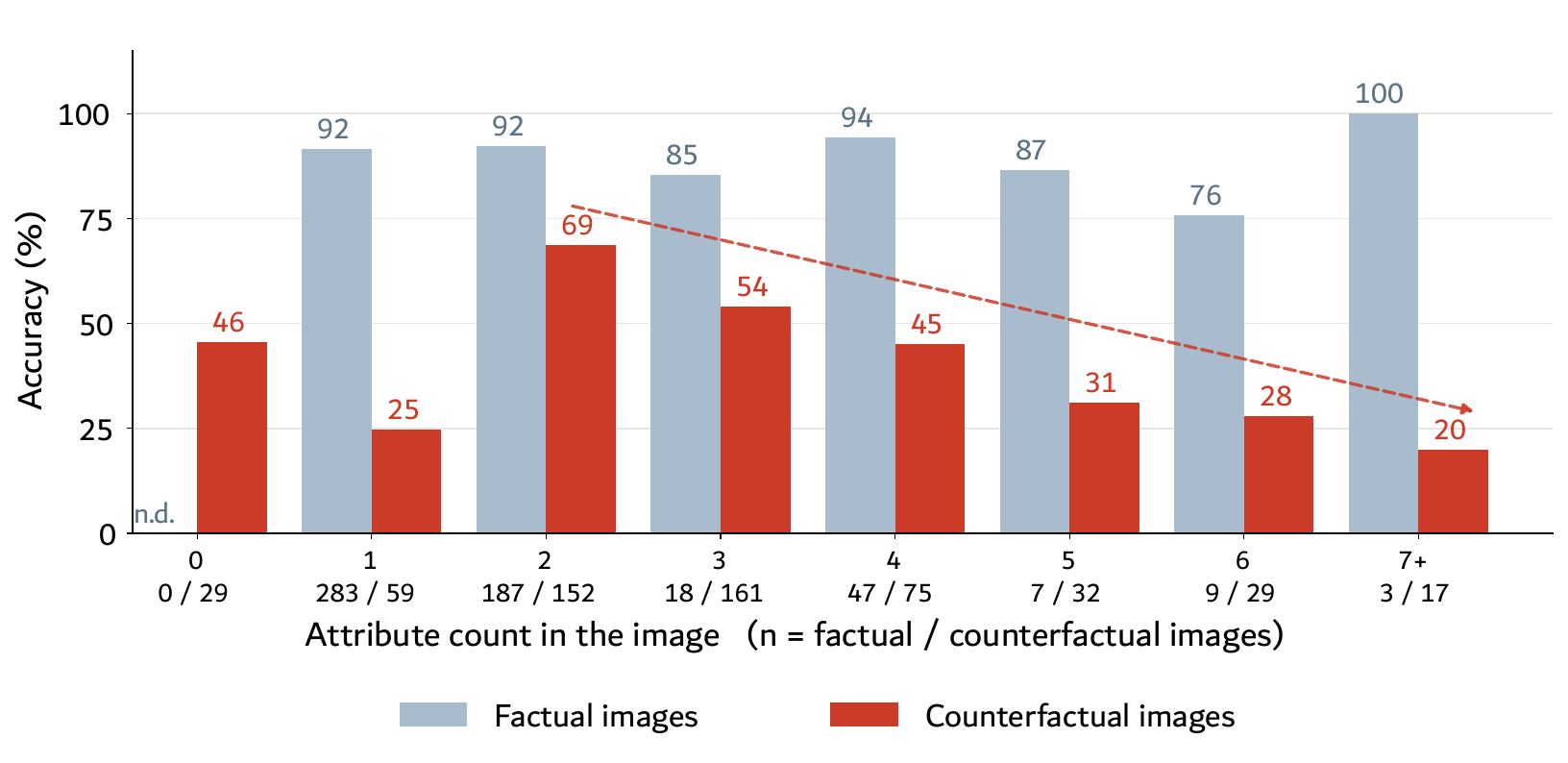}
        \caption{}
        \label{fig:count_a}
    \end{subfigure}
    \hfill
    \begin{subfigure}[b]{0.48\linewidth}
        \centering
        \includegraphics[width=\linewidth]{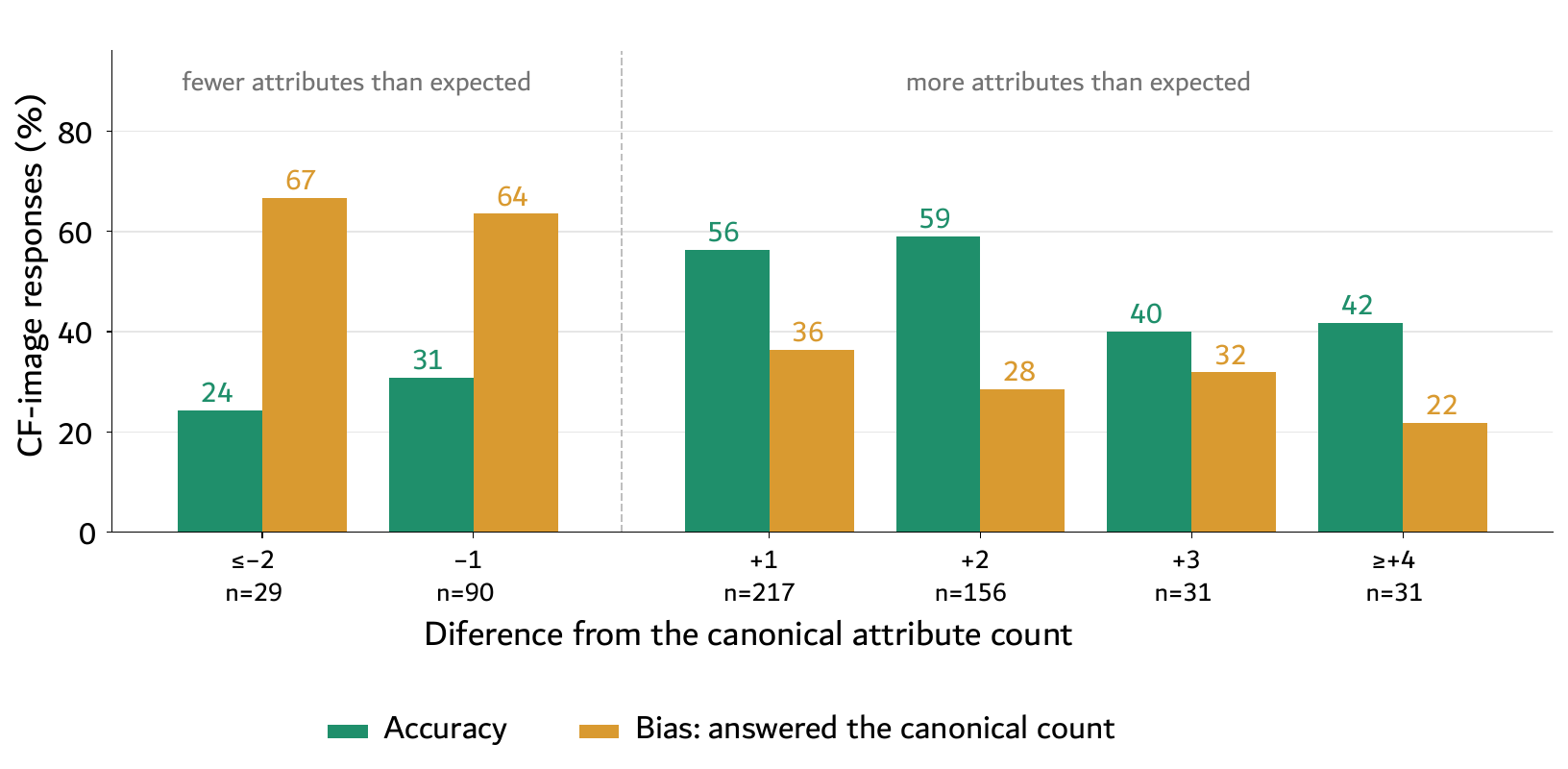}
        \caption{}
        \label{fig:count_b}
    \end{subfigure}
    \caption{\textbf{Effect of attribute count.} (a) MCQ accuracy on factual and CF images by attribute count. (b) CF accuracy and bias rate by the difference from the canonical count, with negative values indicating removal and positive values indicating addition.}
    \label{fig:count_analysis}
    \vspace{-6pt}
\end{figure}

\myparagraph{Does the attribute count matter?}
Consistent with prior work showing that VLM counting degrades as counts increase~\citep{sengupta2025can, qharabagh2025lvlmcount, alghisi2025dereconstructing}, CF accuracy for counts of two or more falls from 69\% at two to 20\% at seven or more (\Cref{fig:count_a}), confirming our choice to keep most counts small.
Factual accuracy, in contrast, varies across counts (76--100\%) without a declining trend, since the canonical count yields the correct answer regardless of how many attributes there are.
Counts of zero and one are exceptions, with low accuracy despite their small size.
These cases mostly involve attribute removal, where models are far more likely to default to the canonical count. The bias rate reaches 64--67\% for removal (\eg $4$ legs $\rightarrow$ $3$ legs), compared with 22--36\% for addition (\eg $1$ tail $\rightarrow$ $3$ tails; \Cref{fig:count_b}).
This suggests that missing attributes are easier to overlook than added ones, leading models to fill the gap with prior knowledge.

% \vspace{-10pt}

% \begin{itemize}
%   \item Baseline Results + Comparison with Human Performance
%   \item Scaling Laws 
%   \item Are All Categories Equally Difficult?
%   \item Attribute Count Analysis
%   \begin{itemize}
%     \item Does performance vary with attribute count?
%     \item Overcount Vs. Undercount
%   \end{itemize}
% \end{itemize}

%\item Is Dataset Size Enough for Diagnosis? % put it in limitations

\section{Understanding model behavior on counterfactuals}
\label{sec:experiments}

Models show a surprising gap in counting capabilities between factual and CF images. In this section, we study this behavior more closely, showing that superimposing a System-1/2 \citep{kahneman2011thinking} model partially correlates with model bias (\Cref{sec:experiments:sys1}), as well as the partial presence of necessary information in the activation space (\Cref{sec:experiments:info}).

\subsection{Models exhibit System 1-like answer behavior}\label{sec:experiments:sys1}

We hypothesize that models exhibit System-1-like thinking \citep{kahneman2011thinking} when answering questions, where they take a glance at the input image, recognize the object, and ultimately fall back on knowledge priors when answering questions about attributes. We study this hypothesis from four angles below. 

\subsubsection{Knowledge priors are caused by visual processing}
\label{sec:experiments:neutral}

A natural question is whether the explicit naming of the subject prompts models to return its canonical count purely based on knowledge. We ablate this by replacing subject names with generic phrasing (e.g., ``parrot'' $\rightarrow$ ``animal''), rerunning our experimental validation (see~\Cref{sup:phrasing} for details), and find almost no change in accuracy or bias. This suggests that the source of the knowledge is not originating from the language alone, but must instead originate from the vision side, which we investigate in the following section.

\begin{figure}[htbp]
    \centering
    \includegraphics[width=0.8\linewidth]{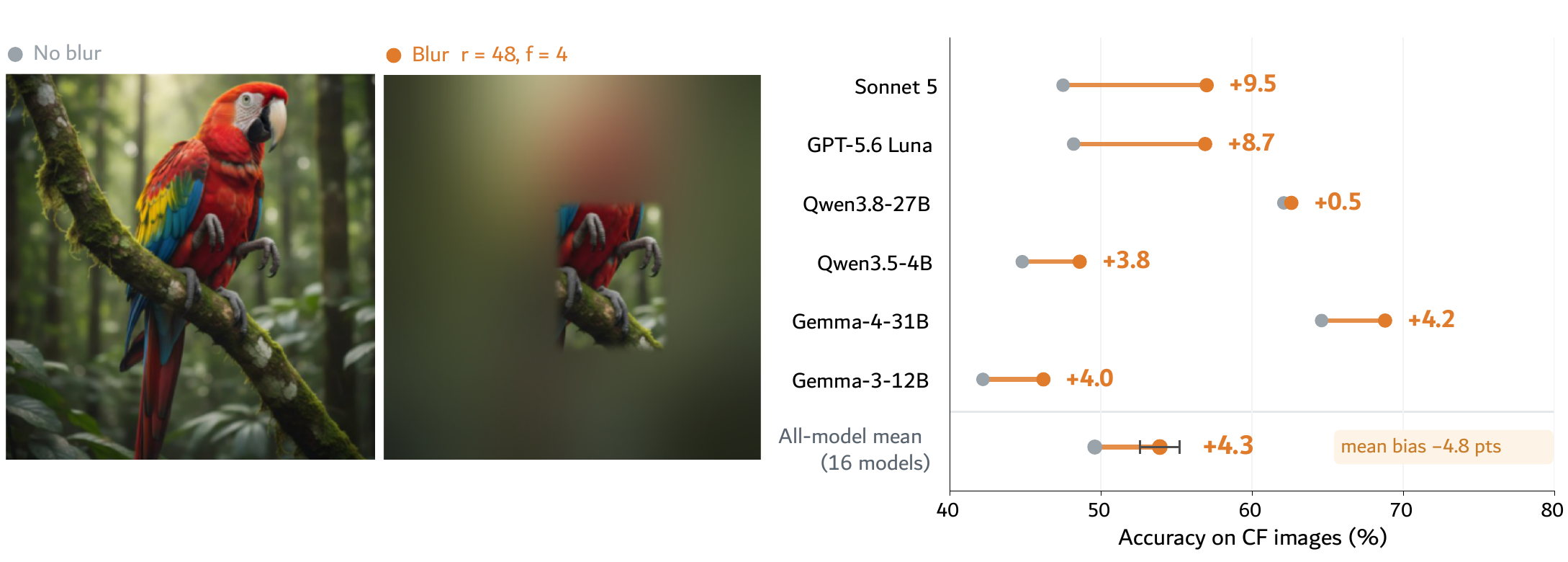}
    \caption{\textbf{Effect of visual context degradation.} Example image before and after blurring (left), alongside changes in CF-image MCQ accuracy (right).}
    \label{fig:blur_fig}
    \vspace{-5pt}
\end{figure}

% We test whether models recognize the subject and default to its expected count without checking the image closely. We investigate this through blurring, context about unusual counts, and explicit reasoning.

\subsubsection{Degrading visual context by blurring}
Models may recognize an object's overall shape and use its canonical count as a shortcut.
Inspired by~\citep{Yang2023visualprompting}, we test whether degrading the visual context reduces this reliance.
Using CounterCount's BB annotations, we keep the counted attribute regions sharp and blur everything else, including the rest of the subject and the background, so that the attributes remain countable while the subject becomes harder to recognize.
We use a blur radius of 48 pixels and feathering of 4 pixels, selected through a parameter sweep on Qwen3-VL-8B (\Cref{sup:blur}).
As shown in \Cref{fig:blur_fig}, blurring improves mean CF MCQ accuracy by 4.3 points and reduces bias by 4.8 points across 16 models, with similar results in OE (\Cref{sup:blur}).
These gains suggest that recognizing the subject is at least partially responsible for the bias.
However, blurring cannot fully prevent recognition, since the attributes themselves often reveal the subject (\eg the shape of a leg or feather suggests a bird).
Removing more visual information would make the image ambiguous, for example, making it unclear whether the attributes belong to one subject or several.

\subsubsection{Cueing the model through instructions}
\label{sec:cueing}
\setlength{\columnsep}{5pt}
\setlength{\intextsep}{2pt}
\begin{wrapfigure}[12]{r}{0.6\textwidth}
    % \vspace{-2em}
    \centering
    \includegraphics[width=\linewidth]{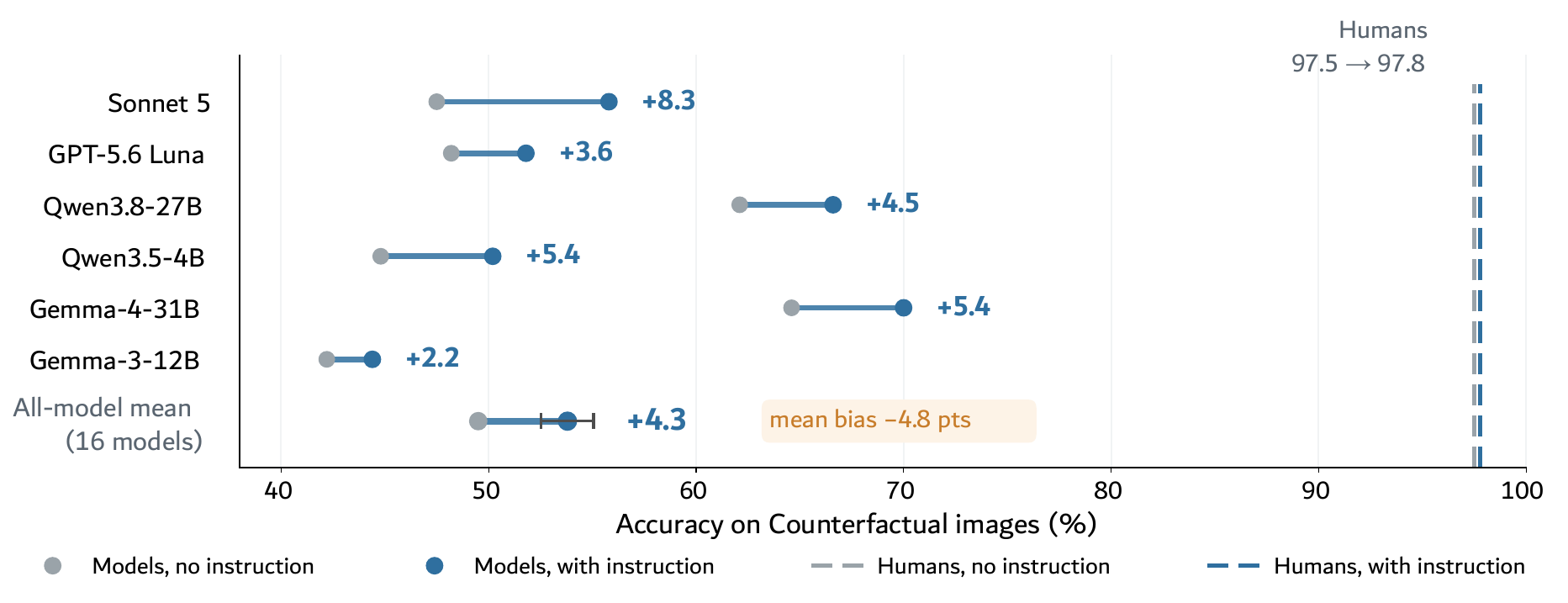}
    \caption{\textbf{Effect of cueing.} MCQ accuracy on CF images with and without the instruction, compared with humans.}
    \label{fig:inst_fig}
\end{wrapfigure}
Human observers typically benefit from \textit{cueing} \citep{kahneman2011thinking}, where they receive a hint that the observed stimuli might be special. We therefore wonder if this applies to models as well, causing them to more carefully examine the visual information. 
We provide them with this instruction: \textit{``Some images are digitally altered, so counts may be unrealistic. Count only what you actually see.''}. 

On factual images, the instruction has little effect for either models or humans, as expected (\Cref{sup:instruction:humanvsmodels}).
On CF images, however, the two diverge.
Models benefit from the cue, with mean MCQ accuracy increasing by 4.3 points and bias decreasing by 4.8 points, suggesting that giving instructions helps them rely more on visual evidence.
Yet their mean accuracy reaches only 53.8\%, still far below humans, whose accuracy remains nearly unchanged (97.5\% without and 97.8\% with the instruction; \Cref{fig:inst_fig}).
Since the task is simple and unambiguous, humans already count from the image and do not need the cue, whereas models improve only partially even when explicitly told that counts may be unusual.
For the human evaluation, we follow the protocol of \Cref{sec2:human} with a new annotator pool validated with the instruction included, for consistency.
Full OE and remaining MCQ results are provided in~\Cref{sup:instruction}.

 %Models benefit from the instruction, with mean MCQ accuracy increasing by 4.3 percentage points and bias decreasing by 4.8 points. 
%  Cueing models therefore reduce bias and cause a more careful observation of visual information, albeit still far below human performance (\Cref{fig:inst_fig}), with a mean accuracy of only 53.8\%. Full OE and remaining MCQ results, along with details of the human experiments, are provided in~\Cref{sup:instruction}.

% We also give the same instruction to humans, using 25 annotators per image with majority voting, selected through the same validation procedure as in our main study but with the instruction included.
% As expected, the instruction has little effect on factual images for either humans or models.
% On CF images, human accuracy remains nearly unchanged (97.5\% without and 97.8\% with the instruction; see \Cref{fig:inst_fig}). We attribute this to the baseline already being near ceiling.

\subsubsection{Thinking modes}

\setlength{\columnsep}{5pt}
\setlength{\intextsep}{4pt}
\begin{wrapfigure}[]{r}{0.5\textwidth}
    % \vspace{-3em}
    \centering
    \includegraphics[width=0.8\linewidth]{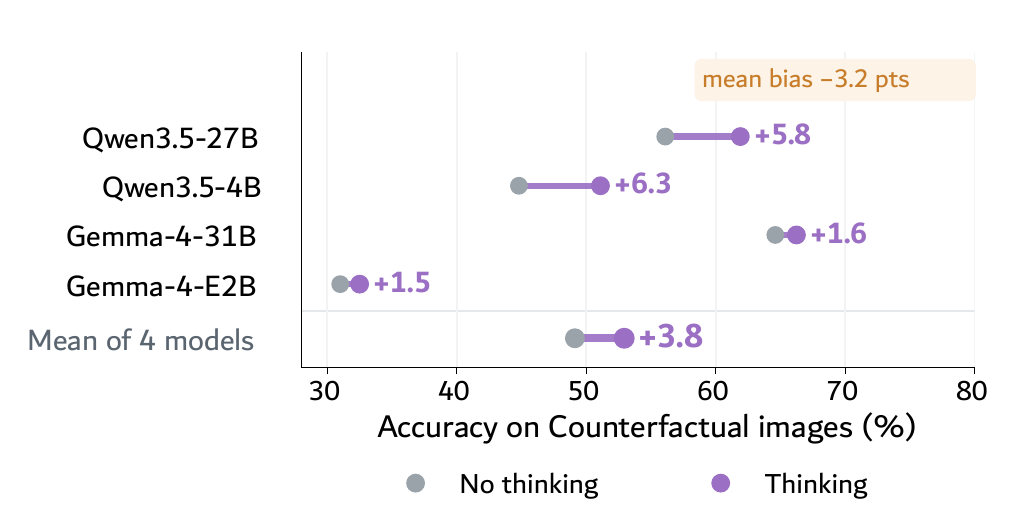}
    \caption{\textbf{Results with thinking.}}
    \label{fig:think_fig}
\end{wrapfigure}
What if models reason before answering?
This most closely resembles System-2 processing~\citep{kahneman2011thinking}.
We enable thinking mode with each model's default sampling parameters and a maximum budget of 16,384 tokens for two model families that support it, Qwen3.5 (4B, 27B) and Gemma-4 (E2B, 31B), selecting a small and a large model from each to examine how the effect varies with size.
As shown in \Cref{fig:think_fig}, thinking improves mean CF accuracy by 3.8 points in MCQ and 4.6 in OE (\Cref{fig:think_oe}), while reducing bias by 3.2 and 4.1 points, respectively. See \Cref{sup:thinking} for details.

\begin{tcolorbox}[
    colback=gray!5,
    colframe=gray!60,
    boxrule=0.5pt,
    arc=2pt,
    left=6pt, right=6pt,
    top=5pt, bottom=5pt
]
\textbf{Takeaway.}
Models fall back to canonical counts upon a shallow recognition of the object, not semantic cues in the prompt. Encouraging System-2-like thinking by breaking visual context, cueing, or thinking mode reduces but does not fully alleviate bias.
\end{tcolorbox}

\subsection{Models have the information internally but under-use it}\label{sec:experiments:info}
\begin{figure}[htbp]
    \centering
    \includegraphics[width=0.9\linewidth]{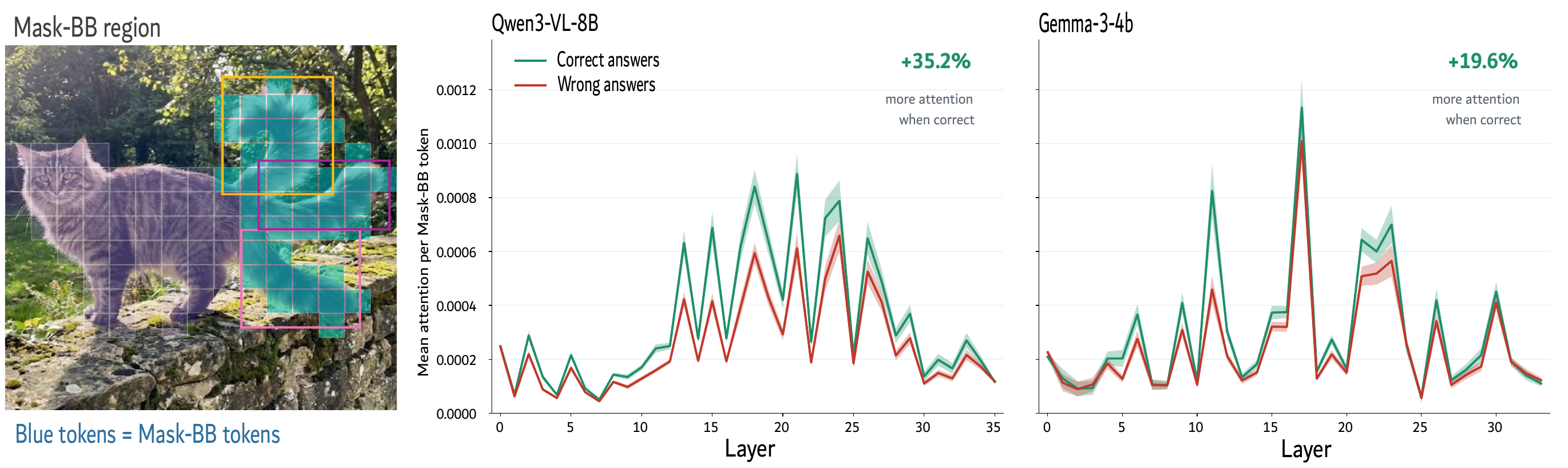}
    \caption{\textbf{Attention to count-relevant regions.} Mean attention per Mask-BB token across layers for correct and incorrect MCQ answers on CF images.}
    \label{fig:corwro_fig}
\end{figure}
Beyond changing the input or the prompt, we ask whether models already have access to the visual evidence but fail to use it. Do models give enough attention to the relevant visual information? For each CF image, we measure the attention from the last token, which predicts the answer, to the Mask-BB tokens at each layer, averaged over heads and normalized by the number of selected tokens, and compare correct and incorrect answers. As shown in \Cref{fig:corwro_fig}, averaged across layers, correct answers receive 35.2\% more attention in Qwen3-VL-8B and 19.6\% more in Gemma-3-4B. Similar patterns appear in OE (\Cref{sup:attention}). This suggests that models may give insufficient attention to count-relevant regions when they answer incorrectly. We therefore test whether boosting attention to these regions can help recover correct answers.

\setlength{\columnsep}{5pt}
\setlength{\intextsep}{2pt}
\begin{wrapfigure}{r}{0.5\textwidth}
    \centering
    \includegraphics[width=\linewidth]{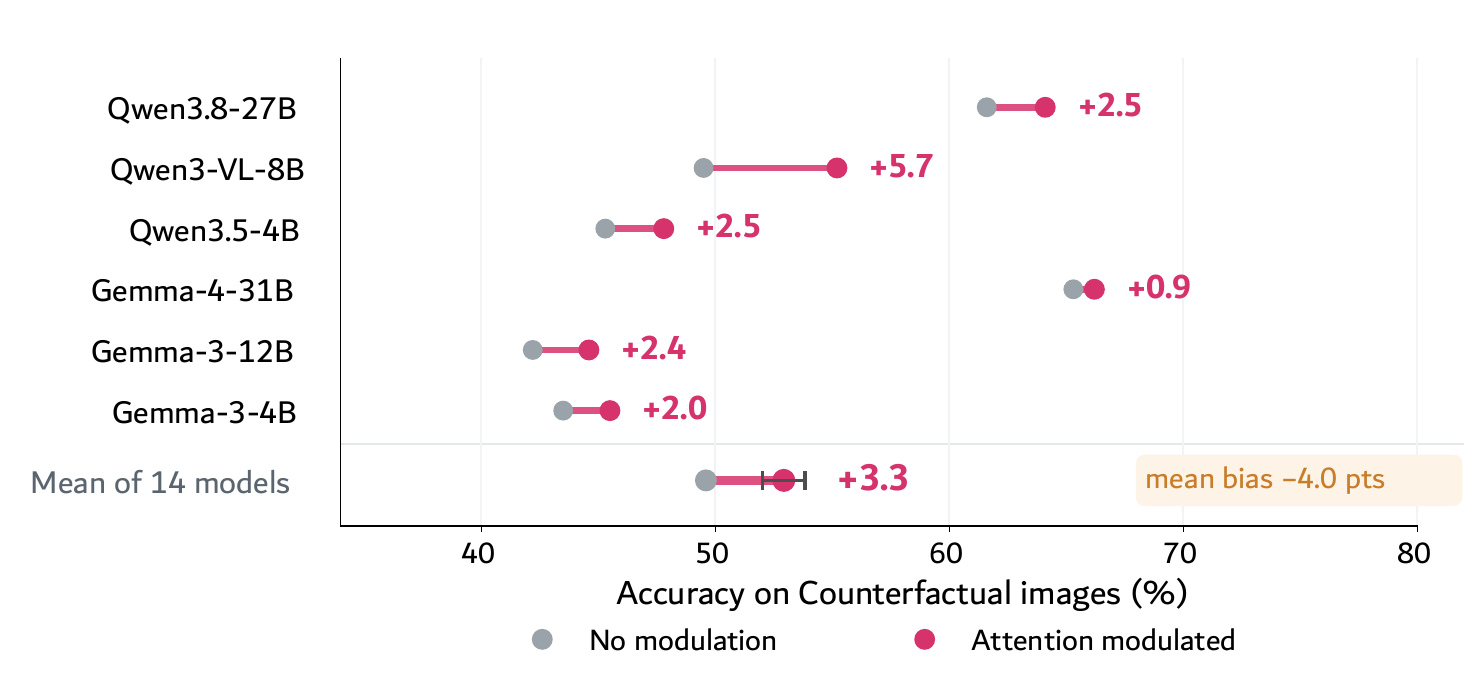}
    \caption{\textbf{Attention modulation results.}}
    \label{fig:att_fig}
\end{wrapfigure}
\textbf{Attention Modulation.}
Following prior work on hallucination in VLMs that modifies attention to visual tokens~\citep{liu2024paying, zhao2025tell, Jiang2025devils, yu2025causally, Tu2025attention}, we use a simple, unified formulation that amplifies, suppresses, or masks attention on selected heads through additive adjustments to attention logits. Let $z_{h,i,j}$ denote the attention logit from query position $i$ to key position $j$ in head $h$. For target visual tokens $T$ and the remaining visual tokens $B$ outside the target region, we apply
\begin{equation}
\tilde{z}_{h,i,j} =
\begin{cases}
z_{h,i,j} + \log(\alpha), & j \in T\\
z_{h,i,j} + \log(\beta), & j \in B
\end{cases}
\label{eq:logit_shift}
\end{equation}
where $\alpha \geq 1$ amplifies target attention and $0 \leq \beta \leq 1$ controls background suppression. A value of one leaves the logit unchanged, while zero masks the token by setting its logit to $-\infty$. We use Mask-BB tokens as the target and test two modes: target amplification alone ($\beta=1$) and target amplification with background masking ($\beta=0$). We rank attention heads using AUROC, computed as the fraction of token pairs in which a visual token inside the BB receives more attention than one outside it. Higher scores indicate stronger attention to the annotated region. We test 12 configurations combining these modes, $\alpha \in \{2,3\}$, and the top 5\%, 7\%, or 10\% of heads, selecting the best-performing configuration for each model as an upper-bound diagnostic. Full methodological and experimental details are provided in~\Cref{sup:attention}.

As shown in \Cref{fig:att_fig}, attention modulation improves mean CF MCQ accuracy by 3.3 points and reduces bias by 4.0 points across 14 open-weight models, with similar gains in OE (\Cref{sup:attention:oe}).
Since modulation only reweights visual tokens the model already receives, this indicates that the evidence needed to count correctly is partially available but underused.

\begin{tcolorbox}[
    colback=gray!5,
    colframe=gray!60,
    boxrule=0.5pt,
    arc=2pt,
    left=6pt, right=6pt,
    top=5pt, bottom=5pt
]
\textbf{Takeaway.}
Models under-attend to count-relevant regions when they fail.
Redirecting attention toward these regions, without adding any information, partially recovers correct answers, indicating that the visual evidence is available but underused in favor of prior knowledge.
\end{tcolorbox}
\section{Related Work}
\label{sec:related}
\myparagraph{Language Priors and Bias in VLMs.} Large Language Models (LLMs) exhibit systematic biases across demographic, cultural, and semantic dimensions~\citep{sheng2019woman, vo2025bscore, parrish2022bbq, wang2024countries, naous2024beer, shin2024ask}. Because Vision-Language Models (VLMs) rely on pretrained LLM backbones, these priors transfer to multimodal reasoning as spurious or stereotypical correlations~\citep{hall2023visogender, raj2024biasdora, Howard2024Social, yang2025escaping, ullman2024illusion}. Consequently, language priors can dominate visual evidence, leading models to produce prior-consistent answers even when the image supports a different response~\citep{yang2025escaping, ullman2024illusion, Deng2025Words, lee2025vlind, vo2025vision, fu2025hidden, golovanevsky2025pixels, liu2025seeing}. Recent analyses suggest relevant visual information is often encoded in cross-modal representations but attenuated during generation by the language backbone~\citep{Tong2024Eyes, zhao2025looking, vo2025vision2, liu2024paying, Favero2024multi, li2026seeing}. This creates a modality imbalance: visual evidence is insufficiently used when it conflicts with linguistic expectations. We study this phenomenon through counting, placing canonical object knowledge (e.g., typical numbers of ears or wheels) in direct conflict with localized visual evidence.

\myparagraph{Counterfactual Images for Diagnosing Bias.} Counterfactual images test whether VLM predictions are grounded in vision or memorized priors~\citep{lee2025vlind, wu2025lanp, ortu2025when, Luo2025probing}. Prior benchmarks often trigger hallucinations via prompt bias or semantically inconsistent scenes, which conflates multiple factors and complicates error attribution. Counting offers a sharper diagnostic because answers are objective and depend purely on visual enumeration~\citep{Paiss2023Teaching, qharabagh2025lvlmcount, Rahmanzadehgervi2024vision, sengupta2025can, campbell2024understanding}. However, VLMs can miscount even when attending to relevant objects~\citep{alghisi2025dereconstructing, hasani2024counting, vo2025vision2, rudman2025forgotten}. Mechanistic counting studies exist~\citep{che2026countingcircuits}, but focus on standard images without prior conflicts. Closest to our setting, \citep{vo2025vision2} uses counterfactual images of familiar subjects to test canonical expectations. Yet, their limited edits (e.g., adding one leg), inclusion of high-count cases, and reliance on geometric shapes make it difficult to isolate prior-driven failures from standard perceptual limits~\citep{rudman2025forgotten}. A similar benchmark, CDH-Bench~\citep{chen2026cdhbench}, studies the same prior-over-evidence failure but does not explicitly focus on counting, treating it as only one of three anomaly types, contains small counting pairs, includes ambiguous images, and lacks localized region annotations. We compare CounterCount with existing datasets in \Cref{tab:dataset_comparisonfinal}. CounterCount provides more image pairs, broader subject and attribute coverage, and is the only dataset with localized region annotations.

% In contrast, \emph{CounterCount} isolates prior-driven counting failures. We provide a focused diagnostic framework across 17 object classes, 379 subjects, and 228 attributes. Each example includes paired original-edited. images, verified answers, and localized annotations of count-relevant regions, supporting controlled interventions on visual tokens.

% \myparagraph{Attention Modulation in VLMs.} The tendency of VLMs to over-rely on language priors has motivated inference-time interventions that modify attention or decoding~\citep{liu2024paying, Tu2025attention, yu2025causally, Jiang2025devils, yang2025tracing, zhao2025tell}. Most methods target general hallucination by amplifying attention to visual tokens, suppressing text-dominant pathways, or using contrastive decoding~\citep{liu2024paying, zhao2025tell, Jiang2025devils, yu2025causally, Tu2025attention, yang2025understanding}. Furthermore, studies show that specific transformer layers and attention heads mediate vision-text conflicts~\citep{Jiang2025devils, yang2025understanding, wang2025vegas, ortu2025when}, motivating localized rather than global interventions. Despite this progress, attention modulation remains underexplored for counterfactual counting, where numerical priors directly conflict with visual evidence. 

% We introduce a simple unified inference-time modulation strategy that can amplify, suppress, or mask selected visual tokens, allowing us to probe and mitigate modality imbalance.

\section{Conclusion}
\label{sec:conclusion}
We introduced \emph{CounterCount}, a diagnostic dataset for studying whether VLMs count from visual evidence or canonical priors.
Across 16 VLMs, accuracy drops from 91.6\% on factual images to 49.5\% on CF images, while humans remain near the ceiling, and most errors return the canonical count.
Our analysis reveals two underlying behaviors:
(1)~Models exhibit System-1-like behavior: correct answers on factual images say little about whether they looked closely, since knowing the subject is enough, so paired counterfactuals are needed to measure visual grounding.
Breaking the visual context, cueing models that counts may be unusual, or letting them reason all lead them to rely more on the visual cues.
(2)~Models have the visual evidence but underuse it, as prior knowledge outweighs it. Redirecting their attention toward count-relevant regions shifts predictions toward the image.
We release CounterCount to support the community in analyzing and understanding such failures, toward VLMs that ground their answers in visual evidence rather than prior knowledge.

\myparagraph{Limitations.}
% CounterCount uses synthetic images, so model behavior on natural images may differ, but synthetic generation is necessary to obtain controlled pairs that differ only in the counted attribute.
CounterCount uses synthetic images, as the controlled construction of factual-counterfactual pairs requires changing a subject's attribute count while holding the remaining visual content fixed--a condition that is exceedingly difficult to satisfy with naturally occurring images. We attempted to cover a broad range of subjects and attributes across 17 categories, yet other subject types with strong canonical priors remain to be explored.
Finally, the hyperparameter search for our attention modulation was limited, and other configurations may have had even stronger debiasing effects.

\bibliography{references}

@inproceedings{Alayrac2022Flamingo,
 author = {Alayrac, Jean-Baptiste and Donahue, Jeff and Luc, Pauline and Miech, Antoine and Barr, Iain and Hasson, Yana and Lenc, Karel and Mensch, Arthur and Millican, Katherine and Reynolds, Malcolm and Ring, Roman and Rutherford, Eliza and Cabi, Serkan and Han, Tengda and Gong, Zhitao and Samangooei, Sina and Monteiro, Marianne and Menick, Jacob L and Borgeaud, Sebastian and Brock, Andy and Nematzadeh, Aida and Sharifzadeh, Sahand and Bi\'{n}kowski, Miko\l aj and Barreira, Ricardo and Vinyals, Oriol and Zisserman, Andrew and Simonyan, Kar\'{e}n},
 booktitle = {Advances in Neural Information Processing Systems},
 editor = {S. Koyejo and S. Mohamed and A. Agarwal and D. Belgrave and K. Cho and A. Oh},
 pages = {23716--23736},
 publisher = {Curran Associates, Inc.},
 title = {Flamingo: a Visual Language Model for Few-Shot Learning},
 url = {https://proceedings.neurips.cc/paper_files/paper/2022/file/960a172bc7fbf0177ccccbb411a7d800-Paper-Conference.pdf},
 volume = {35},
 year = {2022}
}

@InProceedings{Li2022BLIP,
  title = 	 {{BLIP}: Bootstrapping Language-Image Pre-training for Unified Vision-Language Understanding and Generation},
  author =       {Li, Junnan and Li, Dongxu and Xiong, Caiming and Hoi, Steven},
  booktitle = 	 {Proceedings of the 39th International Conference on Machine Learning},
  pages = 	 {12888--12900},
  year = 	 {2022},
  editor = 	 {Chaudhuri, Kamalika and Jegelka, Stefanie and Song, Le and Szepesvari, Csaba and Niu, Gang and Sabato, Sivan},
  volume = 	 {162},
  series = 	 {Proceedings of Machine Learning Research},
  month = 	 {17--23 Jul},
  publisher =    {PMLR},
  url = 	 {https://proceedings.mlr.press/v162/li22n.html}
}

@inproceedings{Liu2023visual,
 author = {Liu, Haotian and Li, Chunyuan and Wu, Qingyang and Lee, Yong Jae},
 booktitle = {Advances in Neural Information Processing Systems},
 editor = {A. Oh and T. Naumann and A. Globerson and K. Saenko and M. Hardt and S. Levine},
 pages = {34892--34916},
 publisher = {Curran Associates, Inc.},
 title = {Visual Instruction Tuning},
 url = {https://proceedings.neurips.cc/paper_files/paper/2023/file/6dcf277ea32ce3288914faf369fe6de0-Paper-Conference.pdf},
 volume = {36},
 year = {2023}
}

@misc{bai2023qwenvl,
      title={Qwen-VL: A Versatile Vision-Language Model for Understanding, Localization, Text Reading, and Beyond}, 
      author={Jinze Bai and Shuai Bai and Shusheng Yang and Shijie Wang and Sinan Tan and Peng Wang and Junyang Lin and Chang Zhou and Jingren Zhou},
      year={2023},
      eprint={2308.12966},
      archivePrefix={arXiv},
      primaryClass={cs.CV},
      url={https://arxiv.org/abs/2308.12966}, 
}

@inproceedings{
vo2025bscore,
title={B-score: Detecting biases in large language models using response history},
author={An Vo and Mohammad Reza Taesiri and Daeyoung Kim and Anh Totti Nguyen},
booktitle={Forty-second International Conference on Machine Learning},
year={2025},
url={https://icml.cc/virtual/2025/poster/44236}
}

@article{Zhang2024vision,
author = {Zhang, Jingyi and Huang, Jiaxing and Jin, Sheng and Lu, Shijian},
title = {Vision-Language Models for Vision Tasks: A Survey},
year = {2024},
issue_date = {Aug. 2024},
publisher = {IEEE Computer Society},
address = {USA},
volume = {46},
number = {8},
issn = {0162-8828},
url = {https://doi.org/10.1109/TPAMI.2024.3369699},
doi = {10.1109/TPAMI.2024.3369699},
journal = {IEEE Trans. Pattern Anal. Mach. Intell.},
month = aug,
pages = {5625–5644},
numpages = {20}
}

@inproceedings{sheng2019woman,
    title = "The Woman Worked as a Babysitter: On Biases in Language Generation",
    author = "Sheng, Emily  and
      Chang, Kai-Wei  and
      Natarajan, Premkumar  and
      Peng, Nanyun",
    editor = "Inui, Kentaro  and
      Jiang, Jing  and
      Ng, Vincent  and
      Wan, Xiaojun",
    booktitle = "Proceedings of the 2019 Conference on Empirical Methods in Natural Language Processing and the 9th International Joint Conference on Natural Language Processing (EMNLP-IJCNLP)",
    month = nov,
    year = "2019",
    address = "Hong Kong, China",
    publisher = "Association for Computational Linguistics",
    url = "https://aclanthology.org/D19-1339/",
    doi = "10.18653/v1/D19-1339",
    pages = "3407--3412"
}

@article{gallegos2024bias,
    title = "Bias and Fairness in Large Language Models: A Survey",
    author = "Gallegos, Isabel O.  and
      Rossi, Ryan A.  and
      Barrow, Joe  and
      Tanjim, Md Mehrab  and
      Kim, Sungchul  and
      Dernoncourt, Franck  and
      Yu, Tong  and
      Zhang, Ruiyi  and
      Ahmed, Nesreen K.",
    journal = "Computational Linguistics",
    volume = "50",
    number = "3",
    month = sep,
    year = "2024",
    address = "Cambridge, MA",
    publisher = "MIT Press",
    url = "https://aclanthology.org/2024.cl-3.8/",
    doi = "10.1162/coli_a_00524",
    pages = "1097--1179"
}

@inproceedings{
vo2025vision,
title={Vision Language Models are Biased: Counting legs of an animal is surprisingly hard},
author={An Vo and Khai-Nguyen Nguyen and Mohammad Reza Taesiri and Vy Tuong Dang and Anh Totti Nguyen and Daeyoung Kim},
booktitle={2nd AI for Math Workshop @ ICML 2025},
year={2025},
url={https://icml.cc/virtual/2025/52429}
}

@misc{
vo2025vision2,
title={Vision Language Models are Biased},
author={An Vo and Khai-Nguyen Nguyen and Mohammad Reza Taesiri and Vy Tuong Dang and Anh Totti Nguyen and Daeyoung Kim},
year={2025},
url={https://openreview.net/forum?id=4GWfYyo6FS}
}

@InProceedings{Rahmanzadehgervi2024vision,
    author    = {Rahmanzadehgervi, Pooyan and Bolton, Logan and Taesiri, Mohammad Reza and Nguyen, Anh Totti},
    title     = {Vision language models are blind},
    booktitle = {Proceedings of the Asian Conference on Computer Vision (ACCV)},
    month     = {December},
    year      = {2024},
    pages     = {18-34}
}

@inproceedings{lee2025vlind,
    title = "{VL}ind-Bench: Measuring Language Priors in Large Vision-Language Models",
    author = "Lee, Kang-il  and
      Kim, Minbeom  and
      Yoon, Seunghyun  and
      Kim, Minsung  and
      Lee, Dongryeol  and
      Koh, Hyukhun  and
      Jung, Kyomin",
    editor = "Chiruzzo, Luis  and
      Ritter, Alan  and
      Wang, Lu",
    booktitle = "Findings of the Association for Computational Linguistics: NAACL 2025",
    month = apr,
    year = "2025",
    address = "Albuquerque, New Mexico",
    publisher = "Association for Computational Linguistics",
    url = "https://aclanthology.org/2025.findings-naacl.231/",
    doi = "10.18653/v1/2025.findings-naacl.231",
    pages = "4129--4144",
    ISBN = "979-8-89176-195-7"
}

@inproceedings{
ortu2025when,
title={When seeing Overrides Knowing: Disentangling Knowledge Conflicts in Vision-Language Models},
author={Francesco Ortu and Zhijing Jin and Diego Doimo and Alberto Cazzaniga},
booktitle={Mechanistic Interpretability Workshop at NeurIPS 2025},
year={2025},
url={https://openreview.net/forum?id=GhSp6uvUuR}
}

@misc{liu2025seeing,
      title={Seeing but Not Believing: Probing the Disconnect Between Visual Attention and Answer Correctness in VLMs}, 
      author={Zhining Liu and Ziyi Chen and Hui Liu and Chen Luo and Xianfeng Tang and Suhang Wang and Joy Zeng and Zhenwei Dai and Zhan Shi and Tianxin Wei and Benoit Dumoulin and Hanghang Tong},
      year={2025},
      eprint={2510.17771},
      archivePrefix={arXiv},
      primaryClass={cs.AI},
      url={https://arxiv.org/abs/2510.17771}, 
}

@misc{wu2025lanp,
      title={LanP: Rethinking the Impact of Language Priors in Large Vision-Language Models}, 
      author={Zongyu Wu and Yuwei Niu and Hongcheng Gao and Minhua Lin and Zhiwei Zhang and Zhifang Zhang and Qi Shi and Yilong Wang and Sike Fu and Junjie Xu and Junjie Ao and Enyan Dai and Lei Feng and Xiang Zhang and Suhang Wang},
      year={2025},
      eprint={2502.12359},
      archivePrefix={arXiv},
      primaryClass={cs.CV},
      url={https://arxiv.org/abs/2502.12359}, 
}

@inproceedings{
yang2025escaping,
title={Escaping the SpuriVerse: Can Large Vision-Language Models Generalize Beyond Seen Spurious Correlations?},
author={Yiwei Yang and Chung Peng Lee and Shangbin Feng and Dora Zhao and Bingbing Wen and Anthony Zhe Liu and Yulia Tsvetkov and Bill Howe},
booktitle={The Thirty-ninth Annual Conference on Neural Information Processing Systems Datasets and Benchmarks Track},
year={2025},
url={https://neurips.cc/virtual/2025/loc/san-diego/poster/121529}
}

@InProceedings{Luo2025probing,
  title = 	 {Probing Visual Language Priors in {VLM}s},
  author =       {Luo, Tiange and Cao, Ang and Lee, Gunhee and Johnson, Justin and Lee, Honglak},
  booktitle = 	 {Proceedings of the 42nd International Conference on Machine Learning},
  pages = 	 {41120--41156},
  year = 	 {2025},
  editor = 	 {Singh, Aarti and Fazel, Maryam and Hsu, Daniel and Lacoste-Julien, Simon and Berkenkamp, Felix and Maharaj, Tegan and Wagstaff, Kiri and Zhu, Jerry},
  volume = 	 {267},
  series = 	 {Proceedings of Machine Learning Research},
  month = 	 {13--19 Jul},
  publisher =    {PMLR},
  url = 	 {https://proceedings.mlr.press/v267/luo25b.html}
}

@inproceedings{golovanevsky2025pixels,
    title = "Pixels Versus Priors: Controlling Knowledge Priors in Vision-Language Models through Visual Counterfacts",
    author = "Golovanevsky, Michal  and
      Rudman, William  and
      Lepori, Michael A.  and
      Bar, Amir  and
      Singh, Ritambhara  and
      Eickhoff, Carsten",
    editor = "Christodoulopoulos, Christos  and
      Chakraborty, Tanmoy  and
      Rose, Carolyn  and
      Peng, Violet",
    booktitle = "Proceedings of the 2025 Conference on Empirical Methods in Natural Language Processing",
    month = nov,
    year = "2025",
    address = "Suzhou, China",
    publisher = "Association for Computational Linguistics",
    url = "https://aclanthology.org/2025.emnlp-main.1262/",
    doi = "10.18653/v1/2025.emnlp-main.1262",
    pages = "24837--24852",
    ISBN = "979-8-89176-332-6"
}

@misc{fu2025hidden,
      title={Hidden in plain sight: VLMs overlook their visual representations}, 
      author={Stephanie Fu and Tyler Bonnen and Devin Guillory and Trevor Darrell},
      year={2025},
      eprint={2506.08008},
      archivePrefix={arXiv},
      primaryClass={cs.CV},
      url={https://arxiv.org/abs/2506.08008}, 
}

@misc{alghisi2025dereconstructing,
      title={[De|Re]constructing VLMs' Reasoning in Counting}, 
      author={Simone Alghisi and Gabriel Roccabruna and Massimo Rizzoli and Seyed Mahed Mousavi and Giuseppe Riccardi},
      year={2025},
      eprint={2510.19555},
      archivePrefix={arXiv},
      primaryClass={cs.CV},
      url={https://arxiv.org/abs/2510.19555}, 
}

@article{hasani2024counting,
  title={Understanding Counting Mechanisms in Large Language and Vision-Language Models},
  author={Hasani, Hosein and Izadi, Amirmohammad and Askari, Fatemeh and Bagherian, Mobin and Mohammadian, Sadegh and Izadi, Mohammad and Baghshah, Mahdieh Soleymani},
  journal={arXiv preprint arXiv:2511.17699},
  year={2024}
}

@article{sengupta2025can,
  title={Can Vision-Language Models Count? A Synthetic Benchmark and Analysis of Attention-Based Interventions},
  author={Sengupta, Saurav and Moradinasab, Nazanin and Liu, Jiebei and Brown, Donald E.},
  journal={arXiv preprint arXiv:2511.17722},
  year={2025}
}

@misc{
qharabagh2025lvlmcount,
title={{LVLM}-{COUNT}: Enhancing the Counting Ability of Large Vision-Language Models},
author={Muhammad Fetrat Qharabagh and Mohammadreza Ghofrani and Kimon Fountoulakis},
year={2025},
url={https://openreview.net/forum?id=GsCMKwyfWm}
}

@misc{chan2026system,
      title={System-Mediated Attention Imbalances Make Vision-Language Models Say Yes}, 
      author={Tsan Tsai Chan and Varsha Suresh and Anisha Saha and Michael Hahn and Vera Demberg},
      year={2026},
      eprint={2601.12430},
      archivePrefix={arXiv},
      primaryClass={cs.CL},
      url={https://arxiv.org/abs/2601.12430}, 
}

@inproceedings{parrish2022bbq,
    title = "{BBQ}: A hand-built bias benchmark for question answering",
    author = "Parrish, Alicia  and
      Chen, Angelica  and
      Nangia, Nikita  and
      Padmakumar, Vishakh  and
      Phang, Jason  and
      Thompson, Jana  and
      Htut, Phu Mon  and
      Bowman, Samuel",
    editor = "Muresan, Smaranda  and
      Nakov, Preslav  and
      Villavicencio, Aline",
    booktitle = "Findings of the Association for Computational Linguistics: ACL 2022",
    month = may,
    year = "2022",
    address = "Dublin, Ireland",
    publisher = "Association for Computational Linguistics",
    url = "https://aclanthology.org/2022.findings-acl.165/",
    doi = "10.18653/v1/2022.findings-acl.165",
    pages = "2086--2105"
}

@inproceedings{wang2024countries,
    title = "Not All Countries Celebrate Thanksgiving: On the Cultural Dominance in Large Language Models",
    author = "Wang, Wenxuan  and
      Jiao, Wenxiang  and
      Huang, Jingyuan  and
      Dai, Ruyi  and
      Huang, Jen-tse  and
      Tu, Zhaopeng  and
      Lyu, Michael",
    editor = "Ku, Lun-Wei  and
      Martins, Andre  and
      Srikumar, Vivek",
    booktitle = "Proceedings of the 62nd Annual Meeting of the Association for Computational Linguistics (Volume 1: Long Papers)",
    month = aug,
    year = "2024",
    address = "Bangkok, Thailand",
    publisher = "Association for Computational Linguistics",
    url = "https://aclanthology.org/2024.acl-long.345/",
    doi = "10.18653/v1/2024.acl-long.345",
    pages = "6349--6384"
}

@inproceedings{naous2024beer,
    title = "Having Beer after Prayer? Measuring Cultural Bias in Large Language Models",
    author = "Naous, Tarek  and
      Ryan, Michael J  and
      Ritter, Alan  and
      Xu, Wei",
    editor = "Ku, Lun-Wei  and
      Martins, Andre  and
      Srikumar, Vivek",
    booktitle = "Proceedings of the 62nd Annual Meeting of the Association for Computational Linguistics (Volume 1: Long Papers)",
    month = aug,
    year = "2024",
    address = "Bangkok, Thailand",
    publisher = "Association for Computational Linguistics",
    url = "https://aclanthology.org/2024.acl-long.862/",
    doi = "10.18653/v1/2024.acl-long.862",
    pages = "16366--16393"
}

@inproceedings{shin2024ask,
    title = "Ask {LLM}s Directly, ``What shapes your bias?'': Measuring Social Bias in Large Language Models",
    author = "Shin, Jisu  and
      Song, Hoyun  and
      Lee, Huije  and
      Jeong, Soyeong  and
      Park, Jong",
    editor = "Ku, Lun-Wei  and
      Martins, Andre  and
      Srikumar, Vivek",
    booktitle = "Findings of the Association for Computational Linguistics: ACL 2024",
    month = aug,
    year = "2024",
    address = "Bangkok, Thailand",
    publisher = "Association for Computational Linguistics",
    url = "https://aclanthology.org/2024.findings-acl.954/",
    doi = "10.18653/v1/2024.findings-acl.954",
    pages = "16122--16143"
}

@inproceedings{
hall2023visogender,
title={VisoGender:  A dataset for benchmarking gender bias in image-text pronoun resolution},
author={Siobhan Mackenzie Hall and Fernanda Gon{\c{c}}alves Abrantes and Hanwen Zhu and Grace Sodunke and Aleksandar Shtedritski and Hannah Rose Kirk},
booktitle={Thirty-seventh Conference on Neural Information Processing Systems Datasets and Benchmarks Track},
year={2023},
url={https://neurips.cc/virtual/2023/poster/73666}
}

@inproceedings{raj2024biasdora,
    title = "{B}ias{D}ora: Exploring Hidden Biased Associations in Vision-Language Models",
    author = "Raj, Chahat  and
      Mukherjee, Anjishnu  and
      Caliskan, Aylin  and
      Anastasopoulos, Antonios  and
      Zhu, Ziwei",
    editor = "Al-Onaizan, Yaser  and
      Bansal, Mohit  and
      Chen, Yun-Nung",
    booktitle = "Findings of the Association for Computational Linguistics: EMNLP 2024",
    month = nov,
    year = "2024",
    address = "Miami, Florida, USA",
    publisher = "Association for Computational Linguistics",
    url = "https://aclanthology.org/2024.findings-emnlp.611/",
    doi = "10.18653/v1/2024.findings-emnlp.611",
    pages = "10439--10455"
}

@InProceedings{Howard2024Social,
    author    = {Howard, Phillip and Madasu, Avinash and Le, Tiep and Moreno, Gustavo Lujan and Bhiwandiwalla, Anahita and Lal, Vasudev},
    title     = {SocialCounterfactuals: Probing and Mitigating Intersectional Social Biases in Vision-Language Models with Counterfactual Examples},
    booktitle = {Proceedings of the IEEE/CVF Conference on Computer Vision and Pattern Recognition (CVPR)},
    month     = {June},
    year      = {2024},
    pages     = {11975-11985}
}

@misc{ullman2024illusion,
      title={The Illusion-Illusion: Vision Language Models See Illusions Where There are None}, 
      author={Tomer Ullman},
      year={2024},
      eprint={2412.18613},
      archivePrefix={arXiv},
      primaryClass={q-bio.NC},
      url={https://arxiv.org/abs/2412.18613}, 
}

@InProceedings{Deng2025Words,
    author    = {Deng, Ailin and Cao, Tri and Chen, Zhirui and Hooi, Bryan},
    title     = {Words or Vision: Do Vision-Language Models Have Blind Faith in Text?},
    booktitle = {Proceedings of the IEEE/CVF Conference on Computer Vision and Pattern Recognition (CVPR)},
    month     = {June},
    year      = {2025},
    pages     = {3867-3876}
}

@InProceedings{Tong2024Eyes,
    author    = {Tong, Shengbang and Liu, Zhuang and Zhai, Yuexiang and Ma, Yi and LeCun, Yann and Xie, Saining},
    title     = {Eyes Wide Shut? Exploring the Visual Shortcomings of Multimodal LLMs},
    booktitle = {Proceedings of the IEEE/CVF Conference on Computer Vision and Pattern Recognition (CVPR)},
    month     = {June},
    year      = {2024},
    pages     = {9568-9578}
}

@inproceedings{zhao2025looking,
    title = "Looking Beyond Text: Reducing Language Bias in Large Vision-Language Models via Multimodal Dual-Attention and Soft-Image Guidance",
    author = "Zhao, Haozhe  and
      Si, Shuzheng  and
      Chen, Liang  and
      Zhang, Yichi  and
      Sun, Maosong  and
      Chang, Baobao  and
      Zhang, Minjia",
    editor = "Christodoulopoulos, Christos  and
      Chakraborty, Tanmoy  and
      Rose, Carolyn  and
      Peng, Violet",
    booktitle = "Proceedings of the 2025 Conference on Empirical Methods in Natural Language Processing",
    month = nov,
    year = "2025",
    address = "Suzhou, China",
    publisher = "Association for Computational Linguistics",
    url = "https://aclanthology.org/2025.emnlp-main.995/",
    doi = "10.18653/v1/2025.emnlp-main.995",
    pages = "19666--19690",
    ISBN = "979-8-89176-332-6"
}

@misc{liu2024paying,
      title={Paying More Attention to Image: A Training-Free Method for Alleviating Hallucination in LVLMs}, 
      author={Shi Liu and Kecheng Zheng and Wei Chen},
      year={2024},
      booktitle = {The European Conference on Computer Vision (ECCV)},
      url={https://eccv.ecva.net/virtual/2024/poster/2599}, 
}

@InProceedings{Favero2024multi,
    author    = {Favero, Alessandro and Zancato, Luca and Trager, Matthew and Choudhary, Siddharth and Perera, Pramuditha and Achille, Alessandro and Swaminathan, Ashwin and Soatto, Stefano},
    title     = {Multi-Modal Hallucination Control by Visual Information Grounding},
    booktitle = {Proceedings of the IEEE/CVF Conference on Computer Vision and Pattern Recognition (CVPR)},
    month     = {June},
    year      = {2024},
    pages     = {14303-14312}
}

@InProceedings{Paiss2023Teaching,
    author    = {Paiss, Roni and Ephrat, Ariel and Tov, Omer and Zada, Shiran and Mosseri, Inbar and Irani, Michal and Dekel, Tali},
    title     = {Teaching CLIP to Count to Ten},
    booktitle = {Proceedings of the IEEE/CVF International Conference on Computer Vision (ICCV)},
    month     = {October},
    year      = {2023},
    pages     = {3170-3180}
}

@inproceedings{
campbell2024understanding,
title={Understanding the Limits of Vision Language Models Through the Lens of the Binding Problem},
author={Declan Iain Campbell and Sunayana Rane and Tyler Giallanza and C. Nicol{\`o} De Sabbata and Kia Ghods and Amogh Joshi and Alexander Ku and Steven M Frankland and Thomas L. Griffiths and Jonathan D. Cohen and Taylor Whittington Webb},
booktitle={The Thirty-eighth Annual Conference on Neural Information Processing Systems},
year={2024},
url={https://neurips.cc/virtual/2024/poster/95266}
}

@InProceedings{Jiang2025devils,
    author    = {Jiang, Zhangqi and Chen, Junkai and Zhu, Beier and Luo, Tingjin and Shen, Yankun and Yang, Xu},
    title     = {Devils in Middle Layers of Large Vision-Language Models: Interpreting, Detecting and Mitigating Object Hallucinations via Attention Lens},
    booktitle = {Proceedings of the IEEE/CVF Conference on Computer Vision and Pattern Recognition (CVPR)},
    month     = {June},
    year      = {2025},
    pages     = {25004-25014}
}

@article{Tu2025attention,
author = {Tu, Chongjun and Ye, Peng and Zhou, Dongzhan and Bai, Lei and Yu, Gang and Chen, Tao and Ouyang, Wanli},
title = {Attention Reallocation: Towards Zero-cost and Controllable Hallucination Mitigation of MLLMs},
year = {2025},
issue_date = {Jan 2026},
publisher = {Kluwer Academic Publishers},
address = {USA},
volume = {134},
number = {1},
issn = {0920-5691},
url = {https://doi.org/10.1007/s11263-025-02607-z},
doi = {10.1007/s11263-025-02607-z},
journal = {Int. J. Comput. Vision},
month = dec,
numpages = {29}
}

@misc{yu2025causally,
      title={Causally-Grounded Dual-Path Attention Intervention for Object Hallucination Mitigation in LVLMs}, 
      author={Liu Yu and Zhonghao Chen and Ping Kuang and Zhikun Feng and Fan Zhou and Lan Wang and Gillian Dobbie},
      year={2025},
      eprint={2511.09018},
      archivePrefix={arXiv},
      primaryClass={cs.CV},
      url={https://arxiv.org/abs/2511.09018}, 
booktitle = {AAAI},
}

@misc{zhao2025tell,
      title={Tell Model Where to Look: Mitigating Hallucinations in MLLMs by Vision-Guided Attention}, 
      author={Jianfei Zhao and Feng Zhang and Xin Sun and Chong Feng and Zhixing Tan},
      year={2025},
      eprint={2511.20032},
      archivePrefix={arXiv},
      primaryClass={cs.CV},
      url={https://arxiv.org/abs/2511.20032}, 
}

@misc{gemma2025gemma3,
      title={Gemma 3 Technical Report}, 
      author={Aishwarya Kamath and Johan Ferret and Shreya Pathak and Nino Vieillard and Ramona Merhej and Sarah Perrin and Tatiana Matejovicova and Alexandre Ramé and Morgane Rivière and Louis Rouillard and Thomas Mesnard and Geoffrey Cideron and Jean-bastien Grill and Sabela Ramos and Edouard Yvinec and Michelle Casbon and Etienne Pot and Ivo Penchev and Gaël Liu and Francesco Visin and Kathleen Kenealy and Lucas Beyer and Xiaohai Zhai and Anton Tsitsulin and Robert Busa-Fekete and Alex Feng and Noveen Sachdeva and Benjamin Coleman and Yi Gao and Basil Mustafa and Iain Barr and Emilio Parisotto and David Tian and Matan Eyal and Colin Cherry and Jan-Thorsten Peter and Danila Sinopalnikov and Surya Bhupatiraju and Rishabh Agarwal and Mehran Kazemi and Dan Malkin and Ravin Kumar and David Vilar and Idan Brusilovsky and Jiaming Luo and Andreas Steiner and Abe Friesen and Abhanshu Sharma and Abheesht Sharma and Adi Mayrav Gilady and Adrian Goedeckemeyer and Alaa Saade and Alex Feng and Alexander Kolesnikov and Alexei Bendebury and Alvin Abdagic and Amit Vadi and András György and André Susano Pinto and Anil Das and Ankur Bapna and Antoine Miech and Antoine Yang and Antonia Paterson and Ashish Shenoy and Ayan Chakrabarti and Bilal Piot and Bo Wu and Bobak Shahriari and Bryce Petrini and Charlie Chen and Charline Le Lan and Christopher A. Choquette-Choo and CJ Carey and Cormac Brick and Daniel Deutsch and Danielle Eisenbud and Dee Cattle and Derek Cheng and Dimitris Paparas and Divyashree Shivakumar Sreepathihalli and Doug Reid and Dustin Tran and Dustin Zelle and Eric Noland and Erwin Huizenga and Eugene Kharitonov and Frederick Liu and Gagik Amirkhanyan and Glenn Cameron and Hadi Hashemi and Hanna Klimczak-Plucińska and Harman Singh and Harsh Mehta and Harshal Tushar Lehri and Hussein Hazimeh and Ian Ballantyne and Idan Szpektor and Ivan Nardini and Jean Pouget-Abadie and Jetha Chan and Joe Stanton and John Wieting and Jonathan Lai and Jordi Orbay and Joseph Fernandez and Josh Newlan and Ju-yeong Ji and Jyotinder Singh and Kat Black and Kathy Yu and Kevin Hui and Kiran Vodrahalli and Klaus Greff and Linhai Qiu and Marcella Valentine and Marina Coelho and Marvin Ritter and Matt Hoffman and Matthew Watson and Mayank Chaturvedi and Michael Moynihan and Min Ma and Nabila Babar and Natasha Noy and Nathan Byrd and Nick Roy and Nikola Momchev and Nilay Chauhan and Noveen Sachdeva and Oskar Bunyan and Pankil Botarda and Paul Caron and Paul Kishan Rubenstein and Phil Culliton and Philipp Schmid and Pier Giuseppe Sessa and Pingmei Xu and Piotr Stanczyk and Pouya Tafti and Rakesh Shivanna and Renjie Wu and Renke Pan and Reza Rokni and Rob Willoughby and Rohith Vallu and Ryan Mullins and Sammy Jerome and Sara Smoot and Sertan Girgin and Shariq Iqbal and Shashir Reddy and Shruti Sheth and Siim Põder and Sijal Bhatnagar and Sindhu Raghuram Panyam and Sivan Eiger and Susan Zhang and Tianqi Liu and Trevor Yacovone and Tyler Liechty and Uday Kalra and Utku Evci and Vedant Misra and Vincent Roseberry and Vlad Feinberg and Vlad Kolesnikov and Woohyun Han and Woosuk Kwon and Xi Chen and Yinlam Chow and Yuvein Zhu and Zichuan Wei and Zoltan Egyed and Victor Cotruta and Minh Giang and Phoebe Kirk and Anand Rao and Kat Black and Nabila Babar and Jessica Lo and Erica Moreira and Luiz Gustavo Martins and Omar Sanseviero and Lucas Gonzalez and Zach Gleicher and Tris Warkentin and Vahab Mirrokni and Evan Senter and Eli Collins and Joelle Barral and Zoubin Ghahramani and Raia Hadsell and Yossi Matias and D. Sculley and Slav Petrov and Noah Fiedel and Noam Shazeer and Oriol Vinyals and Jeff Dean and Demis Hassabis and Koray Kavukcuoglu and Clement Farabet and Elena Buchatskaya and Jean-Baptiste Alayrac and Rohan Anil and Dmitry and Lepikhin and Sebastian Borgeaud and Olivier Bachem and Armand Joulin and Alek Andreev and Cassidy Hardin and Robert Dadashi and Léonard Hussenot},
      year={2025},
      eprint={2503.19786},
      archivePrefix={arXiv},
      primaryClass={cs.CL},
      url={https://arxiv.org/abs/2503.19786}, 
}

@misc{bai2025qwen3vl,
      title={Qwen3-VL Technical Report}, 
      author={Shuai Bai and Yuxuan Cai and Ruizhe Chen and Keqin Chen and Xionghui Chen and Zesen Cheng and Lianghao Deng and Wei Ding and Chang Gao and Chunjiang Ge and Wenbin Ge and Zhifang Guo and Qidong Huang and Jie Huang and Fei Huang and Binyuan Hui and Shutong Jiang and Zhaohai Li and Mingsheng Li and Mei Li and Kaixin Li and Zicheng Lin and Junyang Lin and Xuejing Liu and Jiawei Liu and Chenglong Liu and Yang Liu and Dayiheng Liu and Shixuan Liu and Dunjie Lu and Ruilin Luo and Chenxu Lv and Rui Men and Lingchen Meng and Xuancheng Ren and Xingzhang Ren and Sibo Song and Yuchong Sun and Jun Tang and Jianhong Tu and Jianqiang Wan and Peng Wang and Pengfei Wang and Qiuyue Wang and Yuxuan Wang and Tianbao Xie and Yiheng Xu and Haiyang Xu and Jin Xu and Zhibo Yang and Mingkun Yang and Jianxin Yang and An Yang and Bowen Yu and Fei Zhang and Hang Zhang and Xi Zhang and Bo Zheng and Humen Zhong and Jingren Zhou and Fan Zhou and Jing Zhou and Yuanzhi Zhu and Ke Zhu},
      year={2025},
      eprint={2511.21631},
      archivePrefix={arXiv},
      primaryClass={cs.CV},
      url={https://arxiv.org/abs/2511.21631}, 
}

@misc{wang2025internvl35,
      title={InternVL3.5: Advancing Open-Source Multimodal Models in Versatility, Reasoning, and Efficiency}, 
      author={Weiyun Wang and Zhangwei Gao and Lixin Gu and Hengjun Pu and Long Cui and Xingguang Wei and Zhaoyang Liu and Linglin Jing and Shenglong Ye and Jie Shao and Zhaokai Wang and Zhe Chen and Hongjie Zhang and Ganlin Yang and Haomin Wang and Qi Wei and Jinhui Yin and Wenhao Li and Erfei Cui and Guanzhou Chen and Zichen Ding and Changyao Tian and Zhenyu Wu and Jingjing Xie and Zehao Li and Bowen Yang and Yuchen Duan and Xuehui Wang and Zhi Hou and Haoran Hao and Tianyi Zhang and Songze Li and Xiangyu Zhao and Haodong Duan and Nianchen Deng and Bin Fu and Yinan He and Yi Wang and Conghui He and Botian Shi and Junjun He and Yingtong Xiong and Han Lv and Lijun Wu and Wenqi Shao and Kaipeng Zhang and Huipeng Deng and Biqing Qi and Jiaye Ge and Qipeng Guo and Wenwei Zhang and Songyang Zhang and Maosong Cao and Junyao Lin and Kexian Tang and Jianfei Gao and Haian Huang and Yuzhe Gu and Chengqi Lyu and Huanze Tang and Rui Wang and Haijun Lv and Wanli Ouyang and Limin Wang and Min Dou and Xizhou Zhu and Tong Lu and Dahua Lin and Jifeng Dai and Weijie Su and Bowen Zhou and Kai Chen and Yu Qiao and Wenhai Wang and Gen Luo},
      year={2025},
      eprint={2508.18265},
      archivePrefix={arXiv},
      primaryClass={cs.CV},
      url={https://arxiv.org/abs/2508.18265}, 
}

@InProceedings{Yue2024MMMU,
    author    = {Yue, Xiang and Ni, Yuansheng and Zhang, Kai and Zheng, Tianyu and Liu, Ruoqi and Zhang, Ge and Stevens, Samuel and Jiang, Dongfu and Ren, Weiming and Sun, Yuxuan and Wei, Cong and Yu, Botao and Yuan, Ruibin and Sun, Renliang and Yin, Ming and Zheng, Boyuan and Yang, Zhenzhu and Liu, Yibo and Huang, Wenhao and Sun, Huan and Su, Yu and Chen, Wenhu},
    title     = {MMMU: A Massive Multi-discipline Multimodal Understanding and Reasoning Benchmark for Expert AGI},
    booktitle = {Proceedings of the IEEE/CVF Conference on Computer Vision and Pattern Recognition (CVPR)},
    month     = {June},
    year      = {2024},
    pages     = {9556-9567}
}

@InProceedings{liu2024mmbench,
    title={MMBench: Is Your Multi-modal Model an All-around Player?}, 
    author={Yuan Liu and Haodong Duan and Yuanhan Zhang and Bo Li and Songyang Zhang and Wangbo Zhao and Yike Yuan and Jiaqi Wang and Conghui He and Ziwei Liu and Kai Chen and Dahua Lin},
    booktitle = {The European Conference on Computer Vision (ECCV)},
    year      = {2024}
}

@InProceedings{Antol2015vqa,
author = {Antol, Stanislaw and Agrawal, Aishwarya and Lu, Jiasen and Mitchell, Margaret and Batra, Dhruv and Zitnick, C. Lawrence and Parikh, Devi},
title = {VQA: Visual Question Answering},
booktitle = {Proceedings of the IEEE International Conference on Computer Vision (ICCV)},
month = {December},
year = {2015}
}

@InProceedings{Hudson2019gqa,
author = {Hudson, Drew A. and Manning, Christopher D.},
title = {GQA: A New Dataset for Real-World Visual Reasoning and Compositional Question Answering},
booktitle = {Proceedings of the IEEE/CVF Conference on Computer Vision and Pattern Recognition (CVPR)},
month = {June},
year = {2019}
}

@inproceedings{
lu2024mathvista,
title={MathVista: Evaluating Mathematical Reasoning of Foundation Models in Visual Contexts},
author={Pan Lu and Hritik Bansal and Tony Xia and Jiacheng Liu and Chunyuan Li and Hannaneh Hajishirzi and Hao Cheng and Kai-Wei Chang and Michel Galley and Jianfeng Gao},
booktitle={The Twelfth International Conference on Learning Representations},
year={2024},
url={https://iclr.cc/virtual/2024/oral/19768}
}

@inproceedings{Yu2024mmvet,
author = {Yu, Weihao and Yang, Zhengyuan and Li, Linjie and Wang, Jianfeng and Lin, Kevin and Liu, Zicheng and Wang, Xinchao and Wang, Lijuan},
title = {MM-Vet: evaluating large multimodal models for integrated capabilities},
year = {2024},
publisher = {JMLR.org},
booktitle = {Proceedings of the 41st International Conference on Machine Learning},
articleno = {2381},
numpages = {25},
location = {Vienna, Austria},
series = {ICML'24}
}

@inproceedings{rudman2025forgotten,
    title = "Forgotten Polygons: Multimodal Large Language Models are Shape-Blind",
    author = "Rudman, William  and
      Golovanevsky, Michal  and
      Bar, Amir  and
      Palit, Vedant  and
      LeCun, Yann  and
      Eickhoff, Carsten  and
      Singh, Ritambhara",
    editor = "Che, Wanxiang  and
      Nabende, Joyce  and
      Shutova, Ekaterina  and
      Pilehvar, Mohammad Taher",
    booktitle = "Findings of the Association for Computational Linguistics: ACL 2025",
    month = jul,
    year = "2025",
    address = "Vienna, Austria",
    publisher = "Association for Computational Linguistics",
    url = "https://aclanthology.org/2025.findings-acl.620/",
    doi = "10.18653/v1/2025.findings-acl.620",
    pages = "11983--11998",
    ISBN = "979-8-89176-256-5"
}

@inproceedings{pezeshkpour2025benchmarking,
  author = {Atabuzzaman, Md. and Asgarov, Ali and Thomas, Chris},
  booktitle = {Proceedings of the 2025 Conference on Empirical Methods in Natural Language Processing (EMNLP)},
  title = {Benchmarking and Mitigating MCQA Selection Bias of Large Vision-Language Models},
  url = {https://aclanthology.org/2025.emnlp-main.1703.pdf},
  publisher = {Association for Computational Linguistics},
  year = {2025}
}

@inproceedings{zeno2025choosing,
  author = {Zeno, Giselle and Jedidi, Nour and Gomez, Steven R.},
  booktitle = {Proceedings of the IEEE/CVF Conference on Computer Vision and Pattern Recognition (CVPR) Workshops},
  pages = {535--544},
  title = {Choosing `Right' from Wrong: A Closer Look at Selection Bias in Spatial Multiple-Choice Questions in Large Multimodal Models},
  url = {https://openaccess.thecvf.com/content/CVPR2025W/BEAM/html/Zeno_Choosing_Right_from_Wrong_A_Closer_Look_at_Selection_Bias_CVPRW_2025_paper.html},
  year = {2025}
}

@misc{che2026countingcircuits,
      title={Counting Circuits: Mechanistic Interpretability of Visual Reasoning in Large Vision-Language Models}, 
      author={Liwei Che and Zhiyu Xue and Yihao Quan and Benlin Liu and Zeru Shi and Michelle Hurst and Jacob Feldman and Ruixiang Tang and Ranjay Krishna and Vladimir Pavlovic},
      year={2026},
      eprint={2603.18523},
      archivePrefix={arXiv},
      primaryClass={cs.CV},
    url={https://arxiv.org/abs/2603.18523}, 
}

@misc{qwen35,
  author       = {{Qwen Team}},
  title        = {Qwen3.5: Towards Native Multimodal Intelligence},
  howpublished = {\url{https://qwen.ai/blog?id=qwen3.5}},
  year         = {2026},
  month        = {February},
  note         = {Accessed: 2026-09-10}
}

@misc{qwen36_27b,
  author       = {{Qwen Team}},
  title        = {{Qwen3.6-27B}: Flagship-Level Coding in a {27B} Dense Model},
  howpublished = {\url{https://qwen.ai/blog?id=qwen3.6-27b}},
  year         = {2026},
  month        = {April},
  note         = {Accessed: 2026-09-10}
}

@misc{qwen38repo,
  author       = {{Qwen Team}},
  title        = {{Qwen3.8}: Large Language Model Series},
  year         = {2026},
  publisher    = {GitHub},
  journal      = {GitHub repository},
  howpublished = {\url{https://github.com/QwenLM/Qwen3.8}},
  note         = {Accessed: 2026-09-10}
}

@misc{gemma4,
      title={Gemma 4 Technical Report}, 
      author={Sherif El Abd and Vaibhav Aggarwal and Robin Algayres and Alek Andreev and Olivier Bachem and Ian Ballantyne and Cormac Brick and Victor Cărbune and Michelle Casbon and Mayank Chaturvedi and Aditya Chawla and Victor Cotruta and Alice Coucke and Phil Culliton and Robert Dadashi and Lucas Dixon and Mohamed Elhawaty and Utku Evci and Clément Farabet and Johan Ferret and Filippo Galgani and Sertan Girgin and Jean-Bastien Grill and Maarten Grootendorst and Jiaxian Guo and Cassidy Hardin and Yanzhang He and Steven M. Hernandez and Omri Homburger and Léonard Hussenot and Juyeong Ji and Armand Joulin and Aishwarya Kamath and Parnian Kassraie and Olivier Lacombe and Preethi Lahoti and Gaël Liu and Gus Martins and Luciano Martins and Tatiana Matejovicova and Ramona Merhej and Nikola Momchev and Sneha Mondal and Ryan Mullins and Sindhu Raghuram Panyam and Shreya Pathak and Sarah Perrin and André Susano Pinto and Etienne Pot and Angéline Pouget and Alexandre Ramé and Sabela Ramos and Douglas Reid and David Rim and Morgane Rivière and Karsten Roth and Louis Rouillard and Omar Sanseviero and Pier Giuseppe Sessa and Shane Settle and Danila Sinopalnikov and Sara Smoot and Piotr Stanczyk and Andreas Steiner and Lawrence Stewart and Ilya Tolstikhin and Michael Tschannen and Anton Tsitsulin and Nino Vieillard and Renjie Wu and Pingmei Xu and Haichuan Yang and Edouard Yvinec and Biao Zhang and Li Zhang and Joe Zou and Nicolas Aagnes and Abdelrahman Abdelhamed and Jakub Adamek and Shivani Agrawal and Shubham Agrawal and Ibrahim Alabdulmohsin and Jean Baptiste Alayrac and Uri Alon and Chandramouli Amarnath and Ankesh Anand and Chrysovalantis Anastasiou and Setareh Ariafar and François-Xavier Aubet and Kyriakos Axiotis and Federico Barbero and Joelle Barral and Alexei Bendebury and Urs Bergmann and Stanley Bileschi and Kat Black and Mathieu Blondel and Sebastian Borgeaud and Arthur Bražinskas and Ryan Burnell and Robert Busa-Fekete and Mu Cai and Daniele Calandriello and Glenn Cameron and Charlotte Caucheteux and Rahma Chaabouni and Garima Chadha and Jetha Chan and Blake Jianhang Chen and Jesse Chen and Lin Chen and Xu Chen and Derek Cheng and Tzu-hsiang Chien and Nikolai Chinaev and Yi Chou and Zhaohui Chu and Benjamin Coleman and Pooja Consul and Sam Conway-Rahman and Scott Crowell and Dylan Cutler and Vivek Dani and Samira Daruki and Anil Das and Daniel Deutsch and Nishanth Dikkala and Li Ding and Qiuhan Ding and Shenil Dodhia and Konstantin Donhauser and Tulsee Doshi and Anca Dragan and Alex Druinsky and Sahil Dua and Zoltan Egyed and Danielle Eisenbud and Daniel Eppens and Cindy Fan and Bahare Fatemi and Yassir Fathullah and Vlad Feinberg and Milen Ferev and Sebastian Flennerhag and Takumi Fujimoto and João Gabriel Oliveira and Isaac Galatzer-Levy and João Gante and Simon Geisler and Soham Ghosal and Antonious M. Girgis and Tamara von Glehn and Alec Go and Alhaad Gokhale and Alex Grills and Yiming Gu and Mayank Gupta and Pramod Gupta and Guru Guruganesh and Raia Hadsell and Hamza Harkous and Jitendra Harlalka and Demis Hassabis and Anja Hauth and Joe Heyward and Arian Hosseini and Chih-Yang Hsia and I-Hung Hsu and Xiaopeng Huang and Yangsibo Huang and Kevin Hui and Adrian Hutter and Te I and Fotis Iliopoulos and Advait Jain and Ganesh Jawahar and Ziwei Ji and Qilin Jin and Melvin Johnson and Kandarp Joshi and Arun Kandoor and Wang-Cheng Kang and Koray Kavukcuoglu and Mehran Kazemi and Kathleen Kenealy and Amr Khalifa and Phoebe Kirk and Ivan Korotkov and Suraj Kothawade and Vitaly Kovalev and Neel Kovelamudi and Adam Kraft and Ravin Kumar and Vivek Kumar and Harish Kuppam and Justin Lannin and Chen-Yu Lee and Seungji Lee and Dmitry Lepikhin and Alon Levkovitch and Dongdong Li and Qiujia Li and Valentin Liévin and Ethan Lin and Ziqian Lin and Casper Liu and Tianlin Liu and Tianqi Liu and Xin Liu and Ivan Lobov and Mayank Lunayach and Min Ma and Gagan Madan and Andrii Maksai and Eric Malmi and Michal Matuszak and Daniel McDuff and Gaurav Menghani and Maciej Mikuła and Daniil Mirylenka and Karolis Misiunas and Vedant Misra and Andreea Mitran and Kareem Mohamed and Maksim Mukha and Eric Noland and James O'Donnell and Brendan O'Donoghue and Kate Olszewska and Bernett Orlando and Wanqiong Pan and Rina Panigrahy and Unnati Parekh and Nicolas Perez-Nieves and Chunjong Park and Eric Paskie and Liqian Peng and Bryce Petrini and Slav Petrov and Jonas Pfeiffer and Bilal Piot and Martyna Plomecka and Siim Poder and Octavio Ponce and Arijit Pramanik and David Racz and Anish Rajan and Michelle Ramanovich and Anand Rao and Marvin Ritter and Vitor Rodrigues and Evan Rosen and Mikołaj Rybiński and Noveen Sachdeva and Michaël E. Sander and Rohit Sathyanarayana and Sagar Savla and Samuel Schmidgall and Tal Schuster and George Scrivener and Benoit Seguin and Andrew Sellergren and Aliaksei Severyn and Izhak Shafran and Dhruv Shah and Bobak Shahriari and Yuan Shangguan and Ashish Shenoy and Pradeep Shenoy and Rakesh Shivanna and Pauline Sho and Lucas Spangher and Wojciech Stokowiec and Tim Strother and Yao Su and Yinghao Sun and Mukund Sundararajan and Andrea Tacchetti and Mor Hazan Taege and Pouya Tafti and Jean Tarbouriech and Chetan Tekur and Shantanu Thakoor and Rahul Thapa and Madeleine Traverse and Lenart Treven and Tao Tu and Chien Te Tung and Çağlar Ünlü and Petar Veličković and Malini Pooni Venkat and Sagar Gubbi Venkatesh and Vidya Venkiteswaran and Francesco Visin and Alex Vitvitskyi and Kiran Vodrahalli and Weiyi Wang and Xin Wang and Tris Warkentin and Jan Wassenberg and John Wieting and Cindy Wu and Lechao Xiao and Hao Xu and Yuhui Xu and Fuzhao Xue and Arun Yadav and Jun Yan and Antoine Yang and Lin Yang and Ming-Hsuan Yang and Ziyu Ying and Jae Hyeon Yoo and Morteza Zadimoghaddam and Sajjad Zafar and Fred Zhang and Jiageng Zhang and Jianyi Zhang and Xiaofan Zhang and Chao Zhao and David Zhou and Chen Zou},
      year={2026},
      eprint={2607.02770},
      archivePrefix={arXiv},
      primaryClass={cs.CL},
      url={https://arxiv.org/abs/2607.02770}, 
}

@misc{sonnet5,
  author       = {{Anthropic}},
  title        = {Introducing {Claude Sonnet 5}},
  howpublished = {\url{https://www.anthropic.com/news/claude-sonnet-5}},
  year         = {2026},
  month        = {June},
  note         = {Accessed: 2026-09-13}
}

@misc{gpt56,
  author       = {{OpenAI}},
  title        = {{GPT-5.6}: Frontier Intelligence That Scales with Your Ambition},
  howpublished = {\url{https://openai.com/index/gpt-5-6/}},
  year         = {2026},
  month        = {July},
  note         = {Accessed: 2026-09-13}
}

@misc{chen2026cdhbench,
      title={CDH-Bench: A Commonsense-Driven Hallucination Benchmark for Evaluating Visual Fidelity in Vision-Language Models}, 
      author={Kesheng Chen and Yamin Hu and Qi Zhou and Zhenqian Zhu and Wenjian Luo},
      year={2026},
      eprint={2603.27982},
      archivePrefix={arXiv},
      primaryClass={cs.CV},
      url={https://arxiv.org/abs/2603.27982}, 
}

@InProceedings{Guan2024Hallusion,
    author    = {Guan, Tianrui and Liu, Fuxiao and Wu, Xiyang and Xian, Ruiqi and Li, Zongxia
                 and Liu, Xiaoyu and Wang, Xijun and Chen, Lichang and Huang, Furong
                 and Yacoob, Yaser and Manocha, Dinesh and Zhou, Tianyi},
    title     = {HallusionBench: An Advanced Diagnostic Suite for Entangled Language
                 Hallucination and Visual Illusion in Large Vision-Language Models},
    booktitle = {Proceedings of the IEEE/CVF Conference on Computer Vision and Pattern
                 Recognition (CVPR)},
    year      = {2024},
    pages     = {14375--14385}
}

@inproceedings{Yang2023visualprompting,
 author = {Yang, Lingfeng and Wang, Yueze and Li, Xiang and Wang, Xinlong and Yang, Jian},
 booktitle = {Advances in Neural Information Processing Systems},
 doi = {10.52202/075280-1086},
 editor = {A. Oh and T. Naumann and A. Globerson and K. Saenko and M. Hardt and S. Levine},
 pages = {24993--25006},
 publisher = {Curran Associates, Inc.},
 title = {Fine-Grained Visual Prompting},
 url = {https://proceedings.neurips.cc/paper_files/paper/2023/file/4e9fa6e716940a7cfc60c46e6f702f52-Paper-Conference.pdf},
 volume = {36},
 year = {2023}
}

@misc{carion2026sam3,
      title={SAM 3: Segment Anything with Concepts}, 
      author={Nicolas Carion and Laura Gustafson and Yuan-Ting Hu and Shoubhik Debnath and Ronghang Hu and Didac Suris and Chaitanya Ryali and Kalyan Vasudev Alwala and Haitham Khedr and Andrew Huang and Jie Lei and Tengyu Ma and Baishan Guo and Arpit Kalla and Markus Marks and Joseph Greer and Meng Wang and Peize Sun and Roman Rädle and Triantafyllos Afouras and Effrosyni Mavroudi and Katherine Xu and Tsung-Han Wu and Yu Zhou and Liliane Momeni and Rishi Hazra and Shuangrui Ding and Sagar Vaze and Francois Porcher and Feng Li and Siyuan Li and Aishwarya Kamath and Ho Kei Cheng and Piotr Dollár and Nikhila Ravi and Kate Saenko and Pengchuan Zhang and Christoph Feichtenhofer},
      year={2026},
      eprint={2511.16719},
      archivePrefix={arXiv},
      primaryClass={cs.CV},
      url={https://arxiv.org/abs/2511.16719}, 
}

@article{li2026seeing,
  title={Seeing or Knowing? Visual Context Sensitivity in Multimodal Large Language Models},
  author={Li, Jiaang and Li, Chengzu and An, Zhaochong and Yuan, Yifei and Liu, Xi and Belongie, Serge and Sn{\ae}bjarnarson, V{\'e}steinn},
  journal={arXiv preprint arXiv:2607.26326},
  year={2026}
}

@Book{kahneman2011thinking,
    author={Kahneman, Daniel},
    title={Thinking, fast and slow.},
    year={2011},
    publisher={Farrar, Straus and Giroux},
    address={New York, NY, US},
    pages={499-499},
}

@misc{anh2026countingchorevlms,
      title={Counting to Four is still a Chore for VLMs}, 
      author={Duy Le Dinh Anh and Patrick Amadeus Irawan and Tuan Van Vo},
      year={2026},
      eprint={2604.10039},
      archivePrefix={arXiv},
      primaryClass={cs.CV},
      url={https://arxiv.org/abs/2604.10039}, 
}
\bibliographystyle{iclr2027_conference}

\appendix
% \section{Appendix}
% You may include other additional sections here.
\clearpage
\begingroup
\let\clearpage\relax
\hypersetup{linkcolor=black}
\startcontents[appendix]
\printcontents[appendix]{l}{1}{\section*{Overview of the Appendix}\setcounter{tocdepth}{2}}
\endgroup
% \tableofcontents
\clearpage
\section{Additional Details on CounterCount Data}

\subsection{Dataset Statistics}
\label{sup:dataset_Sta}

\begin{table*}[!h]
  \centering
  \renewcommand{\arraystretch}{1.05}
  \caption{
   Statistics of the CounterCount dataset. The sample CF shows a representative CF instance (factual count\,${\to}$\,CF count). Total subjects and attribute types are counted as unique values across the whole dataset, so they are lower than the sum of the per-category columns.
  }
  \label{tab:dataset}
  \resizebox{\textwidth}{!}{
    \setlength{\tabcolsep}{10pt}
    \small
    \begin{tabular}{l r r r l}
      \toprule
      \textbf{Category}
        & \textbf{Pairs}
        & \textbf{Subjects}
        & \textbf{Attr.\ Types}
        & \textbf{Sample CF} \\
      \midrule
      Mammals              & 39 & 32 & 12 & giraffe (1 head ${\to}$ 5) \\
      Birds                & 31 & 27 &  6 & parrot (2 legs ${\to}$ 4) \\
      Crawling Creatures   & 31 & 19 & 15 & spider (8 legs ${\to}$ 4) \\
      Reptiles             & 30 & 21 & 12 & gecko (4 legs ${\to}$ 1) \\
      Sea Creatures        & 30 & 20 & 15 & narwhal (1 tusk ${\to}$ 3) \\
      Ancient Creatures    & 40 & 15 & 18 & mammoth (1 trunk ${\to}$ 2) \\
      Human               & 35 &  7 & 21 & hand (5 fingers ${\to}$ 7) \\
      Mythical Creatures   & 31 & 17 & 11 & unicorn (1 horn ${\to}$ 4) \\
      Fictional Characters & 32 & 18 & 26 & hello kitty (2 eyes ${\to}$ 1) \\
      Functional Obj.      & 30 & 29 & 25 & T-shirt (2 sleeves ${\to}$ 4) \\
      Housing              & 37 & 33 & 27 & cottage (1 chimney ${\to}$ 3) \\
      Transportation       & 32 & 27 & 26 & car (4 wheels ${\to}$ 6) \\
      Landmarks            & 30 & 23 & 24 & Fuji mountain (1 peak ${\to}$ 4) \\
      Food                 & 30 & 30 & 21 & pineapple (1 crown ${\to}$ 3) \\
      Flags                & 35 & 18 & 21 & Japan flag (1 red circle ${\to}$ 2) \\
      Logos                & 31 & 20 & 25 & Starbucks logo (1 crown ${\to}$ 2) \\
      Currency Sym.        & 30 & 23 &  9 & naira (2 lines ${\to}$ 3) \\
      \midrule
      \textbf{Total}  & \textbf{554} & \textbf{379} & \textbf{228} & \\
      \bottomrule
    \end{tabular}
  }
\end{table*}

\subsection{Dataset Comparison}
\label{sup:dataset_Comp}
\begin{table*}[!h]
\centering
\caption{Comparison of CounterCount with prior counterfactual, hallucination, and counting-oriented
benchmarks. All CounterCount images are human-verified. \textsuperscript{*}Counts estimated from
reported image totals; exact numbers not confirmed in the source papers.
\textsuperscript{**}Unique images only; the original benchmark's totals also include
multiple resolutions of each image.}
\scriptsize
\setlength{\tabcolsep}{4pt}
\renewcommand{\arraystretch}{1.05}
\resizebox{\textwidth}{!}{%
\begin{tabular}{l c c c c c c l c}
\toprule
\textbf{Benchmark} &
\makecell{\textbf{Imgs /}\\\textbf{Pairs}} &
\makecell{\textbf{Local.}\\\textbf{Annot.}} &
\makecell{\textbf{Gen.}\\\textbf{Method}} &
\makecell{\textbf{Question}\\\textbf{Type}} &
\makecell{\textbf{Subj. /}\\\textbf{Attr.}} &
\textbf{Cat.} &
\makecell[l]{\textbf{Conflict Type}} &
\makecell{\textbf{Count.}\\\textbf{Task}} \\
\midrule
VLind-Bench~\citep{lee2025vlind}
& 2576 / $\sim$1288\textsuperscript{*} & \xmark & Auto & Binary & --- & ---
& Language prior & \xmark \\
ViLP~\citep{Luo2025probing}
& 900 / --- & \xmark & Auto & Open & --- & ---
& Visual language prior & \xmark \\
\midrule
VLMs are Biased~\citep{vo2025vision}
& $\sim$464\textsuperscript{**} / --- & \xmark & Semi-auto & Open/Binary & --- / $\sim$19\textsuperscript{*} & 7
& World knowledge & \cmark \\
\midrule
HallusionBench~\citep{Guan2024Hallusion}
& 181 / 165 & \xmark & Manual & Binary & --- & ---
& Lang.\ prior + illusion & \xmark \\
MMVP~\citep{Tong2024Eyes}
& 300 / 150 & \xmark & Manual & MCQ/Binary & --- & ---
& CLIP-blind perception & \xmark \\
LanP~\citep{wu2025lanp}
& 170 / --- & \xmark & Manual & Binary & --- & 4
& Visual ambiguity & Partial \\
CDH-Bench~\citep{chen2026cdhbench}
& 600 / 300 & \xmark & Manual & MCQ/Binary & 66 / 60 & 3
& Object-level knowledge & Partial \\
\midrule
\textbf{CounterCount (ours)}
& \textbf{1003 / 554} & \textbf{Mask+BB} & \textbf{Manual} & \textbf{MCQ/Open} & \textbf{379 / 228} & \textbf{17}
& \textbf{Object-level knowledge} & \cmark \\
\bottomrule
\end{tabular}%
}

\label{tab:dataset_comparisonfinal}
\vspace{-14pt}
\end{table*}

\newpage
\subsection{Factual–Counterfactual pairs}
\label{sup:pairs}

\begin{figure}[h]
    \centering
    \includegraphics[width=1\linewidth]{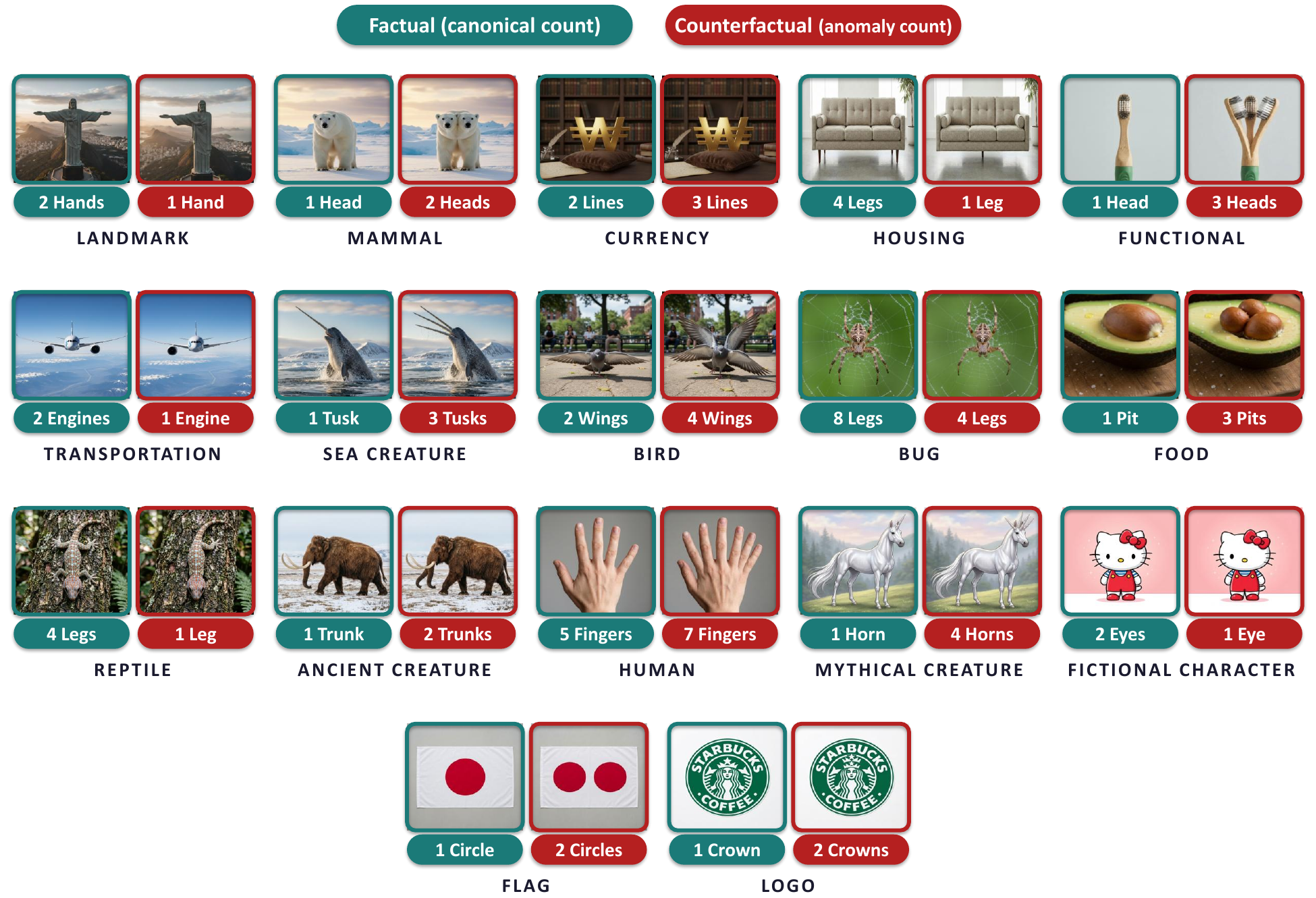}
    \caption{\textbf{Representative factual–counterfactual pairs from CounterCount,
  spanning all 17 semantic categories.}
  Each column pair shows one instance: \textcolor{teal!60!black}{\textbf{left image}}
  (teal border) shows the
  corresponding factual image with canonical counts intact, while \textcolor{red!75!black}{\textbf{right image}}
  (red border) is the counterfactual in which a canonical attribute count
  is deliberately violated.}
 \label{fig:pair}
\end{figure}

\Cref{fig:pair} shows representative factual and counterfactual examples drawn from across the
seventeen categories in CounterCount. Each pair preserves viewpoint, lighting, background, and
object identity, and differs only in the count of a single structural attribute,
illustrating the range of canonical priors the dataset places in conflict with
visual evidence. 

% \newpage
\section{Details on Prompt Design}
\subsection{Distractor Generation for MCQs}
\label{sub:template}
\begin{table}[H]
\centering
\small
\caption{Examples of distractor generation for MCQ options. $c_o$ and $c_a$ denote the \emph{ordinary} and \emph{anomalous} counts, respectively.}
\label{tab:distractor_examples}
\resizebox{0.75\textwidth}{!}{%
\begin{tabular}{cccccl}
\toprule
$c_o$ & $c_a$ & Gap & $d_1$ & $d_2$ & Generated Options \\
\midrule
2 & 4 & 2 & midpoint (3) & outside (5) & \{2, 3, 4, 5\} \\
2 & 7 & 5 & midpoint (4) & outside (8) & \{2, 4, 7, 8\} \\
7 & 2 & 5 & midpoint (4) & outside (8) & \{2, 4, 7, 8\} \\
2 & 3 & 1 & below min (1) & outside (4) & \{1, 2, 3, 4\} \\
1 & 2 & 1 & outside (3) & fallback (4) & \{1, 2, 3, 4\} \\
3 & 0 & 3 & midpoint (1) & outside (4) & \{0, 1, 3, 4\} \\
1 & 0 & 1 & outside (2) & fallback (3) & \{0, 1, 2, 3\} \\
\bottomrule
\end{tabular}
}
\end{table}

Each multiple-choice question consists of four options: the correct anomalous count, the ordinary count, and two distractors. Distractors are generated programmatically based on the relationship between the ordinary count $c_o$ and the anomalous count $c_a$.

When the gap between the two counts is greater than one, the first distractor is set to the midpoint $\lfloor(c_o + c_a) / 2\rfloor$, providing a plausible in-between value. The second distractor is set to one above the maximum of the two counts, serving as an out-of-range option. When the gap equals one, no meaningful midpoint exists, so the first distractor is instead set to one below the minimum of the two counts. If this yields an invalid value (i.e., zero or below), a fallback assigns the distractor to two above the maximum. This fallback also applies when $c_a = 0$, where the minimum is already zero and no smaller valid count exists. In all cases, duplicate options are resolved by incrementing the conflicting value until a unique candidate is found. Finally, all options are converted to word form (e.g., \textit{three}) and shuffled, to prevent position bias, with a fixed per-image random seed, to ensure reproducibility. Table~\ref{tab:distractor_examples} illustrates the distractor generation strategy across representative cases.

\subsection{VLM-as-a-Judge for Answer Extraction}

We employ a VLM-as-a-judge approach to extract the final numeric answer from each response when multiple numbers are present. Specifically, we prompt the judge model with the following:

\begin{quote}
\textit{Question: \{question\}\\
Model response: `\{response\}'\\
The correct answer to this question is \{correct\_answer\}. \\
What single number did the model give as its final answer to the question? Reply with only a single digit, nothing else.}
\end{quote}

The judge is provided with the original question, the full model response, and the ground-truth answer as context, and is asked to identify the single digit the evaluated model committed to as its final answer. This allows us to determine whether the model response was correct, biased, or a miscount.

\subsection{Open-Ended and MCQ Prompts}

Each image is evaluated under two prompting formats: OE and MCQ. Both formats share the same base question, which is constructed from the instance metadata using the object name and anomalous part:

\begin{quote}
\textit{How many \{anomaly\}s does this \{name\} have?}
\end{quote}

For example, for a parrot with an anomalous number of legs, the base question becomes: \textit{How many legs does this parrot have?}

In the OE format, the model is prompted to complete a sentence with just the count and the name of the part:

\begin{quote}
\textit{How many \{anomaly\}s does this \{name\} have? Complete the following sentence with just the count and the name of the part: The \{name\} has}
\end{quote}

In the MCQ format, four options are appended to the base question, and the model is instructed to select one:

\begin{quote}
\textit{How many \{anomaly\}s does this \{name\} have? Choose one of the following options: \{options\}. Reply with only one word from the given options and nothing else.}
\end{quote}

The MCQ options consist of the correct anomalous count, the ordinary count, and two distractors generated as described in \ref{sub:template}.
\section{Additional Experimental Results}
\label{sup:experiments}

\subsection{Multiple-Choice Baseline Results}
\label{sup:mcq_res}

\begin{figure}[htpb]
    \centering
    \includegraphics[width=0.6\linewidth]{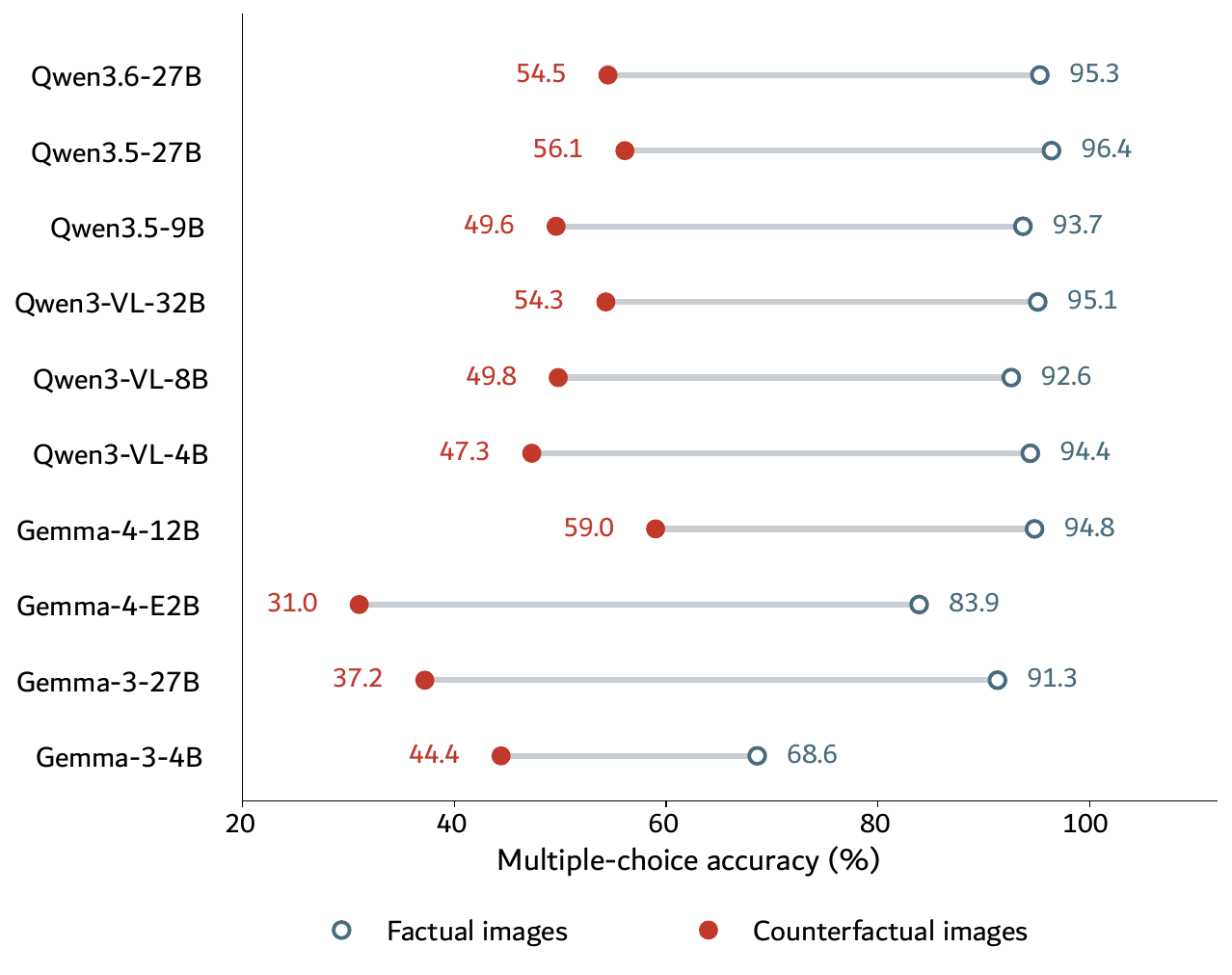}
    \caption{
    MCQ accuracy on factual and CF images.
    }
    \label{sup:mcq_fig_base}
\end{figure}

\Cref{sup:mcq_fig_base} shows baseline MCQ results for the models not included in the main paper.

\subsection{Open-ended Baseline Results}
\label{sup:oe_res}
\begin{figure}[htpb]
    \centering
    \includegraphics[width=0.7\linewidth]{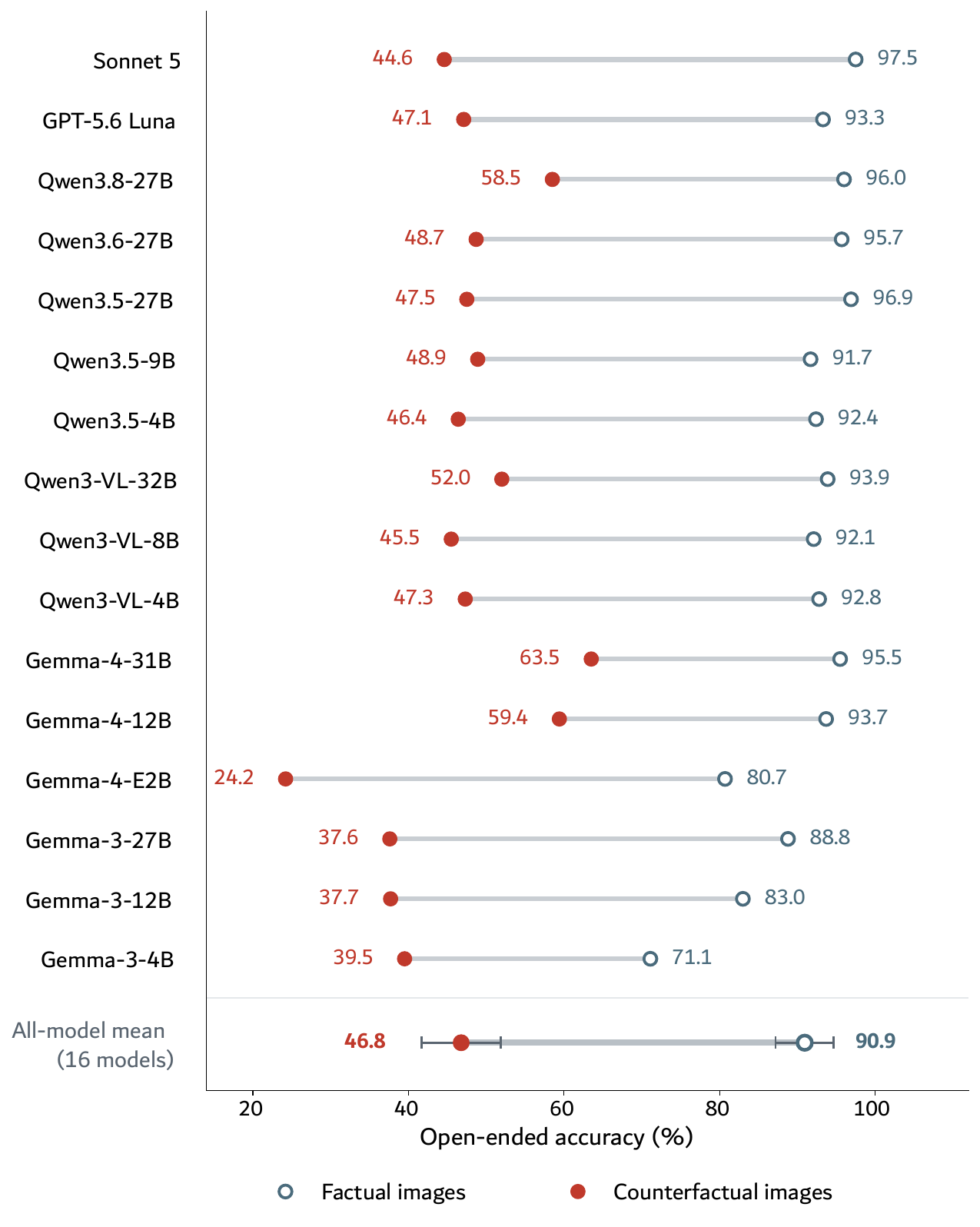}
    \caption{
    OE accuracy on factual and CF images.
    }
    \label{sup:oe_fig_base}
\end{figure}

\Cref{sup:oe_fig_base} shows baseline OE results for all evaluated models.

\clearpage

\subsection{Categories Results}
\label{sup:cat_res}
\begin{figure}[htpb]
    \centering
    \includegraphics[width=0.8\linewidth]{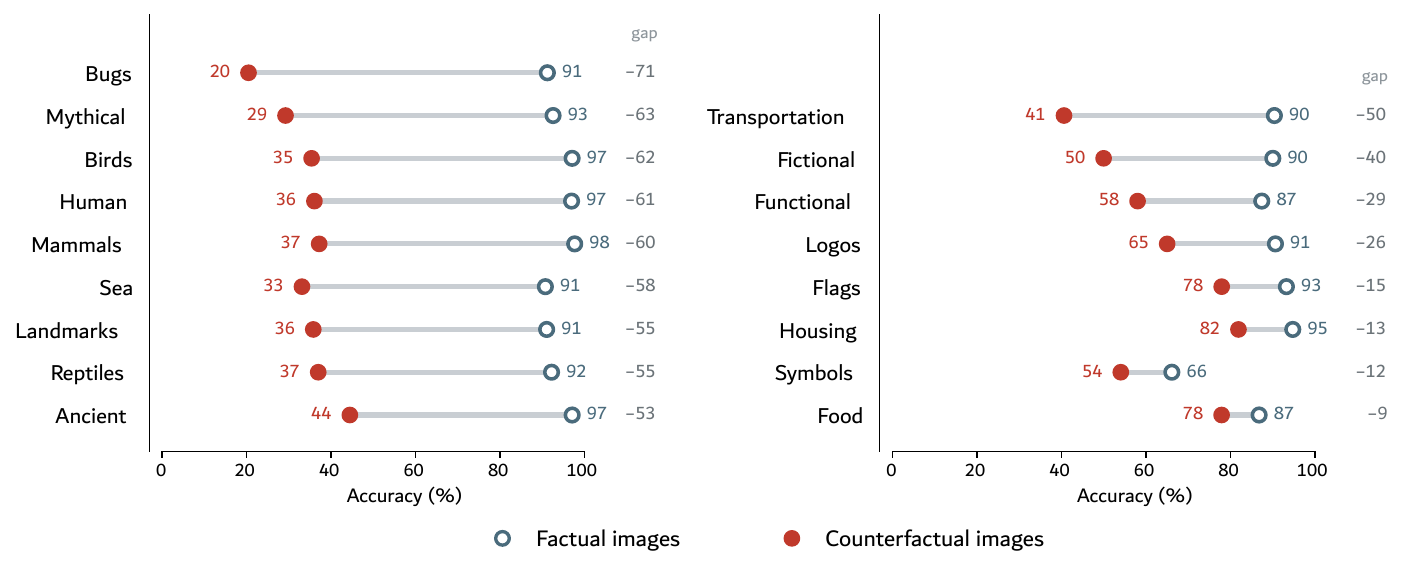}
    \caption{\textbf{Accuracy by category.} Factual and CF accuracy per category, averaged over MCQ and OE.}
    \label{fig:cat_accuracy}
\end{figure}

\begin{figure}[htpb]
    \centering
    \includegraphics[width=0.8\linewidth]{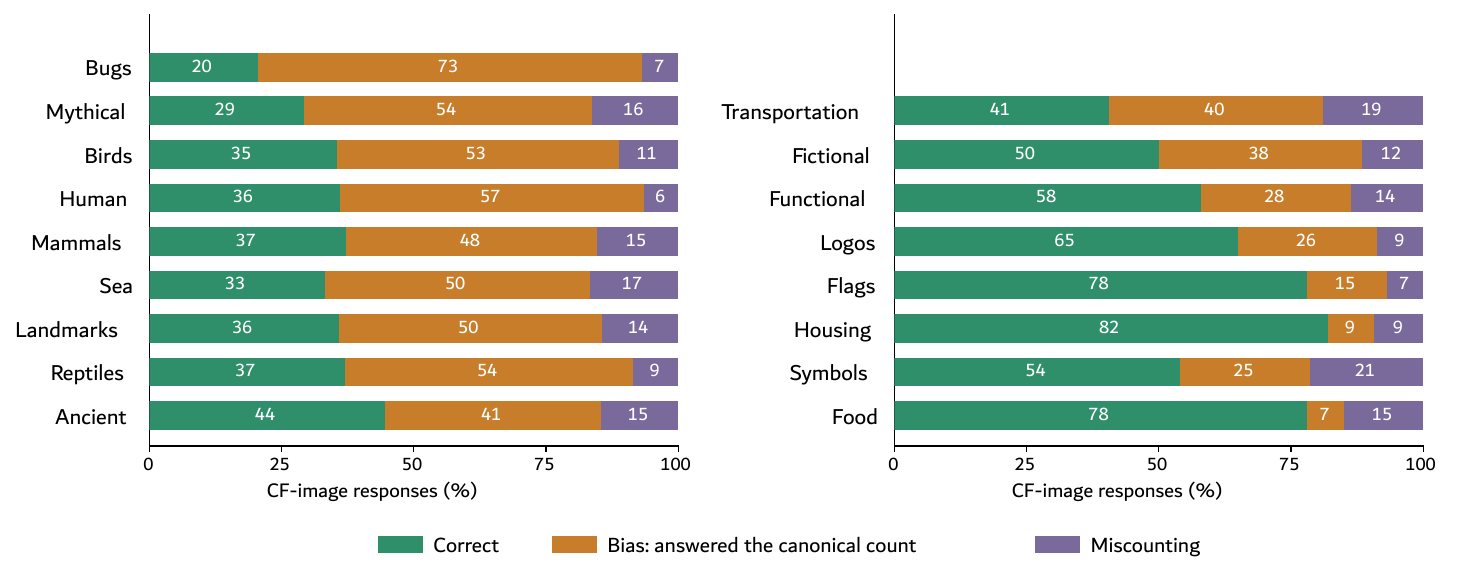}
    \caption{\textbf{Prediction breakdown by category.} Share of correct, biased, and other-error predictions on CF images per category, averaged over MCQ and OE.}
    \label{fig:cat_composition}
\end{figure}

\Cref{fig:cat_accuracy} reports factual and CF accuracy per category, and \Cref{fig:cat_composition} breaks down CF predictions into correct, biased, and other errors.

\subsection {Generic vs. Specific subjects}
\label{sup:phrasing}
\begin{figure}[htpb]
    \centering
    \includegraphics[width=0.7\linewidth]{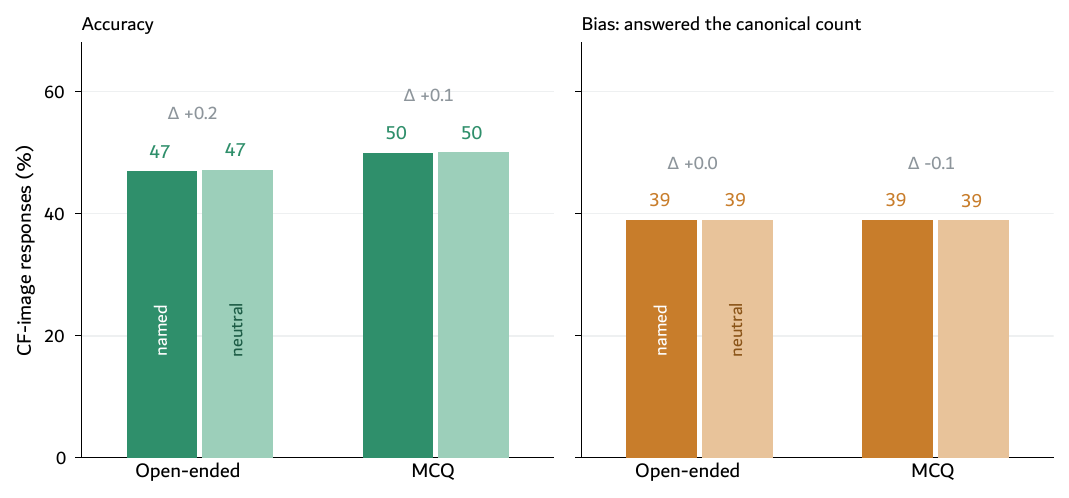}
    \caption{\textbf{Effect of subject naming.} CF accuracy (left) and bias rate (right) with specific and generic subject names, averaged over all models, for OE and MCQ.}
    \label{fig:phrasing}
\end{figure}

To test whether naming the subject triggers the prior, we replace each specific subject name with a generic term assigned per category, leaving the rest of the question unchanged (\eg ``this parrot'' $\rightarrow$ ``this animal'').
We use \textit{animal} for Ancient Creatures, Birds, Bugs, Mammals, Reptiles, and Sea Creatures; \textit{object} for Functional, Housing, and Transportation; \textit{character} for Fictional Characters; \textit{creature} for Mythical Creatures; and the singular category name for the remaining categories (\textit{flag}, \textit{food}, \textit{human}, \textit{landmark}, \textit{logo}, and \textit{symbol}).
As shown in \Cref{fig:phrasing}, CF accuracy and bias remain unchanged in both formats.

% \clearpage

\subsection{Blurring Results}
\label{sup:blur}

\subsubsection{MCQ Results}
\begin{figure}[htpb]
    \centering
    \includegraphics[width=0.6\linewidth]{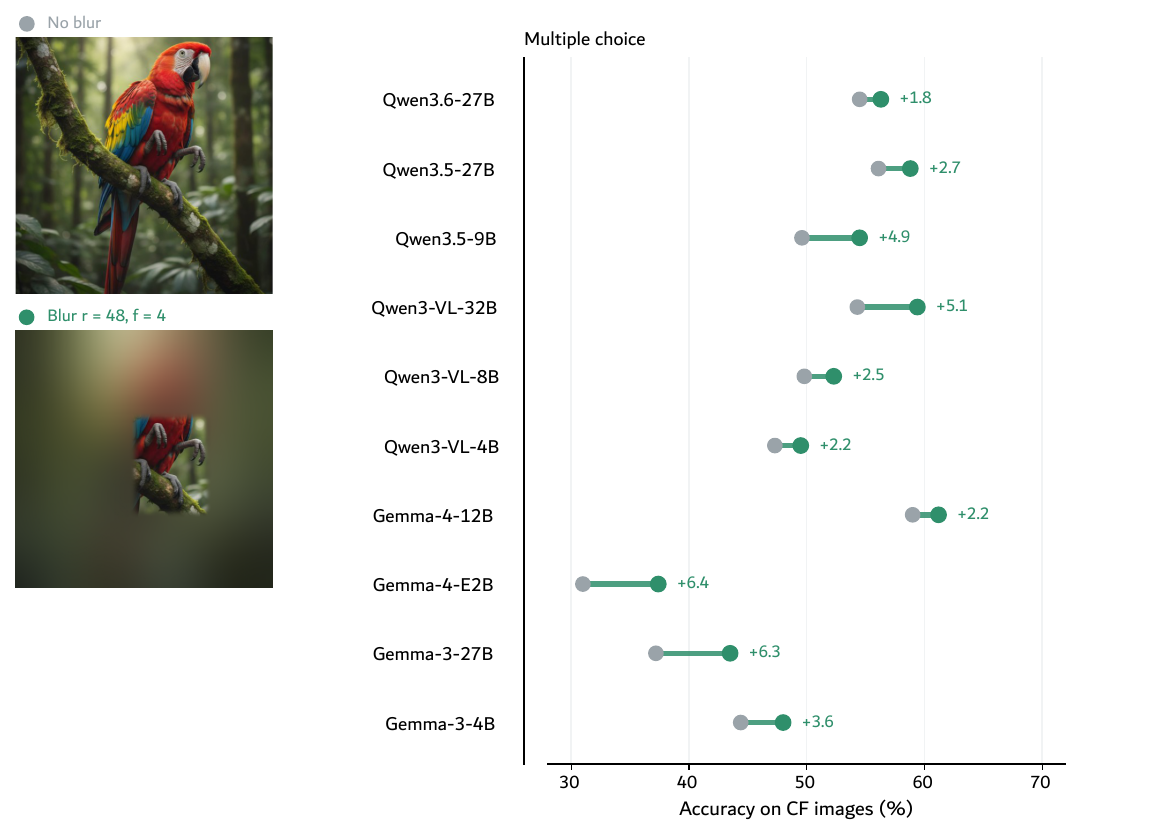}
    \caption{\textbf{Blurring (MCQ).} CF accuracy with and without blurring.}
    \label{fig:blur_mcq_rest}
\end{figure}

\Cref{fig:blur_mcq_rest} shows MCQ results for the models not included in the main paper.

\subsubsection{OE Results}
\begin{figure}[htpb]
    \centering
    \includegraphics[width=0.6\linewidth]{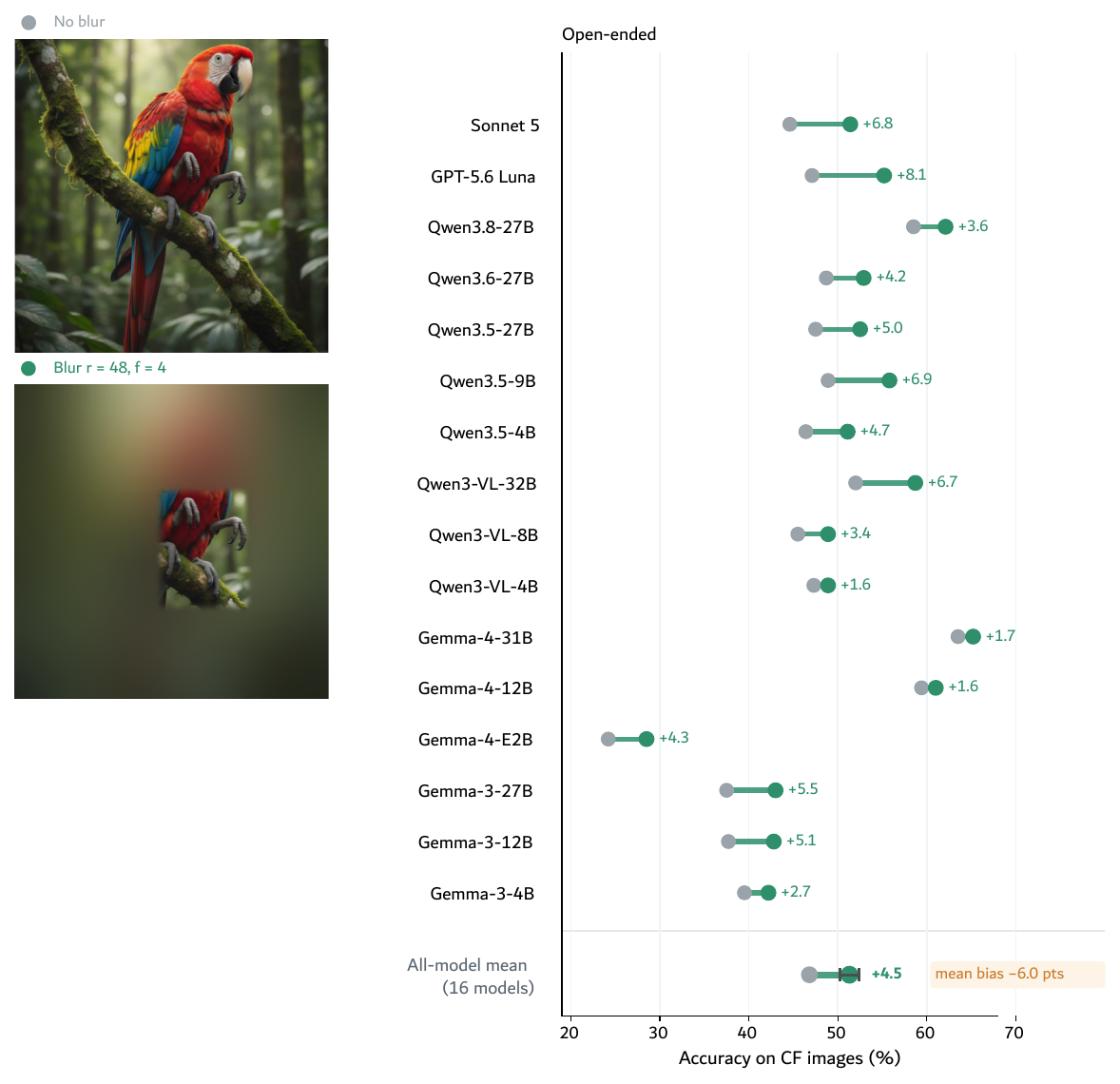}
    \caption{\textbf{Blurring (OE).} CF accuracy with and without blurring.}
    \label{fig:blur_oe_all}
\end{figure}

\Cref{fig:blur_oe_all} shows OE results for all evaluated models.

\subsubsection{Blurring Sweep}
\begin{figure}[htpb]
    \centering
    \includegraphics[width=0.7\linewidth]{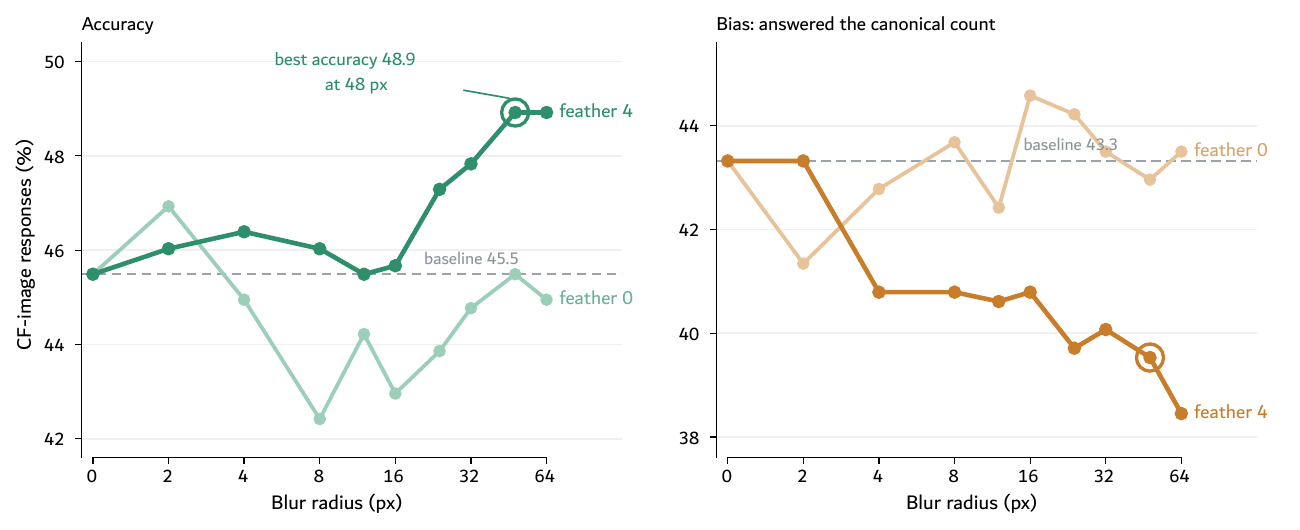}
    \caption{\textbf{Blur parameter sweep on Qwen3-VL-8B.} CF accuracy and bias rate, averaged over MCQ and OE, across blur radii and feathering values.}
    \label{fig:blur_sweep_fig}
\end{figure}

\Cref{fig:blur_sweep_fig} shows CF accuracy and bias rate, averaged over MCQ and OE, across blur radii and feathering values on Qwen3-VL-8B.
We select the setting with the highest accuracy (radius 48, feathering 4).
Larger radii slightly reduce bias but do not improve accuracy, and since the sweep is performed on a single model, we avoid stronger blurring that may degrade performance on other models.

\subsection{Instruction Results}
\label{sup:instruction}

\subsubsection{Humans vs.\ Models}
\label{sup:instruction:humanvsmodels}
\begin{figure}[H]
    \centering
    \includegraphics[width=0.6\linewidth]{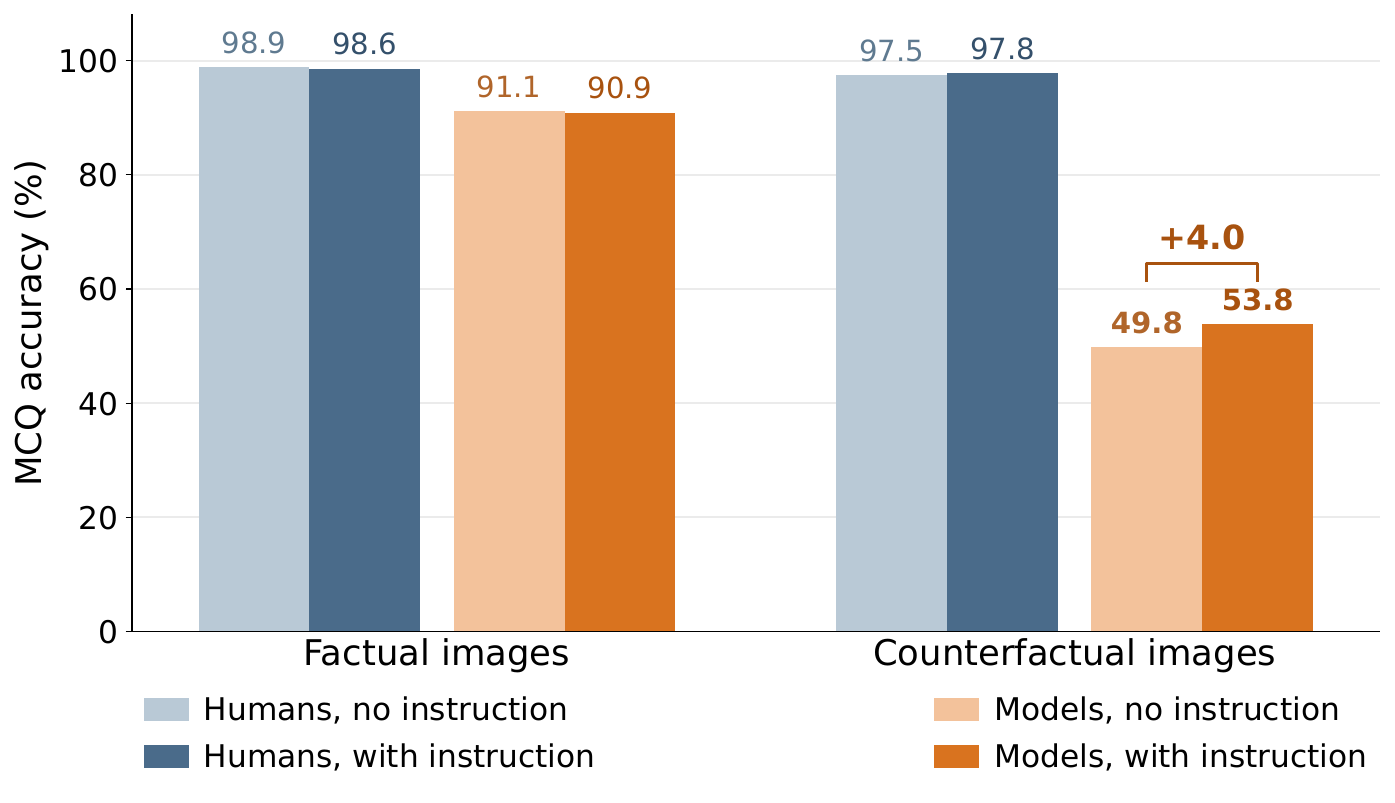}
    \caption{\textbf{Humans vs.\ models with cueing.} MCQ accuracy on factual and CF images with and without the instruction.}
    \label{fig:inst_human_model}
\end{figure}

\Cref{fig:inst_human_model} compares humans and models on factual and CF images with and without the instruction.
On factual images, the instruction has little effect for either, while on CF images, models improve and humans remain near the ceiling.

\subsubsection{MCQ Results}
\begin{figure}[htpb]
    \centering
    \includegraphics[width=0.7\linewidth]{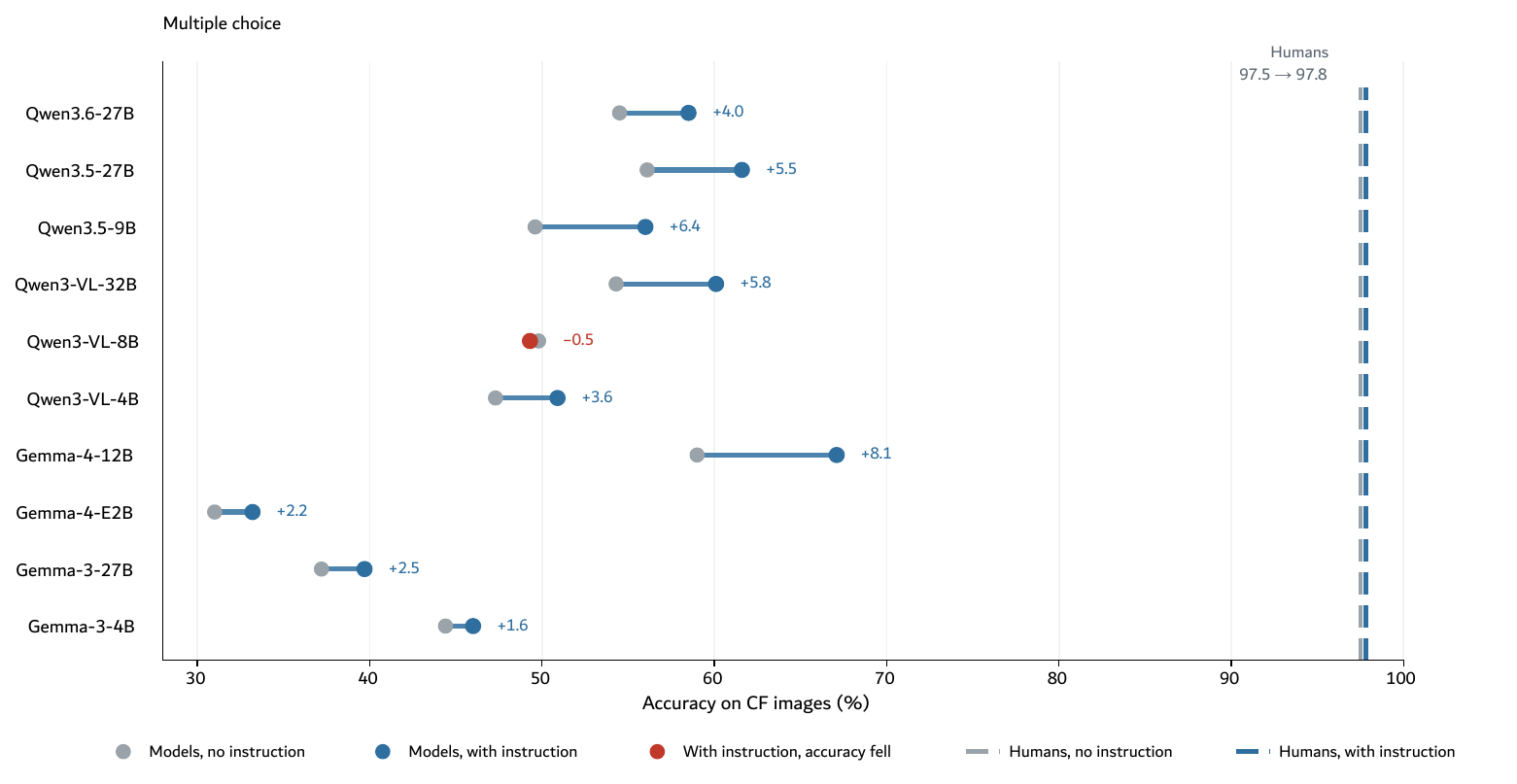}
    \caption{\textbf{Cueing (MCQ).} CF accuracy with and without the instruction for models not shown in \Cref{fig:inst_fig}, compared with humans.}
    \label{fig:inst_mcq_rest}
\end{figure}

\Cref{fig:inst_mcq_rest} shows MCQ results for models not included in the main paper.

\subsubsection{OE Results}
\begin{figure}[H]
    \centering
    \includegraphics[width=0.7\linewidth]{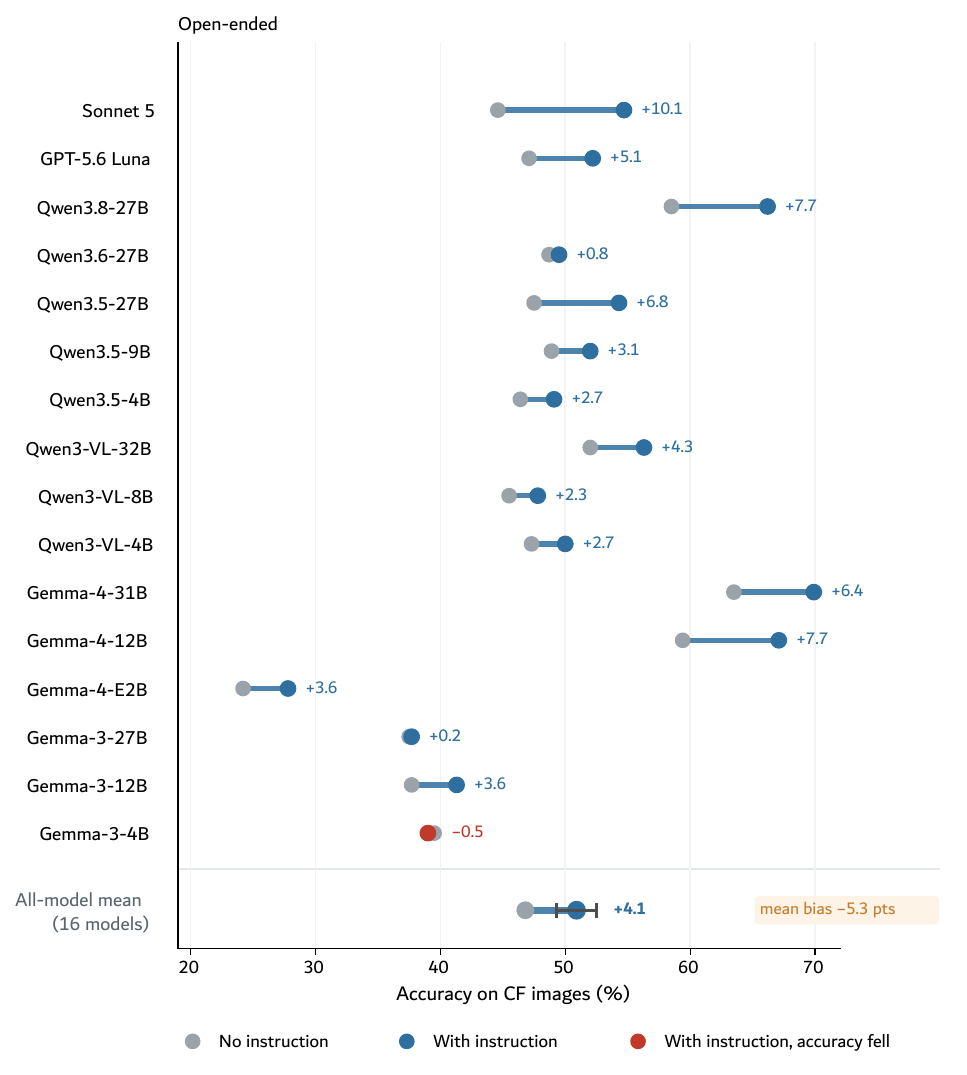}
    \caption{\textbf{Cueing (OE).} CF accuracy with and without the instruction for all models.}
    \label{fig:inst_oe_all}
\end{figure}

\Cref{fig:inst_oe_all} shows OE results for all models.

\subsection{Thinking}
\label{sup:thinking}

\subsubsection{OE Results}
\begin{figure}[H]
    \centering
    \includegraphics[width=0.7\linewidth]{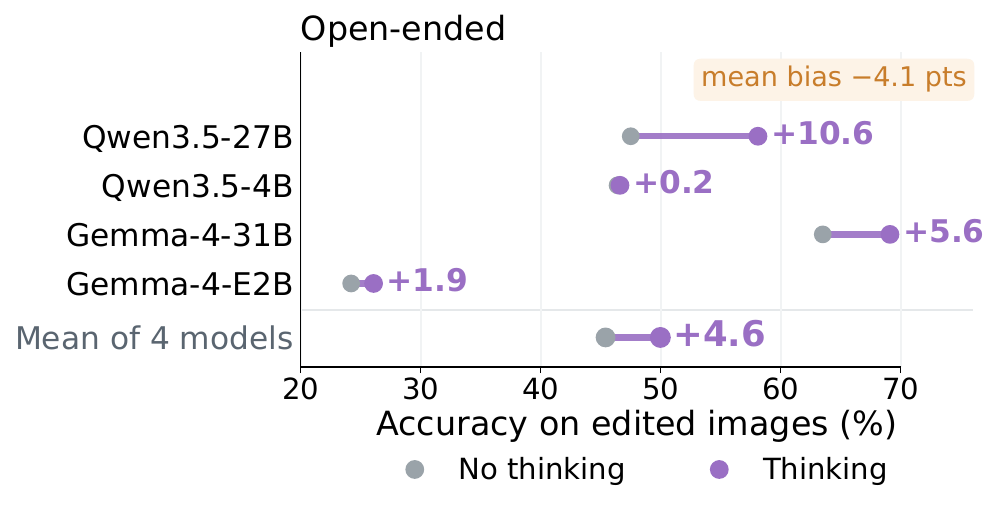}
    \caption{\textbf{Thinking (OE).} CF accuracy with and without thinking mode for Qwen3.5 (4B, 27B) and Gemma-4 (E2B, 31B).}
    \label{fig:think_oe}
\end{figure}

\Cref{fig:think_oe} shows OE results with and without thinking for the four models.

\subsubsection{Thinking Mode Decoding Details}
\label{sup:thinking:decoding}
We enable thinking mode through each model's chat template and keep the prompt unchanged from the baseline.
Generation uses sampling with a maximum budget of 16,384 tokens and each model's default sampling parameters (\Cref{tab:thinking_params}).
Qwen3.5-4B does not ship a generation configuration and therefore uses the library defaults, which are more diffuse than those of Qwen3.5-27B.
After generation, we remove the reasoning block and extract the answer from the remaining response, following the procedure in \Cref{sec:results}.
Since decoding is stochastic and results come from a single run, per-model numbers may vary across runs.

\begin{table}[htpb]
\centering
\small
\caption{\textbf{Sampling parameters for thinking mode.}}
\label{tab:thinking_params}
\begin{tabular}{lccc}
\toprule
\textbf{Model} & \textbf{Temperature} & \textbf{Top-$k$} & \textbf{Top-$p$} \\
\midrule
Qwen3.5-4B$^\dagger$ & 1.0 & 50 & 1.0 \\
Qwen3.5-27B & 0.6 & 20 & 0.95 \\
Gemma-4-E2B & 1.0 & 64 & 0.95 \\
Gemma-4-31B & 1.0 & 64 & 0.95 \\
\bottomrule
\multicolumn{4}{l}{\footnotesize $^\dagger$No generation configuration provided; library defaults used.}
\end{tabular}
\end{table}

\subsection{Attention}
\label{sup:attention}

\subsubsection{Attention Analysis}
\label{sup:attention:analysis}
\begin{figure}[H]
    \centering
    \includegraphics[width=0.7\linewidth]{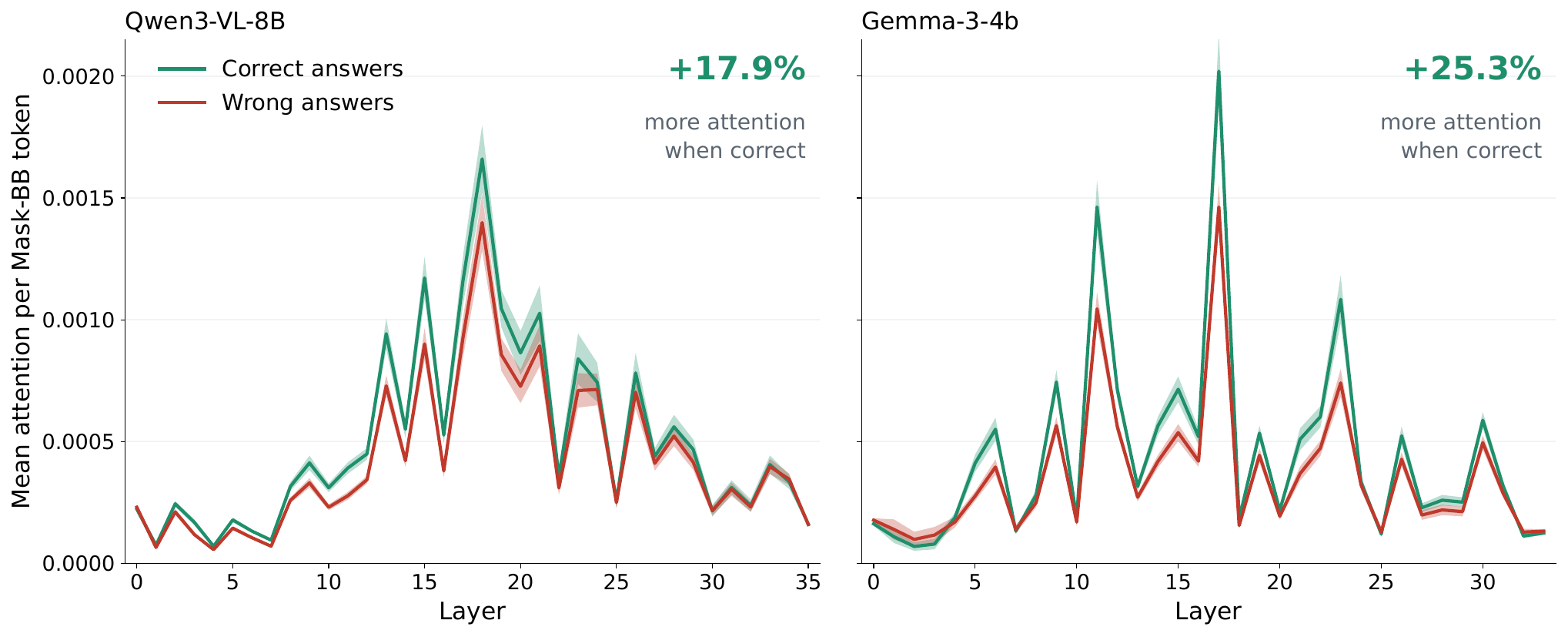}
    \caption{\textbf{Attention to count-relevant regions (OE).} Mean attention per Mask-BB token across layers for correct and incorrect OE answers on CF images.}
    \label{fig:corwro_oe}
\end{figure}

For OE, where the answer spans multiple tokens, we follow the procedure in \Cref{sec:experiments:info} and additionally average attention over the generated answer tokens.
As shown in \Cref{fig:corwro_oe}, correct answers similarly receive more attention to count-relevant regions than incorrect ones.

\subsubsection{Attention Modulation}
\label{sup:attention:modulation}

\begin{figure}[H]
    \centering
    \includegraphics[width=0.75\linewidth]{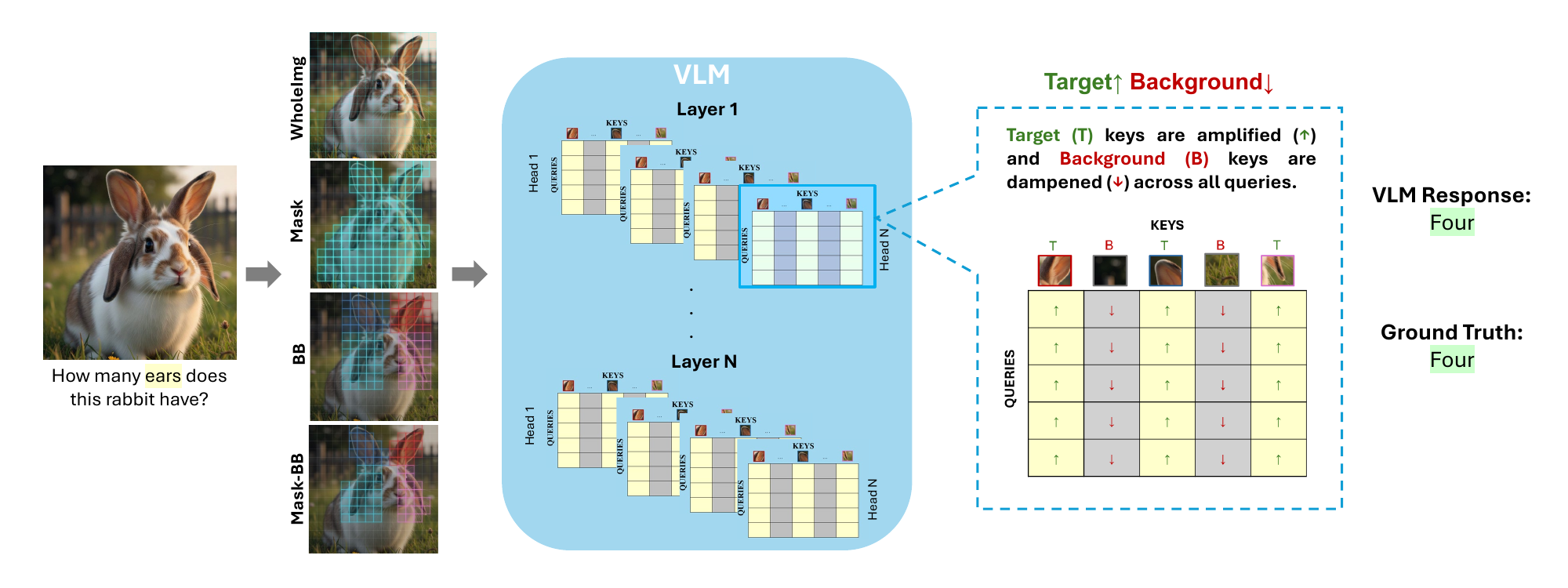}
    \caption{
        \textbf{Inference-time attention modulation.}
        The formulation supports modulating attention toward different visual-token regions (\textit{WholeImg}, \textit{Mask}, \textit{BB}, and \textit{Mask-BB}); we use Mask-BB throughout.
        We illustrate the T$\uparrow$B$\downarrow$ setting, which amplifies count-relevant target tokens while suppressing tokens outside the target.
    }
    \label{fig:methodology}
\end{figure}

The additive logit adjustment in \Cref{eq:logit_shift} corresponds to rescaling attention weights.
Let $z_{h,i,j}$ denote the attention logit for head $h$ from query position $i$ to key position $j$, with standard attention weights
\begin{equation}
    A_{h,i,j} = \frac{\exp(z_{h,i,j})}{\sum_{k=1}^{n}\exp(z_{h,i,k})}.
\end{equation}
Let $\Gamma \in \mathbb{R}^n$ with $\Gamma_k = \alpha$ if key $k$ is targeted and $\Gamma_k = 1$ otherwise.
Scaling the exponentiated logit at a selected position $j$ yields
\begin{equation}
    \tilde{A}_{h,i,j}
    =
    \frac{\alpha\,\exp(z_{h,i,j})}{\sum_{k=1}^{n} \Gamma_k\,\exp(z_{h,i,k})}
    =
    \frac{\exp(z_{h,i,j}+\log\alpha)}{\sum_{k=1}^{n} \exp(z_{h,i,k}+\log \Gamma_k)}.
\end{equation}
Thus, amplification ($\alpha > 1$), suppression ($0 < \alpha < 1$), masking ($\alpha = 0$), and identity ($\alpha = 1$) are special cases of the same operation.
When $\alpha$ is applied uniformly across all keys, it cancels in the softmax and leaves the attention distribution unchanged, so the modulation takes effect only when applied selectively.
Given CounterCount's annotations, the target can be the whole image, the subject mask, the bounding boxes, or their overlap (Mask-BB); we use Mask-BB throughout.

\subsubsection{Selected Configurations}
\label{sup:attention:configs}
\Cref{tab:att_configs} lists the configuration selected for each model from the 12 described in \Cref{sec:experiments:info}, along with the resulting gain in CF accuracy.
All 14 models improve in both formats.
Qwen models mostly favor amplification with outside masking, while Gemma models mostly favor amplification alone, suggesting that the optimal mode depends on the model family.

\begin{table}[H]
\centering
\small
\caption{\textbf{Selected attention modulation configuration per model.} Amp.: target amplification alone ($\beta=1$); Amp.+Mask: target amplification with outside masking ($\beta=0$). Gain: change in CF accuracy (points).}
\label{tab:att_configs}
\begin{tabular}{l ccc c ccc c}
\toprule
& \multicolumn{4}{c}{\textbf{MCQ}} & \multicolumn{4}{c}{\textbf{OE}} \\
\cmidrule(lr){2-5} \cmidrule(lr){6-9}
\textbf{Model} & Mode & $\alpha$ & Heads & Gain & Mode & $\alpha$ & Heads & Gain \\
\midrule
Qwen3-VL-4B   & Amp.      & 3 & 10\% & +5.05 & Amp.      & 3 & 7\%  & +2.35 \\
Qwen3-VL-8B   & Amp.+Mask & 3 & 7\%  & +5.78 & Amp.+Mask & 3 & 5\%  & +5.42 \\
Qwen3-VL-32B  & Amp.+Mask & 3 & 5\%  & +5.23 & Amp.+Mask & 2 & 7\%  & +5.60 \\
Qwen3.5-4B    & Amp.+Mask & 2 & 7\%  & +2.53 & Amp.+Mask & 2 & 7\%  & +3.25 \\
Qwen3.5-9B    & Amp.+Mask & 3 & 5\%  & +4.51 & Amp.+Mask & 2 & 5\%  & +2.89 \\
Qwen3.5-27B   & Amp.+Mask & 3 & 10\% & +2.89 & Amp.+Mask & 2 & 10\% & +2.53 \\
Qwen3.6-27B   & Amp.      & 3 & 7\%  & +1.81 & Amp.+Mask & 2 & 5\%  & +1.26 \\
Qwen3.8-27B   & Amp.+Mask & 3 & 7\%  & +2.53 & Amp.+Mask & 2 & 10\% & +1.81 \\
\midrule
Gemma-3-4B    & Amp.      & 3 & 10\% & +1.99 & Amp.      & 2 & 7\%  & +2.35 \\
Gemma-3-12B   & Amp.      & 2 & 5\%  & +2.35 & Amp.      & 3 & 10\% & +2.89 \\
Gemma-3-27B   & Amp.+Mask & 2 & 10\% & +3.79 & Amp.+Mask & 2 & 7\%  & +1.81 \\
Gemma-4-E2B   & Amp.      & 3 & 5\%  & +5.23 & Amp.      & 3 & 5\%  & +2.53 \\
Gemma-4-12B   & Amp.      & 3 & 5\%  & +1.99 & Amp.      & 2 & 5\%  & +2.53 \\
Gemma-4-31B   & Amp.      & 2 & 7\%  & +0.90 & Amp.      & 2 & 10\% & +0.72 \\
\midrule
\textbf{Mean} & & & & \textbf{+3.33} & & & & \textbf{+2.71} \\
\bottomrule
\end{tabular}
\end{table}

\subsubsection{MCQ Results}
\begin{figure}[H]
    \centering
    \includegraphics[width=0.6\linewidth]{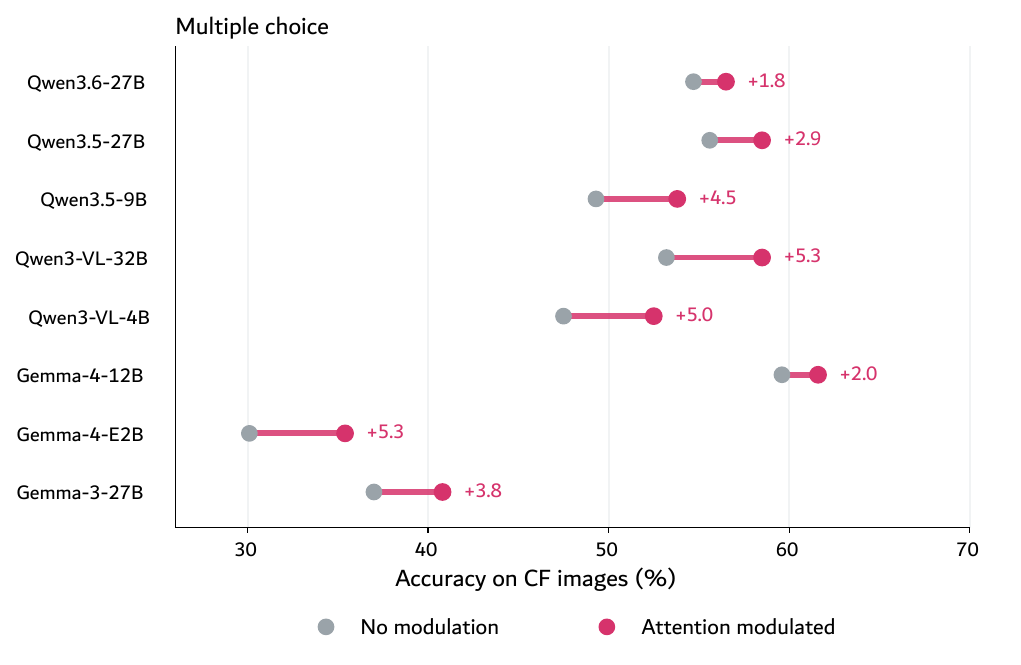}
    \caption{\textbf{Attention modulation (MCQ).} CF accuracy before and after modulation.}
    \label{fig:att_mcq_rest}
\end{figure}

\Cref{fig:att_mcq_rest} shows MCQ results for the eight open-weight models not included in the main paper.

\subsubsection{OE Results}
\label{sup:attention:oe}
\begin{figure}[H]
    \centering
    \includegraphics[width=0.6\linewidth]{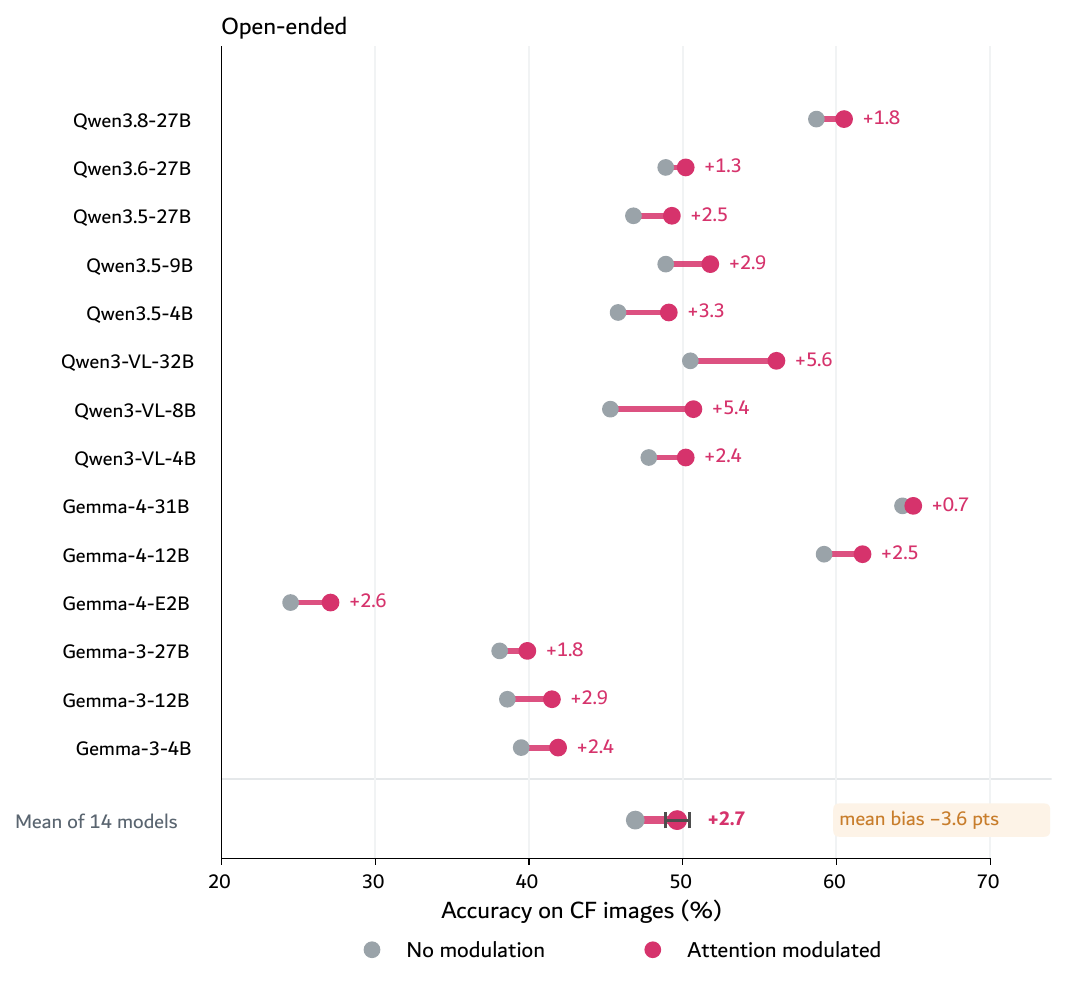}
    \caption{\textbf{Attention modulation (OE).} CF accuracy before and after modulation.}
    \label{fig:att_oe_all}
\end{figure}

\Cref{fig:att_oe_all} shows OE results for all 14 open-weight models.

% \clearpage
\section{Human Study Details}
\label{sup:human_study}

% This appendix provides the full protocol for the human study reported in the main paper.
 
\subsection{Platform and Audience Construction}
\label{app:human_platform}
 
All human annotations were collected through Rapidata\footnote{\url{https://www.rapidata.ai/}}, a high-throughput crowdsourcing platform. It displays each image for at most 25 seconds.
Rather than drawing from Rapidata's open global pool, we constructed a \emph{qualified custom audience} so that careless or inattentive annotators were screened out before the real counting task.

\myparagraph{Qualification trial.}
Each candidate annotator was presented with a 10-image validation set consisting of 4 factual images (canonical counts) and 6 CF images (anomalous counts), all with known ground-truth answers.
Questions were posed in the same four-option multiple-choice (MCQ) format used in the main evaluation.
To be an annotator, a participant had to solve at least 8 of these 10 trial questions at $\geq$87.5\% accuracy (\texttt{min\_tasks}=8, \texttt{target\_accuracy}=0.875).

\myparagraph{Threshold justification.}
We deliberately set the annotator threshold at 0.875 rather than requiring a perfect score of 1.0.
A perfect threshold would have wrongly rejected approximately one-fifth of otherwise excellent annotators who happened to make a single error over eight trials.
The sub-perfect threshold therefore balances annotator quality against pool size, retaining strong counters while filtering out genuinely careless respondents.

\subsection{Validation Set}
\label{app:validation_set}
 
\begin{figure}[ht]
\centering
\includegraphics[width=\textwidth]{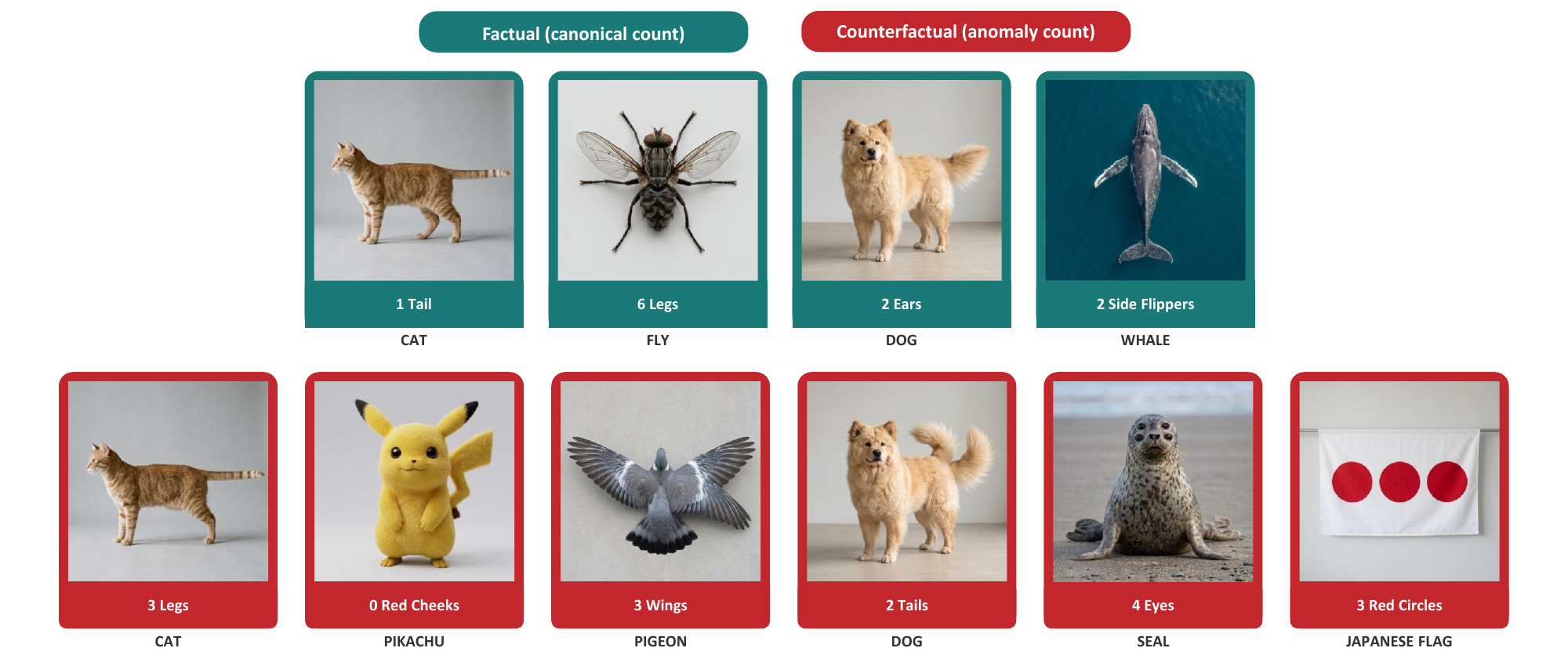}
\caption{The 10 images used in the qualification validation set.
Images with a teal border are factual (canonical count); images with a red border are counterfactual (anomalous count).
See Table~\ref{tab:validation_set} for the corresponding questions and counts.}
\label{fig:validation_images}
\end{figure}

Table~\ref{tab:validation_set} lists the 10 images used in the qualification trial.
Each row shows the subject, the counted attribute, the count shown in the image, the canonical (expected) count, and whether the image is factual or CF.
The corresponding images are shown in Figure~\ref{fig:validation_images}.
 \begin{table}[ht]
\centering
\caption{Qualification validation set.
Images where the shown count equals the canonical count are factual; all others are CF.
The set contains 4 factual and 6 CF images.}
\label{tab:validation_set}
\tiny
\begin{tabular}{lllccl}
\toprule
\textbf{Subject} & \textbf{Attribute} & \textbf{MCQ Question} & \textbf{Shown} & \textbf{Canonical} & \textbf{Type} \\
\midrule
cat          & tail         & How many tails does this cat have?              & 1 & 1 & Factual \\
fly          & leg          & How many legs does this fly have?               & 6 & 6 & Factual \\
dog          & ear          & How many ears does this dog have?               & 2 & 2 & Factual \\
whale        & side flipper & How many side flippers does this whale have?    & 2 & 2 & Factual \\
\midrule
cat          & leg          & How many legs does this cat have?               & 3 & 4 & CF \\
pikachu      & red cheek    & How many red cheeks does this pikachu have?     & 0 & 2 & CF \\
pigeon       & wing         & How many wings does this pigeon have?           & 3 & 2 & CF \\
dog          & tail         & How many tails does this dog have?              & 2 & 1 & CF \\
seal         & eye          & How many eyes does this seal have?              & 4 & 2 & CF \\
Japanese flag & red circle  & How many red circles does this Japanese flag have? & 3 & 1 & CF \\
\bottomrule
\end{tabular}
\end{table}

\subsection{Annotation Protocol}
\label{app:human_protocol}
 
Once qualified, annotators proceeded to the main evaluation.
Each of the 554 counterfactual images and their factual counterparts was independently annotated by 25 qualified annotators.
Rapidata displays each screen for at most 25 seconds, encouraging quick perceptual judgments rather than deliberate pixel-by-pixel analysis.
This platform-imposed time limit mirrors natural human counting, which is rapid and largely automatic for small quantities.
The final answer for each image was determined by majority voting over the 25 responses.

\subsection{Instruction Experiment: Audience Construction}
\label{app:human_instruction}
 
To test whether cueing annotators with the instruction \emph{``Some images are digitally altered, so counts may be unrealistic. Count only what you actually see.''} affects human performance (Section~\ref{sec:cueing}), we constructed two separate audiences from the \emph{same} 10 validation images, following an identical qualification procedure.
The two audiences differed in exactly one respect:
 
\begin{itemize}[leftmargin=*]
    \item \textbf{No-instruction audience.} Qualification and main-task screens showed only the counting question and MCQ options, with no additional instruction.
    \item \textbf{Instruction audience.} Both qualification and main-task screens additionally displayed the ``images may be altered'' caveat before the question.
\end{itemize}
 
\noindent Because the validation images, qualification criteria, and annotation interface were otherwise identical, any difference in performance between the two runs can be attributed to the instruction alone.
Each audience independently annotated the full image set with 25 annotators per image, and final answers were again determined by majority voting.

\begin{figure}[htpb]
\centering
\begin{subfigure}[t]{0.45\textwidth}
    \centering
    \includegraphics[width=0.7\textwidth]{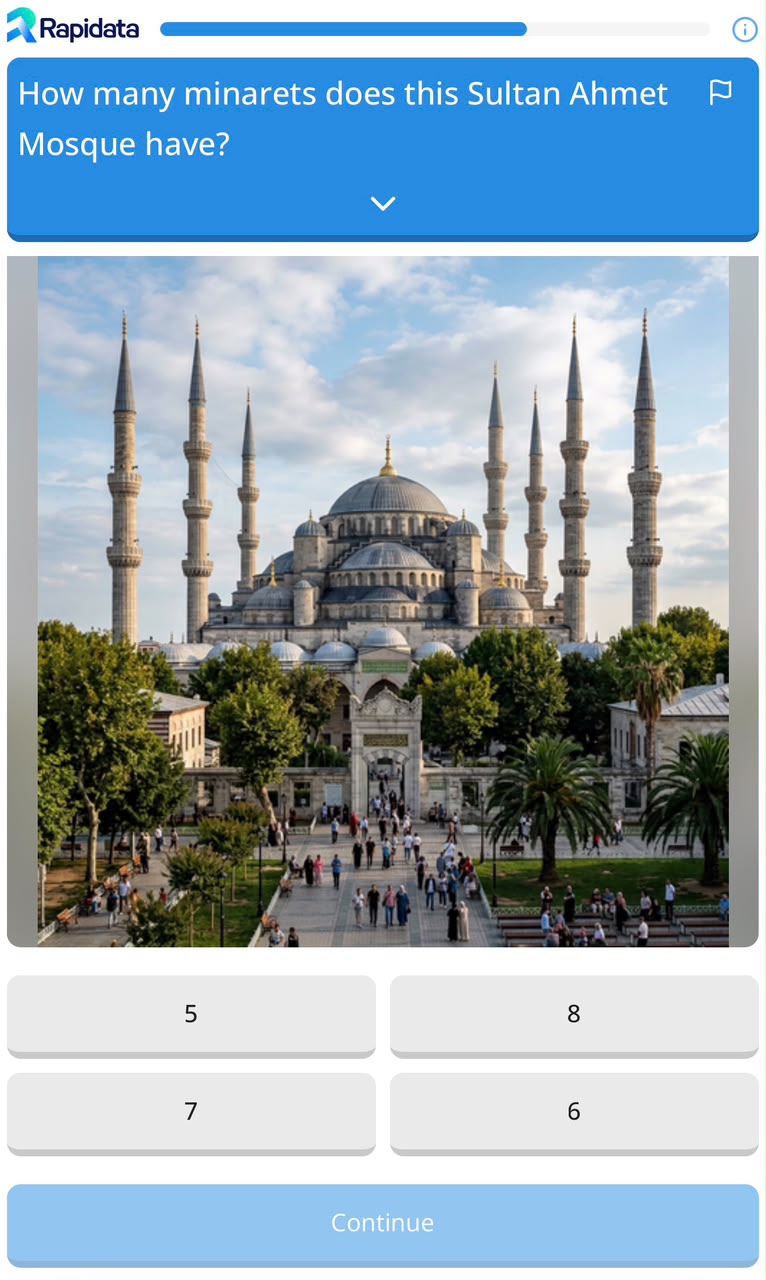}
    \caption{Without instruction.}
    \label{fig:rapidata_no_instruction}
\end{subfigure}
\hfill
\begin{subfigure}[t]{0.45\textwidth}
    \centering
    \includegraphics[width=0.7\textwidth]{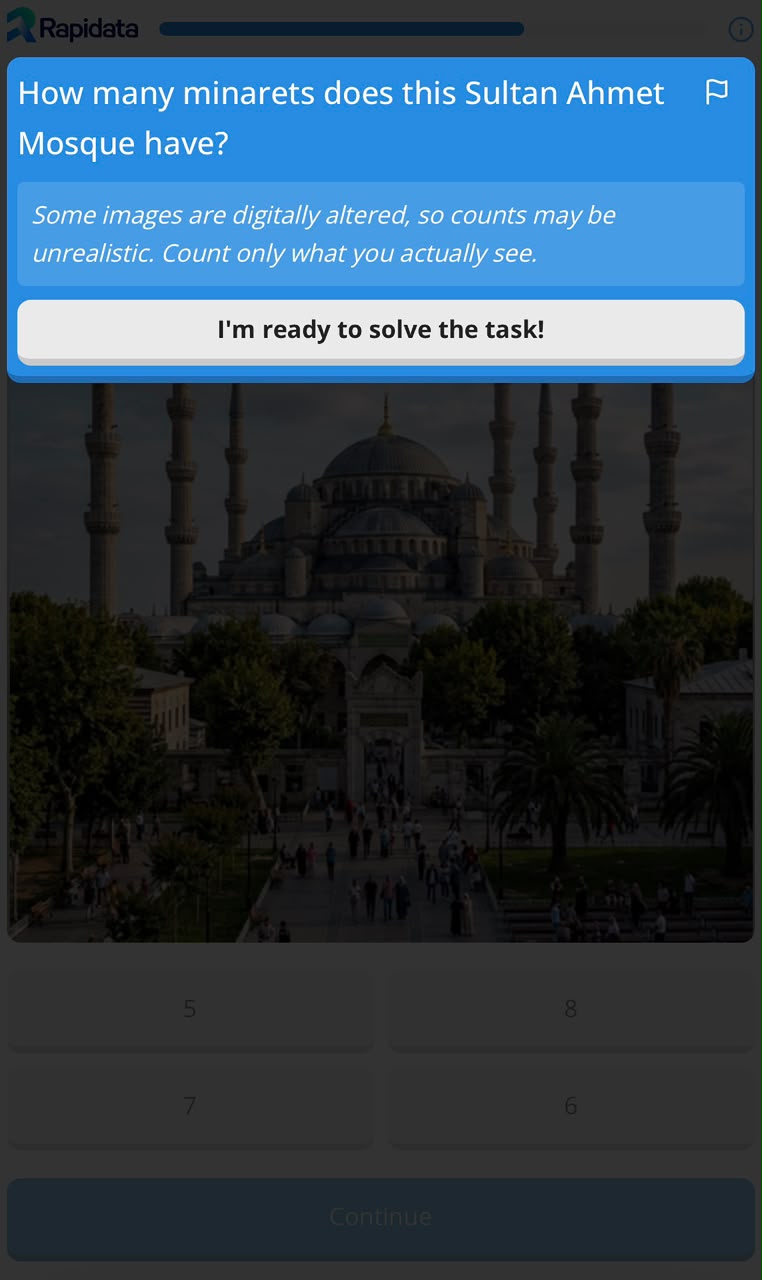}
    \caption{With instruction.}
    \label{fig:rapidata_with_instruction}
\end{subfigure}
\caption{Rapidata annotation interface for the two experimental conditions. \textbf{(a)}~The neutral condition shows only the counting question and four MCQ options. \textbf{(b)}~The instruction condition additionally displays the caveat \emph{ "Some images are digitally altered, so counts may be unrealistic. Count only what you actually see."} above the image.}
\label{fig:rapidata_interface}
\end{figure}

Figure~\ref{fig:rapidata_interface} shows the Rapidata annotation interface for both conditions.
 
\newpage
\end{document}